\documentclass[lettersize,journal]{IEEEtran}
\usepackage{amsmath,amsfonts}
\pdfoutput=1
\usepackage{algorithmic}
\usepackage{algorithm}
\usepackage{array}
\usepackage[caption=false,font=normalsize,labelfont=sf,textfont=sf]{subfig}
\usepackage{textcomp}
\usepackage{stfloats}
\usepackage{url}
\usepackage{verbatim}
\usepackage{graphicx}
\usepackage{cite}
\usepackage{amsmath}
\usepackage[colorlinks,linkcolor=blue]{hyperref}
\usepackage{verbatim}
\usepackage{multirow} 
\usepackage{xcolor}
\usepackage{hyperref}
\usepackage{makecell}
\usepackage{graphics}
\usepackage{threeparttable}
\newcommand{\etal}{\emph{et~al.~}}
\newcommand{\ie}{\emph{i.e.}}

\newcommand{\etc}{\emph{etc}}

\begin{document}

\title{Benchmarking Adversarial Patch Against\\ Aerial Detection}

\author{Jiawei~Lian,~\IEEEmembership{Graduate Student Member,~IEEE},
        Shaohui~Mei,~\IEEEmembership{Senior Member,~IEEE},\\
        Shun~Zhang,~\IEEEmembership{Member,~IEEE},
        and~Mingyang~Ma,~\IEEEmembership{Graduate Student Member,~IEEE}

\thanks{This work was supported in part by the National Natural Science Foundation of China (62171381 and 62271409), and in part by the Fundamental Research Funds for the Central Universities. (Corresponding author: Shaohui Mei.)}

\thanks{Jiawei Lian, Shaohui Mei, Shun Zhang, and Mingyang Ma are with the School of Electronics and Information, Northwestern Polytechnical University, Xi'an 710129, China (email: lianjiawei@mail.nwpu.edu.cn; meish@nwpu.edu.cn; szhang@nwpu.edu.cn; mamingyang@mail.nwpu.edu.cn).}
}

\markboth{Journal of \LaTeX\ Class Files,~Vol.~14, No.~8, August~2021}%
{Shell \MakeLowercase{\textit{et al.}}: A Sample Article Using IEEEtran.cls for IEEE Journals}


\maketitle

\begin{abstract}

Deep neural networks (DNNs) have become essential for aerial detection. However, DNNs are vulnerable to adversarial examples, which poses great security concerns for security-critical systems. To physically evaluate the vulnerability of DNNs-based aerial detection methods, researchers recently devised adversarial patches. Nonetheless, adversarial patches generated by existing algorithms are not strong enough and extremely time-consuming. Moreover, the complicated physical factors are not accommodated well during the optimizing process. In this paper, a novel adaptive-patch-based physical attack (AP-PA) framework is proposed to alleviate the above problems, which achieves state-of-the-art performance in both accuracy and efficiency. Specifically, the AP-PA aims to generate adversarial patches that are adaptive in both physical dynamics and varying scales, and by which the particular targets can be hidden from being detected. Furthermore, the adversarial patch is also gifted with attack effectiveness against all targets of the same class with a patch outside the target (No need to smear targeted objects) and robust enough in the physical world. In addition, a new loss is devised to consider more available information of detected objects to optimize the adversarial patch, which can significantly improve the patch's attack efficacy (Average precision drop up to $87.86\%$ and $85.48\%$ in white-box and black-box settings, respectively) and optimizing efficiency. We also establish one of the first comprehensive, coherent, and rigorous benchmarks to evaluate the attack efficacy of adversarial patches on aerial detection tasks. Finally, several proportionally scaled experiments are performed physically to demonstrate that the elaborated adversarial patches can successfully deceive aerial detection algorithms in dynamic physical circumstances. The code is available at \url{https://github.com/JiaweiLian/AP-PA}.

\end{abstract}


\section{Introduction}

\begin{figure}[t!]
\begin{center}
\includegraphics[width=4.3cm]{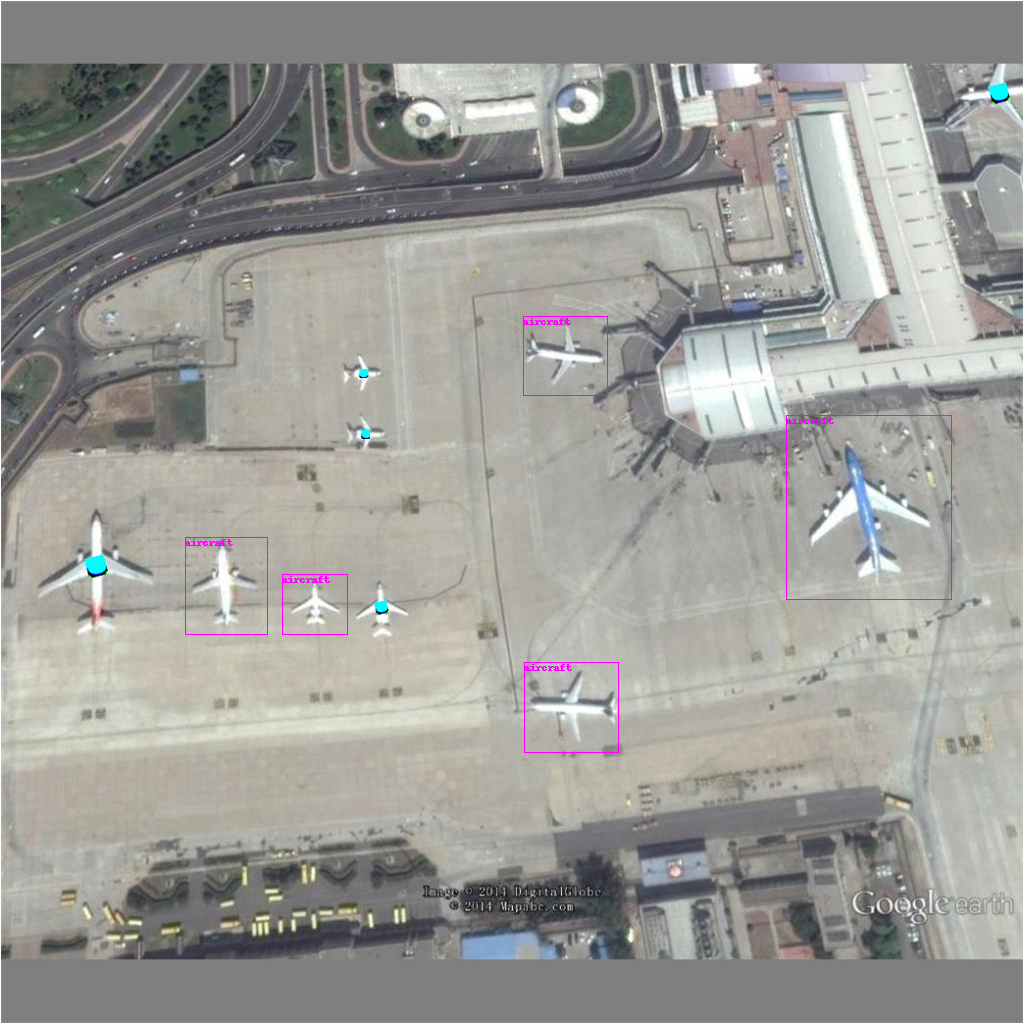}
\includegraphics[width=4.3cm]{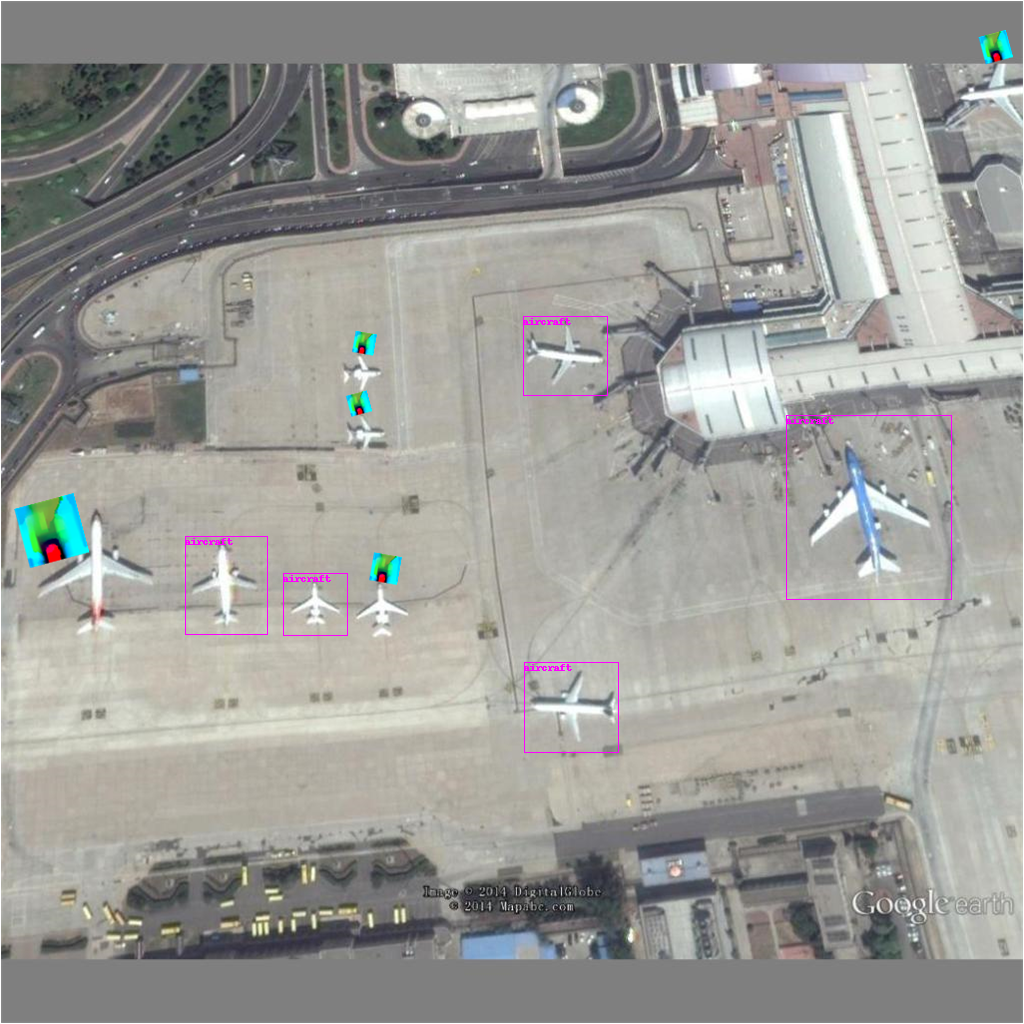}
\end{center}
\caption{The attack effectiveness of adversarial patches with different positions (Patches are pasted on targets or placed outside targets).}
\label{fig_on_outside}
\end{figure}

\IEEEPARstart{W}{ith} the development of deep neural networks (DNNs), DNNs-based aerial detection approaches \cite{mei2017learning,zhang2022spectral,mei2019unsupervised,mei2021accelerating,mei2021hyperspectral} have shown excellent performance both in accuracy and efficiency. However, DNNs are vulnerable to elaborately designed adversarial examples \cite{szegedy2014intriguing,ian2015explaining}. By adding a small malicious perturbation to the clean examples, the DNNs-based systems can make a completely different prediction, which may cause severe consequences in some security-critical areas. In this context, adversarial robustness is regarded as the key performance of DNNs-based intelligent aerial detection systems. The study of new adversarial attack methods provides a data basis for improving the adversarial robustness of DNNs, which also provides ideas for explaining the vulnerability of DNNs to adversarial examples. Nonetheless, most of the existing adversarial attack methods focus on digital attacks and individual object detectors. In addition, attacking object detectors is more challenging than attacking image classifiers, especially extending the digital attack to the physical world, because it requires the adversarial perturbation to be robust enough to survive real-world distortions from many uncontrollable physical dynamics.

In the physical world, however, DNNs-based aerial detection systems work by directly scanning objects. So most of the existing works change the object's appearance in the physical scenarios to provide adversarial examples to the remote sensing detection devices, which poses great challenges, especially needs to solve complicated physical conditions such as different viewing distances, object scales, and lighting conditions. To make adversarial examples practical in real scenarios, some latest works propose the adversarial patch \cite{brown2017adversarial}. They elaborate an adversarial patch that does not attempt to subtly sway an existing object to another. Instead, this attack method generates an image-agnostic patch that is extremely salient to a neural network. This patch can then be pasted anywhere within the field of view of the classifier and renders the classifier to predict a targeted wrong class. Up to now, adversarial patches have been applied to different tasks, such as face recognition \cite{sharif2016accessorize,dong2019efficient,wei2022adversarial}, object detection \cite{xie2017adversarial,song2018physical}, pedestrian detection \cite{thys2019fooling,wang2019advpattern}, image retrieval\cite{wei2022adversarial,chen2022adversarial}, aerial detection \cite{du2022physical,lu2021scale}, \etc.

Albeit the great success of adversarial patches for physical attacks, they have several limitations. First of all, the procedure of generating an adversarial patch is extremely time-consuming, because the adversarial patch is iteratively optimized on a large amount of data. Secondly, the adversarial patch will face a complex transformation from the digital domain transferring to the physical world. Hence, these operations lead to high computation costs. Thirdly, the patch's pixel values will inevitably become distorted due to the limitation of patch printing devices and image capture devices. Last but not least, the current adversarial patches are painted or pasted on the surface of objects, which is not flexible and convenient enough to be applied in real scenarios and is prone to arouse human suspicion. 

Considering the above reasons, this paper dedicates to solving the following problems: under the real physical scenarios, how to generate the adversarial patch to easily achieve stealthy attack effectiveness; in addition, the adversarial patch is robust to complex physical changes and can be used flexibly and conveniently.

Technically, to search for the appropriate physical attack framework, we propose to construct a novel state-of-the-art method to generate an adversarial patch to hide the targets from being detected as shown in Fig.\ref{fig_on_outside}. We devote to designing a physically robust adversarial patch, in which the physical varying factors and different object scales are properly accommodated and the patch is gifted with strong attack effectiveness against all targets of the same class. Moreover, we make full use of the information from all of the detected objects to optimize the adversarial patch, which can significantly improve the adversarial patch's attack efficacy and optimize efficiency. To make a comprehensive evaluation, we also establish one of the first coherent and rigorous benchmarks to evaluate the attack efficacy of adversarial patches on aerial detection tasks. Our method is comprehensively verified in several state-of-the-art object detection frameworks, such as one-stage detector, two-stage detector, CNN-based detector, Transformer-based detector, \etc. Extensive experiments demonstrate the proposed method is effective and robust in complex physical conditions and has a certain transferability for different aerial object detectors.

In summary, our contributions are four-fold:

\begin{enumerate}

\item A novel adaptive-patch-based framework AP-PA is devised to conduct the physical attacks and it achieves state-of-the-art attack performance. On the one hand, our method can elaborate an adaptive adversarial patch accommodating both physical dynamics and varying scales to hide the particular targets from being detected. On the other hand, the adversarial patch is gifted with strong attack effectiveness against all targets of the same class. Additionally, the patch can be easily and conveniently used in real scenarios, simply placed beside targets, which can make the attack happen.

\item A new objective loss is proposed to make full use of the detected information, which can not only accelerate the optimization process of the adversarial patch but also strengthen its attack efficacy both in white-box (Average precision drop up to $87.86\%$) and black-box settings (Average precision drop up to $85.48\%$). Moreover, the elaborated adversarial patches can also transfer their attack effectiveness well between different aerial detectors.

\item To the best of our knowledge, we are the first to comprehensively benchmark adversarial patches against several mainstream aerial detection methods with different frameworks (One-stage, Two-stage, CNN-based, and Transformer-based detectors). In addition, we also delve into the impact of resolution and location of the adversarial patch on attack efficacy.

\item We verify the proposed method on all kinds of object detectors, and the experimental results show that our method naturally maintains attack efficacy and with a certain generalization between different detectors. In addition, we also conduct proportionally scaled validation experiments in the physical world, which demonstrate that the adversarial patch crafted by AP-PA can be robust enough to successfully fool aerial detectors in dynamic physical conditions.

\end{enumerate}

The remainder of this paper is organized as follows. Section II briefly reviews the related work of adversarial attacks. Then, we introduce the details of the proposed framework AP-PA for generating adversarial patches against aerial detection tasks in Section III. We evaluate the proposed attack method and demonstrate the effectiveness of our generated adversarial patches in Section IV. Finally, we conclude our proposed AP-PA and discuss some future work concerning adversarial patches in Section V.

\section{Related work}

In this section, we first provide the background knowledge of the adversarial attack. Additionally, we also review the related works about digital attacks, physical attacks, and physical attacks in aerial detection, respectively.

\subsection{Digital Attacks}

Most existing works concerning adversarial attacks focus on image classification in the digital domain \cite{szegedy2014intriguing,ian2015explaining,liu2022towards,shi2022query,ma2021simulating,mahmood2021robustness,ilyas2019adversarial}. Given an image classifier $ f(\boldsymbol{x}) : \boldsymbol{x} \in \boldsymbol{X} \to y \in \boldsymbol{Y} $ that outputs a prediction $y$ as the result for an input image $\boldsymbol{x}$, the purpose of adversarial attack is to elaborate an adversarial example $ \boldsymbol{x} ^ {*} $ near to clean example $\boldsymbol{x}$ but leading to the classifier making a wrong prediction. Technically, adversarial attack methods can be divided as \textbf{non-targeted} and \textbf{targeted} ones according to attacker's intentions. For a properly classified input image $\boldsymbol{x}$ with ground-truth label $y$ such that $ f(\boldsymbol{x}) = y $, non-targeted attack methods design adversarial example $ \boldsymbol{x} ^ {*} $  by adding imperceptible perturbation to clean images $ \boldsymbol{x} $, but fools the classifier as $ f(\boldsymbol{x}^*) \not= y $, which mainly used for image classification, automatic driving, and object detection tasks; while the targeted adversarial attack methods aim to misguide the classifier by predicting a particular label as $ f(\boldsymbol{x}^*) = y^* $, where $y^*$ is the target label specified by the attacker and $ y^* \not= y $, which are often applied to attacking face recognition, image classification, and automatic driving tasks. Usually, the $L_p$ norm is adopted as the visibility metric of the adversarial noise. For digital attack, the adversarial noise is required to be invisible to human eyes, namely less than an allowed value $\epsilon$ as $ {\Vert \boldsymbol{x}^* - \boldsymbol{x} \Vert}_p \le \epsilon $.

Existing methods can be categorized into three types according to how the adversarial samples are generated. In this paper, we focus on the non-targeted version of attack approaches, and the targeted version can be derived similarly.

\textbf{Optimization-based methods} L-BFGS \cite{szegedy2014intriguing}, Deepfool \cite{moosavi2016deepfool}, C\&W \cite{carlini2017towards}, \etc. directly minimize the distance between the clean and adversarial examples subject to the misclassification of adversarial examples, which can be defined as:
\begin{equation}
    \label{eq_optimization-based}
    \mathop{\arg\min}\limits_{\theta} \lambda \cdot {\Vert \boldsymbol{x}^* - \boldsymbol{x} \Vert}_p - L(\boldsymbol{x}^*, y),
\end{equation}
where $L(\boldsymbol{x}^*, y)$ is the loss function w.r.t. $\boldsymbol{x}^*$. Since it directly minimizes the distance between an adversarial example and the corresponding clean example, the $L_p$ norm is not necessarily inferior to a specified value. 

\textbf{Gradient-based one-step methods}, such as the fast gradient sign method (FGSM) \cite{ian2015explaining}, seek an adversarial example $\boldsymbol{x}^*$ by maximizing $L(\boldsymbol{x}^*, y)$. FGSM generates adversarial examples to meet the $L_\infty$ norm limitation $ {\Vert \boldsymbol{x}^* - \boldsymbol{x} \Vert}_p \le \epsilon $ as:
\begin{equation}
    \label{eq_one-step1}
    \boldsymbol{x}^* = \boldsymbol{x} + \epsilon \cdot {\rm sign} (\nabla_{\boldsymbol{x}}L(\boldsymbol{x},y)),
\end{equation}
where $\nabla_{\boldsymbol{x}}L(\boldsymbol{x},y)$ is the gradient of the loss function w.r.t. $\boldsymbol{x}$. A generalization of FGSM is to meet the $L_2$ norm constraint $ {\Vert \boldsymbol{x}^* - \boldsymbol{x} \Vert}_2 \le \epsilon $ as:
\begin{equation}
    \label{eq_one-step2}
    \boldsymbol{x}^* = \boldsymbol{x} + \epsilon \cdot \frac{\nabla_{\boldsymbol{x}}L(\boldsymbol{x},y)}{{\Vert         \nabla_{\boldsymbol{x}}L(\boldsymbol{x},y) \Vert}_2}.
\end{equation}

\textbf{Gradient-based iterative methods} I-FGSM \cite{kurakin2018adversarial}, MI-FGSM \cite{dong2018boosting}, and PGD \cite{madry2018towards} iteratively apply one-step methods multiple times with a small step size $\alpha$. The iterative attack methods can be defined as:
\begin{equation}
    \label{eq_iterative}
    \boldsymbol{x}^*_0 = \boldsymbol{x}, \quad \boldsymbol{x}^*_{t+1} = \boldsymbol{x}^*_{t} + \alpha \cdot {\rm sign} (\nabla_{\boldsymbol{x}}L(\boldsymbol{x}^*_{t},y)).
\end{equation}
To make the generated adversarial perturbations imperceptible to humans, \ie, meet the $L_p$ constraint, which can be achieved by simply clipping $\boldsymbol{x}^*_t$ into the $\epsilon$ vicinity of $\boldsymbol{x}$ or simply set $\alpha = \epsilon / T $ with $T$ being the number of iterations.

\subsection{Physical Attacks}

Physical attacks play a progressively critical role considering their considerable practical values. To make adversarial perturbations effective in real scenarios, bountiful works have been introduced. In \cite{kurakin2018adversarial}, the feasibility of physical attacks is verified by the fact that the adversarial examples being captured by the imaging device still have attack efficacy. The expectation over transformation (EOT) \cite{athalye2018synthesizing} algorithm makes adversarial examples robust to dynamic physical conditions.

The adversarial patch \cite{brown2017adversarial} is the most frequently used physical attack approach and has been widely applied in many computer vision tasks, such as automatic driving, face recognition, and object detection. We give the reviews in detail as follows:

For automatic driving systems, \cite{eykholt2018robust} devises Robust Physical Perturbations to generate physical perturbations that can steadily fool a DNN-based classifier under physical dynamic conditions. The authors in \cite{duan2020adversarial} elaborate and camouflage adversarial noises into a natural appearance that looks legitimate to human observers to design adversarial traffic signs. The translucent patch in \cite{zolfi2021translucent} is the first camera-based physical attack method, in which the patch is placed on a camera lens rendering the automatic driving system failed to detect the traffic sign. Some other works that take the safety of autonomous driving into account can also be found in \cite{eykholt2017note,sitawarin2018darts}.

For face recognition systems, \cite{sharif2016accessorize} shows how an attacker that is aware or unaware of the internals of a face recognition system can physically achieve impersonation and dodge attacks. Later, some researchers generate eyeglasses \cite{sharif2019general}es, makeup \cite{zhu2019generating}, light \cite{nguyen2020adversarial}s, and hat \cite{komkov2021advhat} with adversarial perturbations attached to deceive the face recognition systems. In \cite{wei2022adversarial}, they propose the meaningful adversarial sticker, by manipulating the fusing operation and parameters of real stickers on the objects instead of designing perturbation patterns like most existing works. Several other relevant studies \cite{kaziakhmedov2019real,pautov2019adversarial} place patches with an attacking effect onto the face or the wearable accessory. 

For objection detection systems, Song \etal \cite{song2018physical} broaden physical attacks to more difficult object detection tasks and introduce the ``Disappearance Attack''. In their work \cite{wu2020making}, the authors delve into physical scenario attacks using printed patches and clothes and quantify their attack effectiveness with different metrics.  Adversarial T-shirts \cite{xu2020adversarial} adopt the deformable adversarial patch on the T-shirts to attack the person detector. The work  \cite{thys2019fooling} applies patch-based adversarial examples to hide a person from being detected. Some other relevant researches \cite{vellaichamy2022detectordetective,cai2022zero} are also devoted to fooling object detectors.

\subsection{Adversarial Attacks in Aerial Images}

Most of the adversarial attack studies focus on earth-based imagery, such as face recognition, person detection, autonomous driving and so on. Recently, some latest works \cite{chen2021empirical,burnel2021generating,xu2020assessing,xu2022universal} devote to crafting imperceptible perturbations to attack aerial image classifiers in the digital domain, while physical attacks on satellite and aerial imagery have not been extensively exploited. Some researchers apply adversarial patches to fool aerial imagery classifiers \cite{czaja2018adversarial} and detectors \cite{den2020adversarial,lu2021scale,du2021adversarial} in the digital domain without verifying the attack efficacy in the physical world. Du \etal \cite{du2022physical} elaborate an adversarial patch that encloses the target object, which is robust enough against atmospheric conditions and temporal variability.

\section{Methodology}

In this article, we propose a brand-new method called \textbf{a}daptive-\textbf{p}atch-based \textbf{p}hysical \textbf{a}ttack (AP-PA), which aims to generate adversarial patch to hide objects from aerial detectors in the physical world. We choose aircraft as the target object, and the different target objects can be simply derived. In this section, we first define the problem to be solved, and then we give an overview of the proposed AP-PA. Finally, we introduce the design of the adaptive patch and objective function in detail respectively.

\subsection{Problem Formulation}

In the aerial detection task, given a benign aerial image $\boldsymbol{x}$, the purpose of the adversarial attack is to make the aerial detection method ignore the target object of the maliciously designed aerial imagery $\boldsymbol{x}^*$. Technically, the adversarial example with elaborated patches can be formulated as:
\begin{equation}
    \label{eq_problem-formulation}
    \boldsymbol{x}^* = (1-\boldsymbol{M}_{\boldsymbol{p}^*}) \odot \boldsymbol{x} + \boldsymbol{M}_{\boldsymbol{p}^*} \odot \boldsymbol{p}^*,
\end{equation}
where $\odot$ and $\boldsymbol{p}^*$ means Hadamard product and adversarial patch, respectively. Mask matrix $\boldsymbol{M}_{\boldsymbol{p}^*}$ is used to constrict the size, shape, and location of the adversarial patch, where the value of the patch position area is 1.

The existing studies mainly focus on optimizing adversarial patches with them pasted on the targets. In contrast, our approach also delves into the physical attack with the patch outside the targets. In the following sections, we will illustrate how to obtain the excellent adversarial patch speedily with strong adversarial attack efficacy.

\subsection{Overview of AP-PA}

\begin{figure*}[!t]
\begin{center}
\includegraphics*[width=18cm]{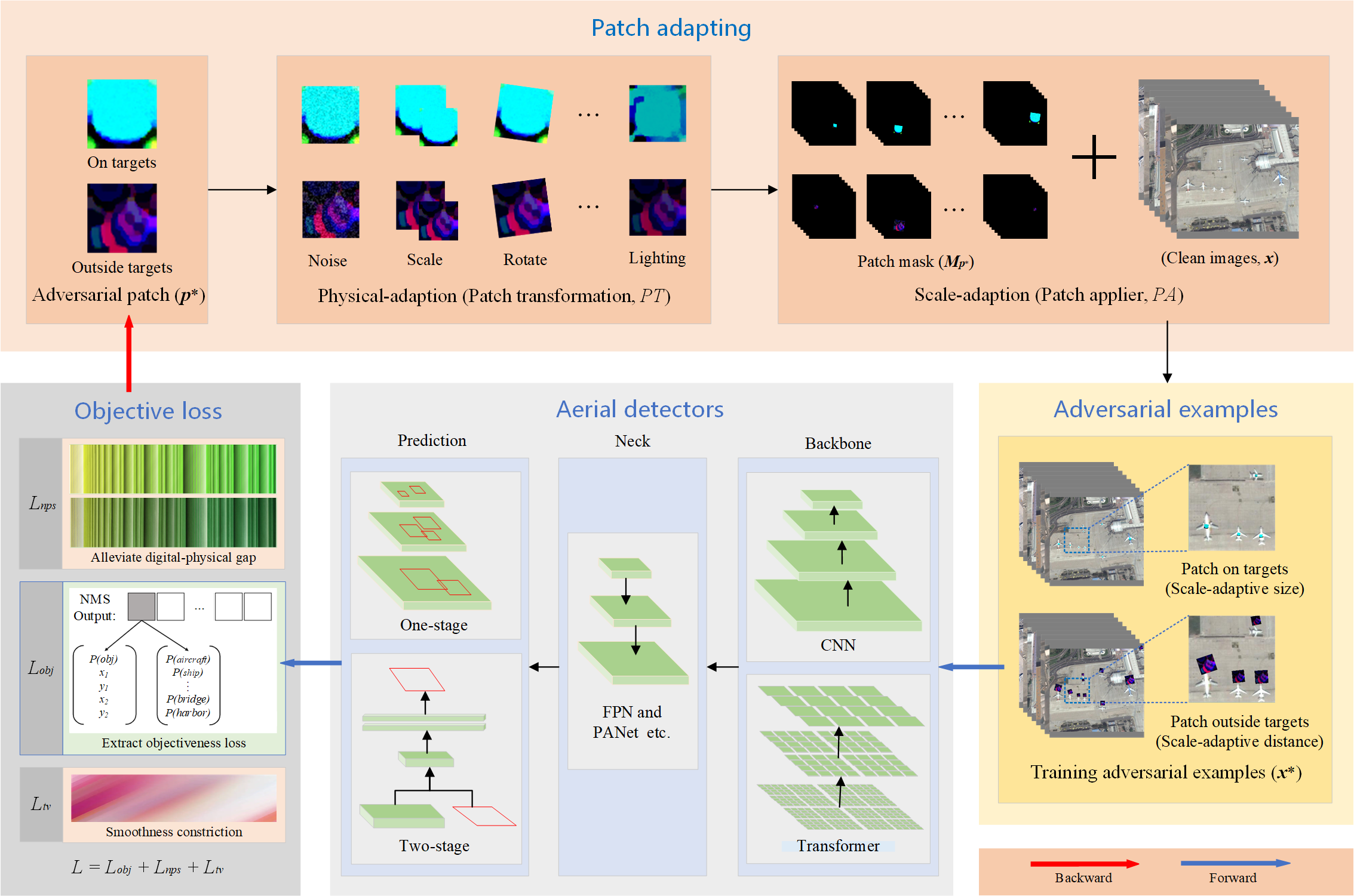}
\end{center}
\caption{The illustration of the proposed AP-PA method for physical attack. Firstly, a series of transform operations are conducted to make the adversarial patch accommodate physical dynamic conditions. Secondly, paste adversarial patches on or outside targets in the proper size and location. Thirdly, the adversarial examples will be fed into a target aerial detector. Next, the objectiveness scores extracted from the detection result are used as part of the total loss. Then, the pixel values of the adversarial patch are optimized to minimize the loss function, including adversarial objectiveness loss ($L_{obj}$), non-printability score ($L_{nps}$), and total variation ($L_{tv}$). Finally, repeat the above procedures until the end of the training
process.}
\label{fig_pipeline}
\end{figure*}

The pipeline of the proposed AP-PA physical attack method is displayed in Fig.\ref{fig_pipeline}. Our purpose is to craft an adversarial patch, which is robust enough in real scenarios and with strong attack effectiveness, and the patch can also be applied outside the target objects. To reach this, the weights and bias of the targeted aerial detectors should be fixed during the training process, instead, the pixel values of the adversarial patch should be updated iteratively, which means we are ``training'' a patch instead of a model with a big set of images containing bountiful of aircraft.

We give a detailed description of the optimizing procedures of the AP-PA approach in the Algorithm \ref{alg:alg1}. Specifically, given an original patch $\boldsymbol{p}^0$, after a series of physical and scale adaptive transformations, the patch $\boldsymbol{p}^*$ will be placed on the clean image $\boldsymbol{x}$ to form an adversarial image $\boldsymbol{x}^*$. Next, the adversarial example will be fed into a targeted aerial detector. Then, the objectiveness scores extracted from the detection result can be used as part of the total loss. The next step is backpropagation, the adversarial patch $\boldsymbol{p}^*$ will be updated. Finally, repeat the above steps until the end of the training process.

\begin{algorithm}
\caption{AP-PA}
\label{alg:alg1}
\begin{algorithmic}[1] 
\REQUIRE Detector $D(\cdot)$, benign aerial image $\boldsymbol{x}$ and ground truth $\boldsymbol{y}$, original patch $\boldsymbol{p}^0$, the adversarial attack loss function $L(\cdot)$, the number of epochs $N_{epo}$ and the number of iterations of each epoch $N_{ite}$, image size $s$, hyperparameter $\alpha, \beta, \eta$
\ENSURE Adversarial patch $\boldsymbol{p}^*$
    \STATE Initialize $\boldsymbol{p}^0$ randomly in [0, 255], $\boldsymbol{p}^*=\boldsymbol{p}^0$;
    \FOR{$i=0$ to $N_{epo}$}
        \FOR{$j=0$ to $N_{ite}$}
            \STATE Patch transformation,
            \STATE $\boldsymbol{p}^* = PT(\boldsymbol{p}^*)$;
            \STATE Patch applier,
            \STATE $\boldsymbol{x}^* = PA({\boldsymbol{p}^*},\boldsymbol{l}_{\boldsymbol{p}^*},w_{\boldsymbol{p}^*},h_{\boldsymbol{p}^*})$;
            \STATE Detection,
            \STATE $\boldsymbol{r} = D(\boldsymbol{x}^*)$;
            \STATE Extract objectiveness loss,
            \STATE $L_{obj} = E(\boldsymbol{r})$;
            \STATE Total loss,
            \STATE $L = L_{obj} + \alpha \cdot L_{tv} + \beta \cdot L_{nps}$;
            \STATE Update patch,
            \STATE $\boldsymbol{p}^*_{i,j+1} = \boldsymbol{p}^*_{i,j} + \eta \cdot \nabla_{\boldsymbol{p}^*_{i,j}}L$;
        \ENDFOR
    \ENDFOR
    \STATE $\boldsymbol{p}^* = \boldsymbol{p}^*_{N_{epo},N_{ite}}$;
    \STATE \textbf{return} $\boldsymbol{p}^*$
\end{algorithmic}
\end{algorithm}

\subsection{Patch Adapting}

\begin{figure}[t!]
\begin{center}
\includegraphics[width=4.3cm]{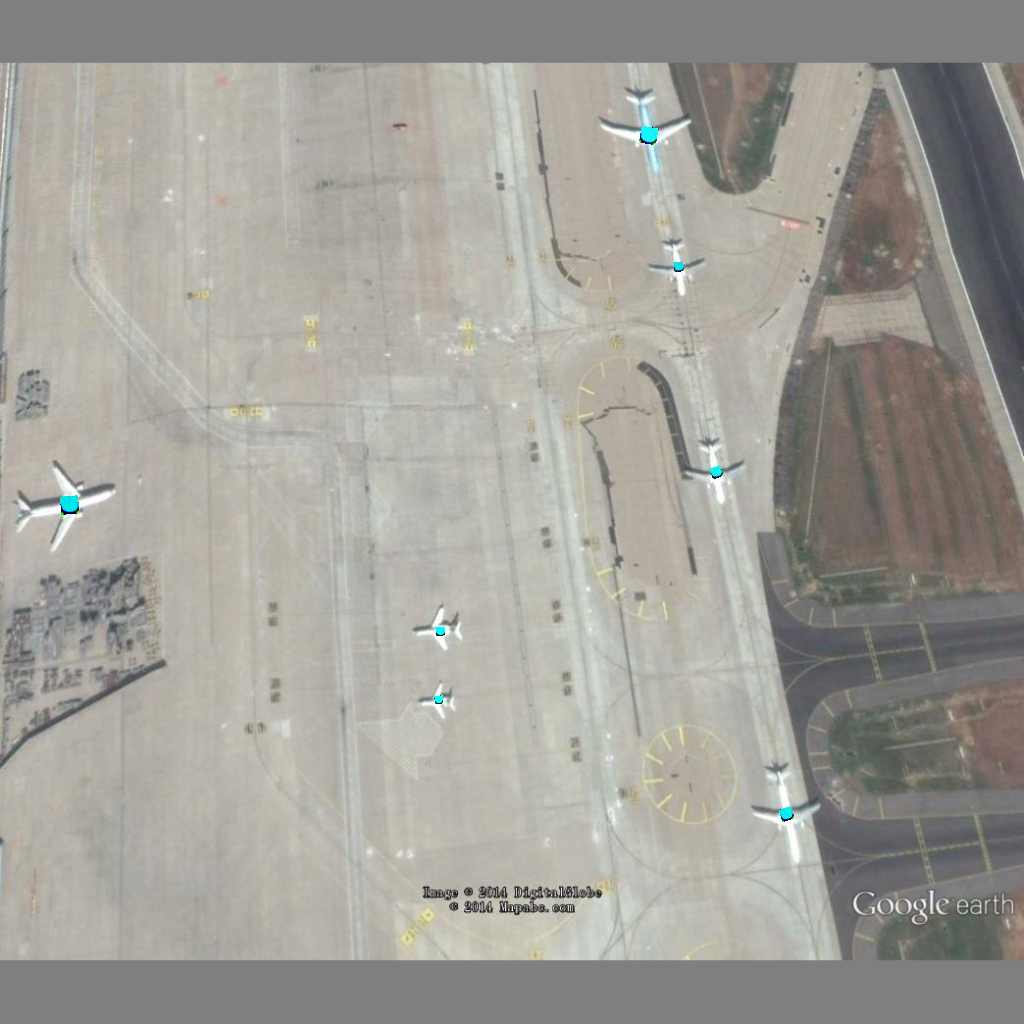}
\includegraphics[width=4.3cm]{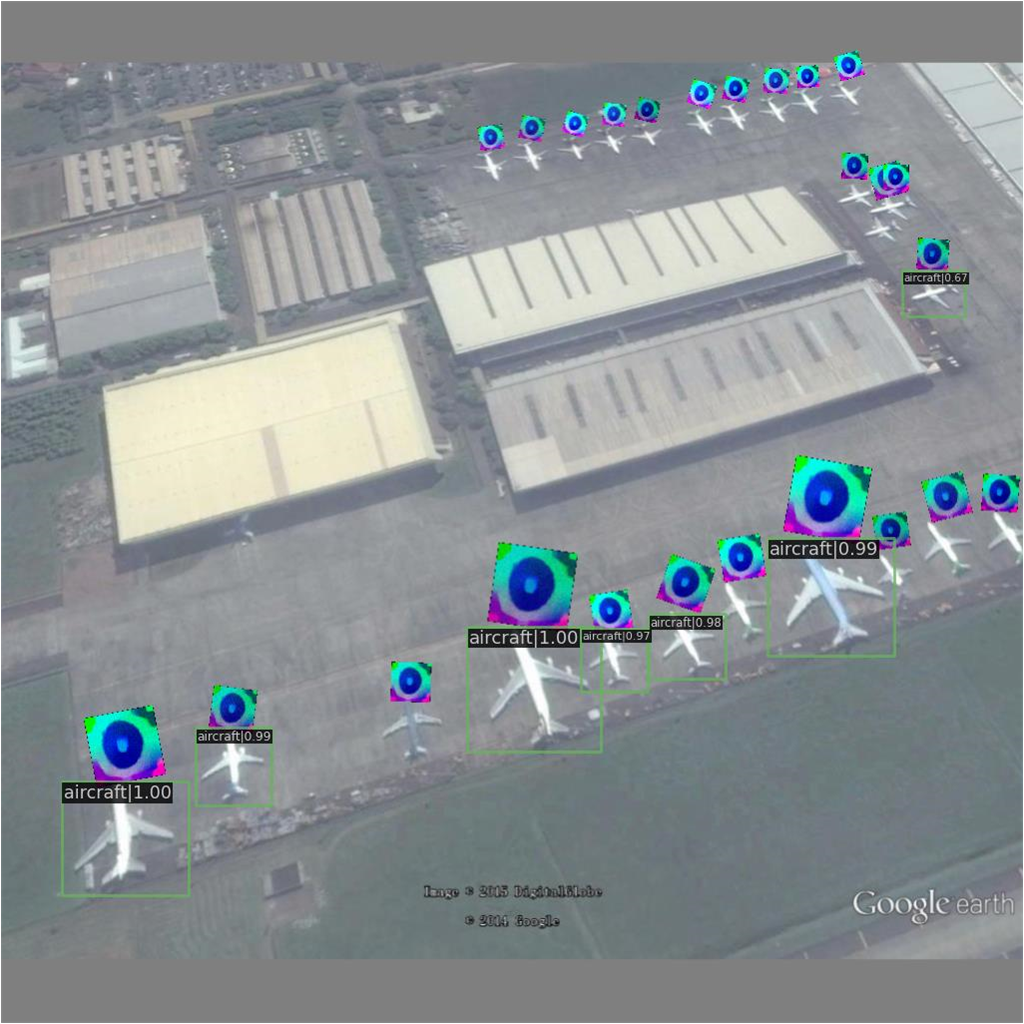}
\end{center}
\caption{Visual examples of scale-adaptive patch size and distance.}
\label{fig_scaled_patch}
\end{figure}

To make the adversarial patch crafted by the AP-PA algorithm successfully fool aerial detection systems in real scenarios, we accommodate the dynamic conditions of the physical world during the process of optimizing the adversarial patches. The real scenarios usually contain varying conditions, including dynamic viewpoint, natural noise, varying lighting, \etc. We adopt several physical transformations to simulate such dynamic factors. Technically, we take the transformations of accommodating physical fluctuated conditions into account, such as adding noise, varying scales, random rotation, lighting shift and so on. The above physical adaptive operations are packed in patch transformation function $PT(\cdot)$, then the adversarial example can be written as:
\begin{equation}
    \label{eq_patch_transformation}
    \boldsymbol{x}^* = (1-\boldsymbol{M}_{\boldsymbol{p}^*}) \odot \boldsymbol{x} + \boldsymbol{M}_{\boldsymbol{p}^*} \odot PT(\boldsymbol{p}^*).
\end{equation}

Next, we focus on how to place the adversarial patches in the proper position with the adaptive size due to the varying scales of objects, as shown in Fig.\ref{fig_scaled_patch}. 

For the patch on target, our goal is to paste an adversarial patch in the center of the object with a proper size. To achieve that, we use the coordinate $(x_1,y_1,x_2,y_2)$ of the detection result to compute the center coordinate of adversarial patch $\boldsymbol{l}_{\boldsymbol{p}^*}$ as:
\begin{equation}
    \label{eq_patch_location}
    \boldsymbol{l}_{\boldsymbol{p}^*} = (\frac{x_1+x_2}{2},\frac{y_1+y_2}{2}).
\end{equation}
Then, considering the different scales of the objects, a scale-adaptive patch method is proposed to tackle this problem. To make the area of the adversarial patch and targeted object keep a proper ratio $r_s$:
\begin{equation}
    \label{eq_patch_size_ratio}
    r_s = \frac{w_{\boldsymbol{p}^*} \cdot h_{\boldsymbol{p}^*}}{w_{\boldsymbol{t}} \cdot h_{\boldsymbol{t}}},
\end{equation}
where the scale adaptive patch size can be calculated by:
\begin{equation}
    \label{eq_patch_size}
    w_{\boldsymbol{p}^*} = h_{\boldsymbol{p}^*} = \sqrt[2]{r_s \cdot w_{\boldsymbol{t}} \cdot h_{\boldsymbol{t}}}.
\end{equation}
Next, the mask of adversarial $\boldsymbol{M}_{\boldsymbol{p}^*}$ is formulated as:
\begin{equation}
    \label{eq_patch_applier}
    \boldsymbol{M}_{\boldsymbol{p}^*} = PA({\boldsymbol{p}^*},\boldsymbol{l}_{\boldsymbol{p}^*},w_{\boldsymbol{p}^*},h_{\boldsymbol{p}^*}),
\end{equation}
where the patch applier function $PA({\boldsymbol{p}^*},\boldsymbol{l}_{\boldsymbol{p}^*},w_{\boldsymbol{p}^*},h_{\boldsymbol{p}^*})$ aims to paste adversarial patch on the corresponding position with an adaptive size. Finally, the original formulated problem (\ref{eq_problem-formulation}) can be transformed as:
\begin{equation}
    \label{eq_patch_transformation2}
    \begin{aligned}
    \boldsymbol{x}^* = (1-PA({\boldsymbol{p}^*},\boldsymbol{l}_{\boldsymbol{p}^*},w_{\boldsymbol{p}^*},h_{\boldsymbol{p}^*})) \odot \boldsymbol{x}\\ + PA({\boldsymbol{p}^*},\boldsymbol{l}_{\boldsymbol{p}^*},w_{\boldsymbol{p}^*},h_{\boldsymbol{p}^*}) \odot PT(\boldsymbol{p}^*).
    \end{aligned}
\end{equation}

For patch outside the target, our strategy is to put the adversarial patch on the top of the target in a scale adaptive distance $d_{\boldsymbol{p}^*}$, \ie, to make $d_{\boldsymbol{p}^*}$ and the height of the target keep an appropriate ratio $r_d$:
\begin{equation}
    \label{eq_patch_distance_ratio}
    r_d = \frac{y_2 - y_1}{d_{\boldsymbol{p}^*}},
\end{equation}
where scale adaptive distance $d_{\boldsymbol{p}^*}$ can be acquired by:
\begin{equation}
    \label{eq_patch_distance}
     d_{\boldsymbol{p}^*}= \frac{y_2 - y_1}{r_d}.
\end{equation}
Next, the central position of the adversarial patch is given by:
\begin{equation}
    \label{eq_patch_location2}
    \boldsymbol{l}_{\boldsymbol{p}^*} = (\frac{x_1+x_2}{2},\frac{y_1+y_2}{2} - d_{\boldsymbol{p}^*}).
\end{equation}
The rest steps can be derived from the procedures of training the adversarial patch on the target.

\subsection{Objective Function Design}

In this article, we aim to design a novel algorithm that can be used to craft a printable adversarial patch, which is capable to deceive aerial detectors. To achieve that, we adopt an optimization process (update patch pixel values) to train an adversarial patch that, on a big dataset, significantly drops the average precision (AP) of the particular target of aerial detection.

Our objective function contains three parts:

\subsubsection{Objectiveness loss $L_{obj}$}

We use the mean of all objectiveness scores of detected objects after non-maximum suppression operation as adversarial objective loss, which can be written as:
\begin{equation}
    \label{eq_Lobj}
    L_{obj} = E(\boldsymbol{r}) = \frac{1}{n} \sum\limits_{i=1}^n P_i(obj),
\end{equation}
where $\boldsymbol{r}$ is the detection results of aerial detectors, and $E(\boldsymbol{r})$ means extracting objectiveness loss $L_{obj}$ from $\boldsymbol{r}$ that contains $n$ detected object(s), including the coordinate $(x_1,y_1,x_2,y_2)$, objective score $P(obj)$, and class scores such as $(P(aircraft),P(ship),...,P(bridge),P(harbor))$ of each object. The purpose of the adversarial patch is to hide aircraft in the aerial image. To achieve this, we aim to lower the object or class score predicted by the aerial detector. The reason why we do not consider class scores in the loss function is that minimizing the class score of aircraft tends to increase the score of a different class. Moreover, \cite{thys2019fooling} demonstrates that taking the class score into account can not acquire a stronger attack efficacy.

\subsubsection{Total variation loss $L_{tv}$}

To overcome the problem that the value gap between adjacent pixels is difficult to capture by image acquisition devices, we add total variation as described in \cite{sharif2016accessorize} into objective function. $L_{tv}$ tends to guarantee that the optimizer favors the adversarial patch with a smooth pattern and color shift. This loss can be calculated from adversarial patch $\boldsymbol{p}^*$ as follows:
\begin{equation}
    \label{eq_Ltv}
    L_{tv} = \sum\limits_{i,j} \sqrt{(p_{i+1,j}-p_{i,j})^2 + (p_{i,j+1}-p_{i,j})^2},
\end{equation}
where $p_{i,j}$ represents the pixel value of $i$th row, $j$th column of the adversarial patch.

\subsubsection{ Non-printable score loss $L_{nps}$}

Due to the colors shift of the adversarial patch from the digital domain transform to the physical domain, the non-printability score in the work \cite{sharif2016accessorize} is introduced to show how well the colors in the adversarial patch can be printed in the physical world, which represents the distance of adversarial patch between the digital domain and physical world that printed by a normal printer. Here $L_{nps}$ is formulated as:
\begin{equation}
    \label{eq_Lnps}
    L_{nps} = \sum\limits_{i,j} \min_{c_{print} \in C} \mid p_{i,j} - c_{print} \mid,
\end{equation}
where $c_{print}$ is one color in a group of physical printable colors set $C$. Taking this loss into account makes the pixel values of our elaborated adversarial patch favor printable colors from printable colors set $C$.

Out of the above three components follows the total objective function, written as:
\begin{equation}
    \label{eq_total_loss}
    L = L_{obj} + \alpha \cdot L_{tv} + \beta \cdot L_{nps}.
\end{equation}
We use hyperparameter $\alpha$ and $\beta$ to scale $L_{tv}$ and $L_{nps}$ respectively and add up the three parts, then optimize $L$ with Adam \cite{kingma2015adam}. Our proposed AP-PA aims to minimize the objective function $L$ and optimize the adversarial patch, so we freeze all weights and biases in the aerial detection model and only update the pixel values of the adversarial patch. The initial patch $\boldsymbol{p}^0$ is gifted with random values at the start of the optimizing process. 

\section{Experiments}

\begin{figure*}[t!]
\begin{center}

\includegraphics[width=1.7cm]{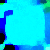}
\includegraphics[width=1.7cm]{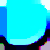}
\includegraphics[width=1.7cm]{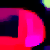}
\includegraphics[width=1.7cm]{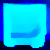}
\includegraphics[width=1.7cm]{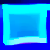}
\includegraphics[width=1.7cm]{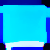}
\includegraphics[width=1.7cm]{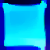}
\includegraphics[width=1.7cm]{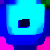}
\includegraphics[width=1.7cm]{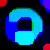}
\includegraphics[width=1.7cm]{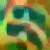}
\end{center}
\begin{center}
\includegraphics[width=1.7cm]{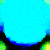}
\includegraphics[width=1.7cm]{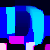}
\includegraphics[width=1.7cm]{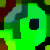}
\includegraphics[width=1.7cm]{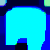}
\includegraphics[width=1.7cm]{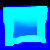}
\includegraphics[width=1.7cm]{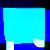}
\includegraphics[width=1.7cm]{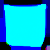}
\includegraphics[width=1.7cm]{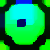}
\includegraphics[width=1.7cm]{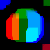}
\includegraphics[width=1.7cm]{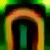}
\end{center}
\begin{center}
\includegraphics[width=1.7cm]{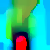}
\includegraphics[width=1.7cm]{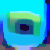}
\includegraphics[width=1.7cm]{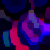}
\includegraphics[width=1.7cm]{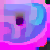}
\includegraphics[width=1.7cm]{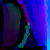}
\includegraphics[width=1.7cm]{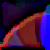}
\includegraphics[width=1.7cm]{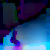}
\includegraphics[width=1.7cm]{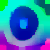}
\includegraphics[width=1.7cm]{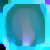}
\includegraphics[width=1.7cm]{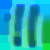}

\end{center}
\caption{Elaborated adversarial patches by different methods. From left to right, the target detectors are YOLOv2, YOLOv3, YOLOv5n, YOLOV5s, YOLOv5m, YOLOv5l, YOLOv5x, Faster R-CNN, SSD, and Swin Transformer, respectively.}
\label{fig_patches}
\end{figure*}

\begin{table}[t!]
\caption{The detailed description of RSOD and DOTA datasets.}
\label{table_datasets}
\centering
\begin{tabular*}{\hsize}{cccccc}
\hline\hline
Datasets    & Categories    & Images    & Instances     & Image width   & Year      \\ \hline\hline
RSOD        & 4             & 976       & 6950          & $\sim$1000    & 2017      \\
DOTA        & 15            & 2806      & 188282        & 800-4000      & 2018      \\ \hline\hline
\end{tabular*}
\end{table}

In this part, we perform comprehensive experiments to verify the attack efficacy of the proposed AP-PA algorithm. We first describe the  experimental settings in detail in Sec. \ref{section1}. Then we specify the results for the digital attack in Sec. \ref{section2} and physical attack in Sec. \ref{section3}.

\subsection{Experimental Settings}
\label{section1}

\subsubsection{Target models}
We choose several representative aerial detection models, such as one-stage detectors (YOLOv2 \cite{redmon2017yolo9000}, YOLOv3 \cite{redmon2018yolov3}, YOLOv5 \cite{jocher2020yolov5}, and SSD \cite{liu2016ssd}), two-stage detector (Faster R-CNN \cite{ren2015faster}), and Transformer-based detector (Swin Transformer \cite{liu2021swin}) as attack target models. The open-source codes (YOLOv2{\footnote{\url{https://github.com/ringringyi/DOTA_YOLOv2}}, YOLOv3\footnote{\url{https://github.com/ultralytics/yolov3}}, YOLOv5\footnote{\url{https://github.com/ultralytics/yolov5}}, MMDetection\footnote{\url{https://github.com/open-mmlab/mmdetection}}}) are adopted to train the aforementioned aerial detectors. 

\subsubsection{Datasets}
We conduct experiments on two public datasets: RSOD\footnote{\url{https://github.com/RSIA-LIESMARS-WHU/RSOD-Dataset-}} and DOTA\footnote{\url{https://captain-whu.github.io/DOTA/index.html}} \cite{xia2018dota}. The detailed information of the above two datasets is described in Table \ref{table_datasets}. We use the DOTA dataset to train aerial detectors because of its diverse object categories and rich data volumes, and the RSOD dataset is adopted to optimize adversarial patches due to its separate aircraft images. 

\subsubsection{Metrics}
Three metrics, Recall, Precision, and Average Precision (AP), are adopted to evaluate the attack effectiveness of adversarial patches. To verify the attack efficacy towards aerial detectors, it is regarded as a successful attack if the aircraft can be ignored by the aerial detector.

\subsubsection{Implementation}
We refer to the settings in the work \cite{thys2019fooling} and set $\alpha$ and $\beta$ of Eq.(\ref{eq_total_loss}) to 2.5 and 0.01 respectively to balance the three parts of the objective loss $L$. In addition, we empirically set the maximum number of epochs to 600. The iterations $T$ equals the number of training data divided by the batch size, and the thresholds of the intersection of union (IOU) and objectiveness confidence are 0.45 and 0.4 respectively for all aerial detectors both in testing and training. In this paper, the experiments are conducted with the PyTorch platform \cite{paszke2019pytorch} using NVIDIA GeForce RTX3080 (10GB) GPUs.

\subsection{Experimental Results in Digital Domain}
\label{section2}

\subsubsection{Elaborated adversarial patches}
First of all, we present the elaborately crafted adversarial patches by our proposed AP-PA method against YOLOv2, YOLOv3, YOLOv5n, YOLOV5s, YOLOv5m, YOLOv5l, YOLOv5x, Faster R-CNN, SSD, and Swin Transformer respectively, as shown in Fig.\ref{fig_patches}. 

Some interesting properties of adversarial patches can be observed in Fig.\ref{fig_patches} as follows:
\begin{itemize}
\item{The more similar aerial detectors, the more similar corresponding adversarial patches, which have the same pattern style, such as the adversarial patches generated by YOLOv5s, YOLOv5m, YOLOv5l, and YOLOv5x both by the work \cite{thys2019fooling} and AP-PA (On targets). For AP-PA (Outside targets), the adversarial patches corresponding to YOLOv5n, YOLOv5m, YOLOv5l, and YOLOv5x have the same styles;}
\item{Different aerial detectors craft adversarial patches with different styles. For example, a one-stage detector (YOLO) and a two-stage detector (Faster R-CNN) generate adversarial patches with totally different styles. Moreover, CNN-based and Transformer-based detectors also generate adversarial patches with different pattern styles;}
\item{The position of the adversarial patch has a significant influence on the pattern style of adversarial patches. Specifically, there is a slight difference between the patch crafted by Thys \etal \cite{thys2019fooling} and our AP-PA with the patch on targets, while a huge gap exists between patches on and outside targets.}
\end{itemize}

\subsubsection{Attack efficacy}

\begin{table*}[t!]
\caption{The experimental results of the adversarial attack with the patches on targets.}
\label{table_on_target}
\centering
\setlength{\tabcolsep}{1.4mm}
\begin{threeparttable}
\begin{tabular*}{\hsize}{ccccccccccc}
\hline\hline
\diaghead{\theadfont Diag ColumnmnHead}{Patches}{Detectors} &YOLOv2 &YOLOv3 &YOLOv5n    & YOLOv5s   &YOLOv5m    &YOLOv5l    &YOLOv5x    &Faster R-CNN   &SSD    &Swin Transformer
\\ \hline\hline
YOLOv2\cite{redmon2017yolo9000}              &\textbf{6.33}\%    &65.80\%    &80.18\%	&73.05\%	&80.77\%	&78.83\%	&78.44\%	&68.88\%	&47.80\%	&85.98\%     \\
YOLOv3\cite{redmon2018yolov3}              &19.38\%	&\textbf{59.24}\%	&75.36\%	&66.43\%	&74.20\%	&72.54\%	&75.40\%	&35.35\%	&28.88\%	&82.78\%    \\
YOLOv5n\cite{jocher2020yolov5}             &63.90\%	&88.57\%	&\textbf{83.94}\%	&85.28\%	&91.60\%	&89.49\%	&92.36\%	&37.16\%	&39.97\%	&81.17\%    \\
YOLOv5s\cite{jocher2020yolov5}             &9.25\%	    &65.17\%	&78.28\%	&\textbf{63.60}\%	&75.13\%	&74.07\%	&76.48\%	&53.72\%	&38.13\%	&83.81\%    \\
YOLOv5m\cite{jocher2020yolov5}             &12.79\%	&66.94\%	&78.47\%	&67.49\%	&\textbf{73.53}\%	&75.23\%	&78.31\%	&54.49\%	&42.00\%	&83.75\%    \\
YOLOv5l\cite{jocher2020yolov5}             &11.69\%	&65.50\%	&78.31\%	&67.17\%	&74.20\%	&\textbf{72.08}\%	&75.30\%	&56.37\%	&41.16\%	&84.34\%    \\
YOLOv5x\cite{jocher2020yolov5}             &8.71\%	    &65.89\%	&77.64\%	&65.67\%	&75.08\%	&74.53\%	&\textbf{73.90}\%	&55.30\%	&40.28\%	&83.62\%    \\
Faster R-CNN\cite{ren2015faster}        &14.27\%	&76.86\%	&84.57\%	&80.58\%	&85.02\%	&84.35\%	&84.84\%	&\textbf{32.90}\%	&34.29\%	&80.05\%    \\
SSD\cite{liu2016ssd}                 &27.54\%	&72.54\%	&81.50\%	&77.81\%	&80.22\%	&80.60\%	&81.73\%	&30.98\%	&\textbf{25.14}\%	&78.99\%    \\
Swin Transformer\cite{liu2021swin}    &66.06\%	&81.43\%	&84.70\%	&83.82\%	&85.87\%	&86.31\%	&85.75\%	&29.63\%	&33.90\%	&\textbf{73.61}\%    \\
Noise               &94.19\%	&95.09\%	&94.88\%	&95.24\%	&96.37\%	&96.71\%	&96.60\%	&81.42\%	&75.72\%	&89.05\%    \\
\hline\hline
\end{tabular*}
    \begin{tablenotes}
        \footnotesize
        \item White-box attack results are highlighted in \textbf{bold}, and the rest results belong to the black-box attack.
        \item The noise is added to compare the effect of patch occlusion.
    \end{tablenotes}
\end{threeparttable}
\end{table*}

\begin{table*}[t!]
\caption{The experimental results of the adversarial attack with patches outside targets.}
\label{table_outside_target}
\centering
\setlength{\tabcolsep}{1.4mm}
\begin{threeparttable}
\begin{tabular*}{\hsize}{ccccccccccc}
\hline\hline
\diaghead{\theadfont Diag ColumnmnHead}{Patches}{Detectors} &YOLOv2 &YOLOv3 &YOLOv5n    & YOLOv5s   &YOLOv5m    &YOLOv5l    &YOLOv5x    &Faster R-CNN   &SSD    &Swin Transformer
\\ \hline\hline
YOLOv2\cite{redmon2017yolo9000}              &\textbf{20.72}\%	   &72.18\%	&50.86\%	&64.55\%	&64.87\%	&65.96\%	&67.08\%	&42.23\%	&48.27\%	&68.88\%    \\
YOLOv3\cite{redmon2018yolov3}              &54.83\%	&\textbf{64.50}\%	&57.03\%	&66.91\%	&67.38\%	&72.76\%	&68.05\%	&46.40\%	&49.35\%	&57.87\%    \\
YOLOv5n\cite{jocher2020yolov5}             &58.02\%	&64.36\%	&\textbf{39.41}\%	&66.07\%	&66.60\%	&69.62\%	&67.51\%	&40.87\%	&46.64\%	&66.90\%    \\
YOLOv5s\cite{jocher2020yolov5}             &55.53\%	&68.68\%	&58.07\%	&\textbf{60.68}\%	&70.35\%	&73.60\%	&71.47\%	&34.58\%	&47.98\%	&59.42\%    \\
YOLOv5m\cite{jocher2020yolov5}             &55.26\%	&63.88\%	&51.83\%	&53.57\%	&\textbf{55.21}\%	&62.67\%	&61.53\%	&41.23\%	&41.06\%	&63.61\%    \\
YOLOv5l\cite{jocher2020yolov5}             &56.50\%	&67.09\%	&57.96\%	&57.39\%	&63.93\%	&\textbf{54.17}\%	&63.86\%	&33.96\%	&42.60\%	&63.08\%    \\
YOLOv5x\cite{jocher2020yolov5}             &52.90\%	&65.70\%	&57.87\%	&61.31\%	&64.42\%	&69.03\%	&\textbf{62.65}\%	&40.45\%	&48.60\%	&65.09\%    \\
Faster R-CNN\cite{ren2015faster}        &53.15\%	&72.19\%	&56.72\%	&70.27\%	&72.01\%	&75.19\%	&74.19\%	&\textbf{30.27}\%	&49.42\%	&57.72\%    \\
SSD\cite{liu2016ssd}                 &44.84\%	&68.52\%	&54.12\%	&66.02\%	&62.93\%	&71.29\%	&67.11\%	&42.62\%	&\textbf{44.62}\%	&62.94\%    \\
Swin Transformer\cite{liu2021swin}    &47.02\%	&72.05\%	&61.75\%	&71.20\%	&69.90\%	&72.57\%	&70.05\%	&47.02\%	&51.84\%	&\textbf{57.91\%}   \\
\hline\hline
\end{tabular*}
    \begin{tablenotes}
        \footnotesize
        \item For adversarial patches outside targets, there is no need for considering the impact of occlusion.
    \end{tablenotes}
\end{threeparttable}
\end{table*}

\begin{figure*}[!t]
\begin{center}
\includegraphics*[width=18cm]{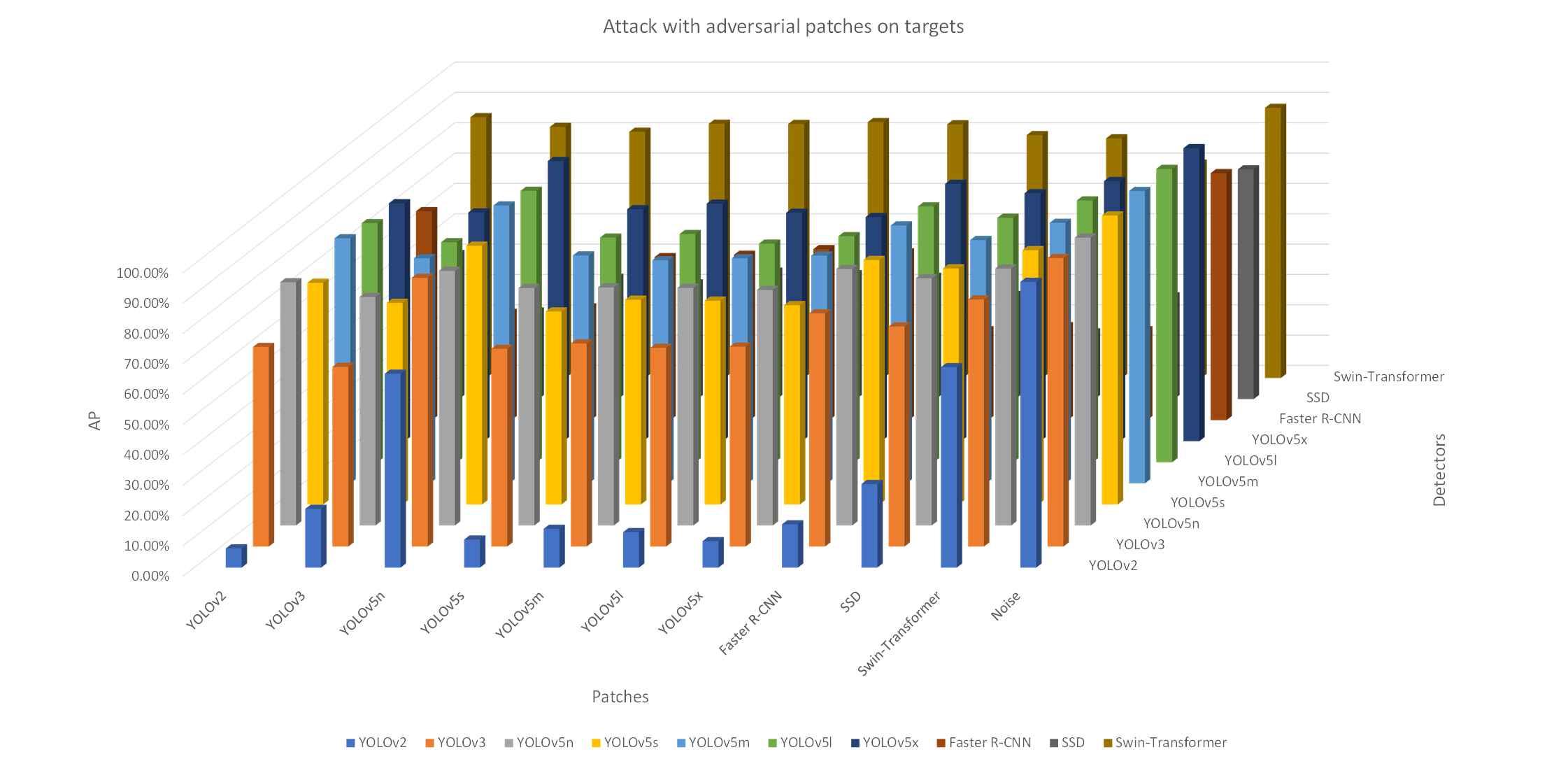}
\end{center}
\caption{The transferability of adversarial attack between different aerial detectors with patches on targets.}
\label{fig_3d_bar_on}
\end{figure*}

\begin{figure*}[!t]
\begin{center}
\includegraphics*[width=18cm]{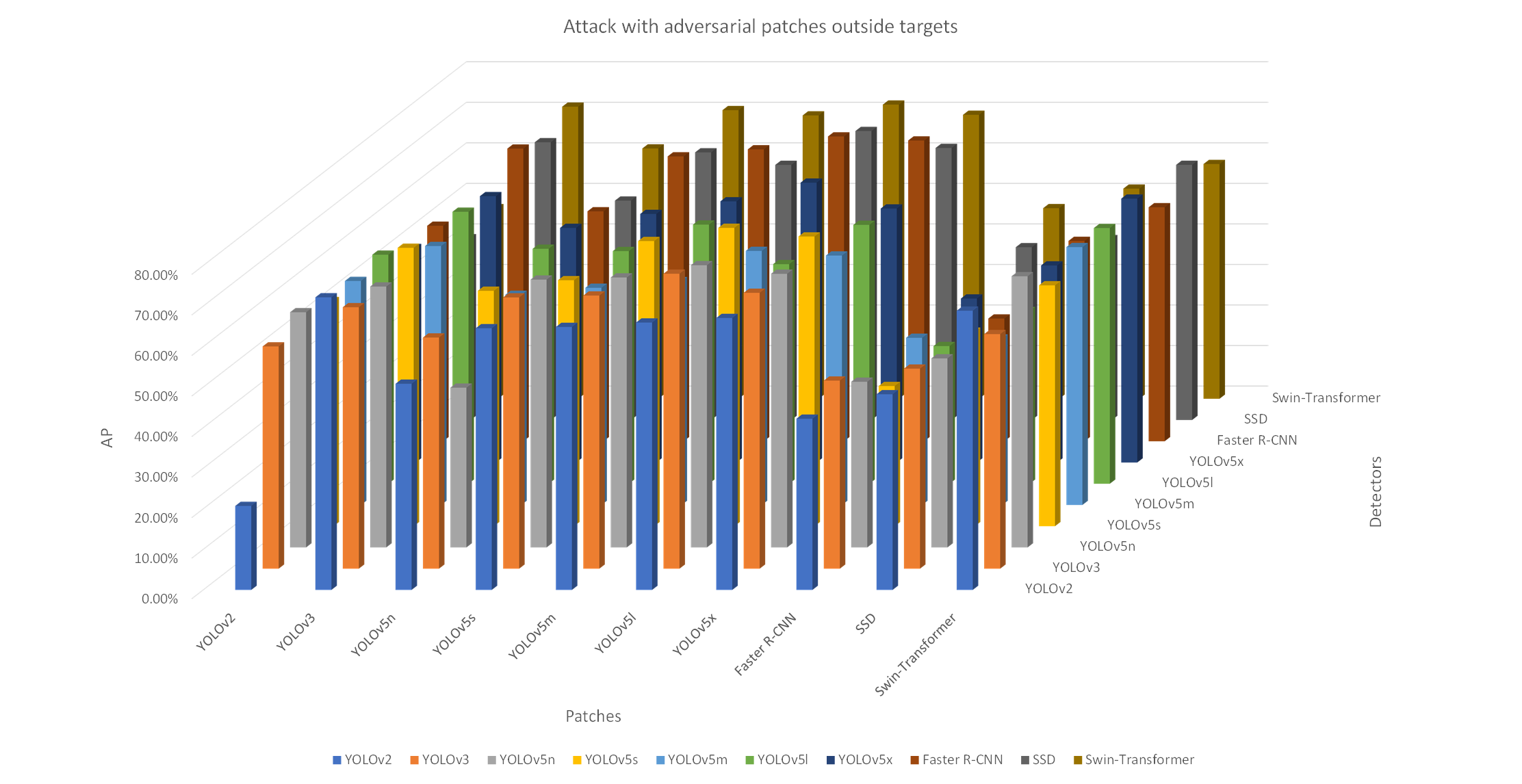}
\end{center}
\caption{The transferability of adversarial attack between different aerial detectors with patches outside targets.}
\label{fig_3d_bar_outside}
\end{figure*}

\begin{figure*}[t!]
\begin{center}
\includegraphics[width=3.5cm]{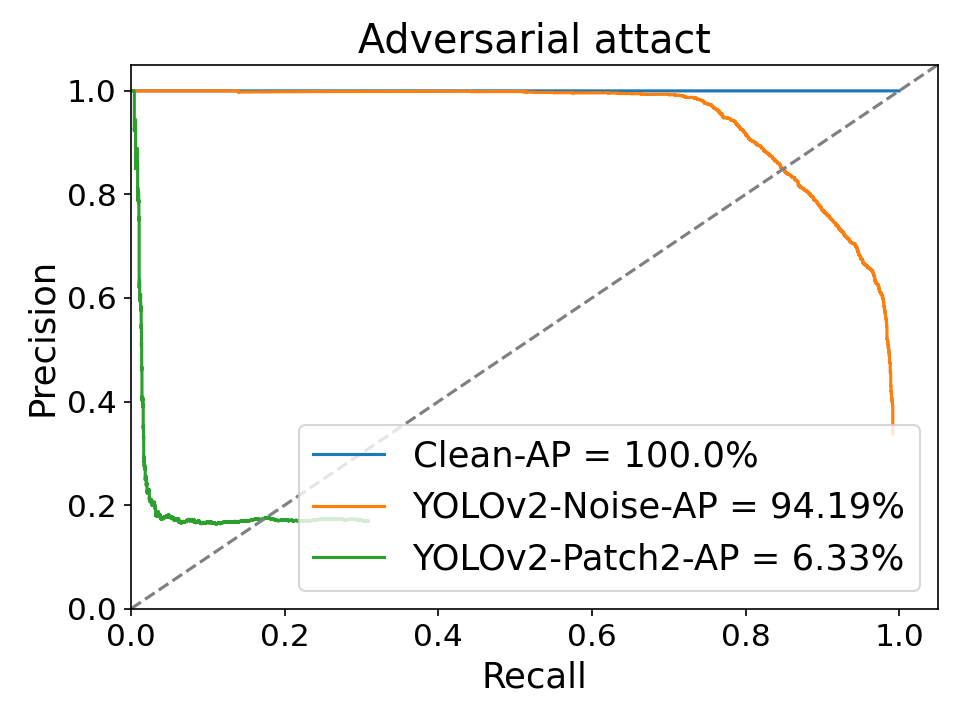}
\includegraphics[width=3.5cm]{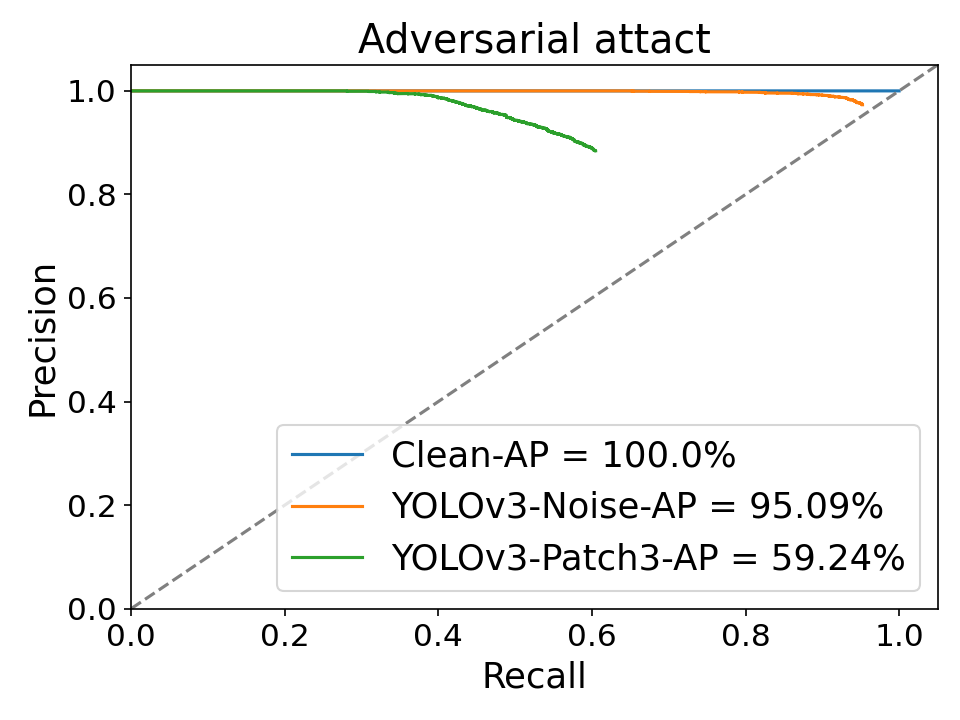}
\includegraphics[width=3.5cm]{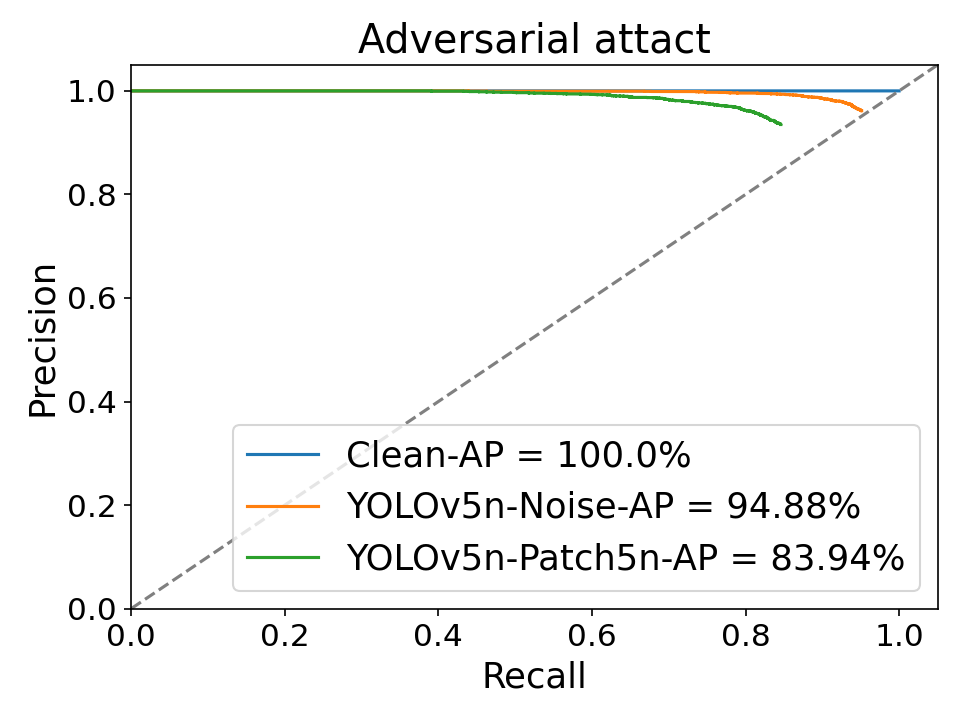}
\includegraphics[width=3.5cm]{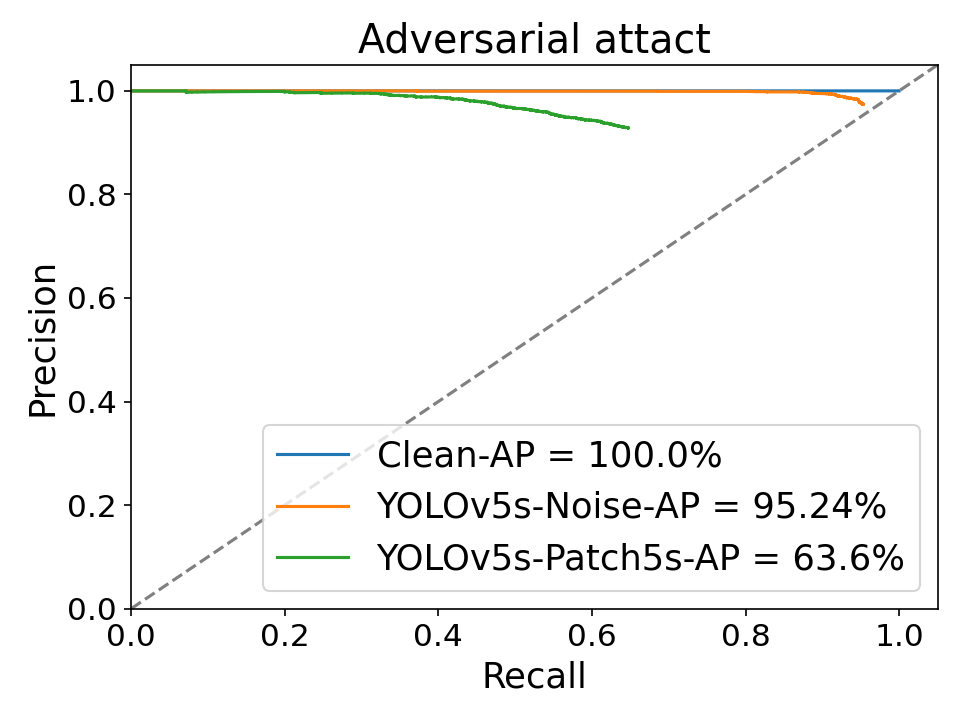}
\includegraphics[width=3.5cm]{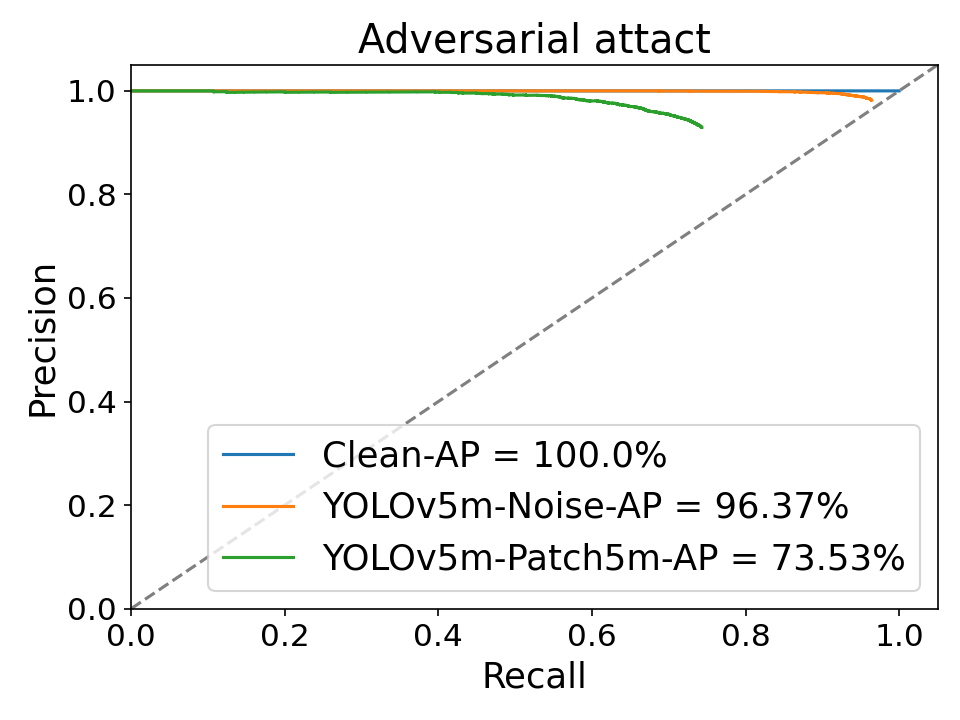}
\includegraphics[width=3.5cm]{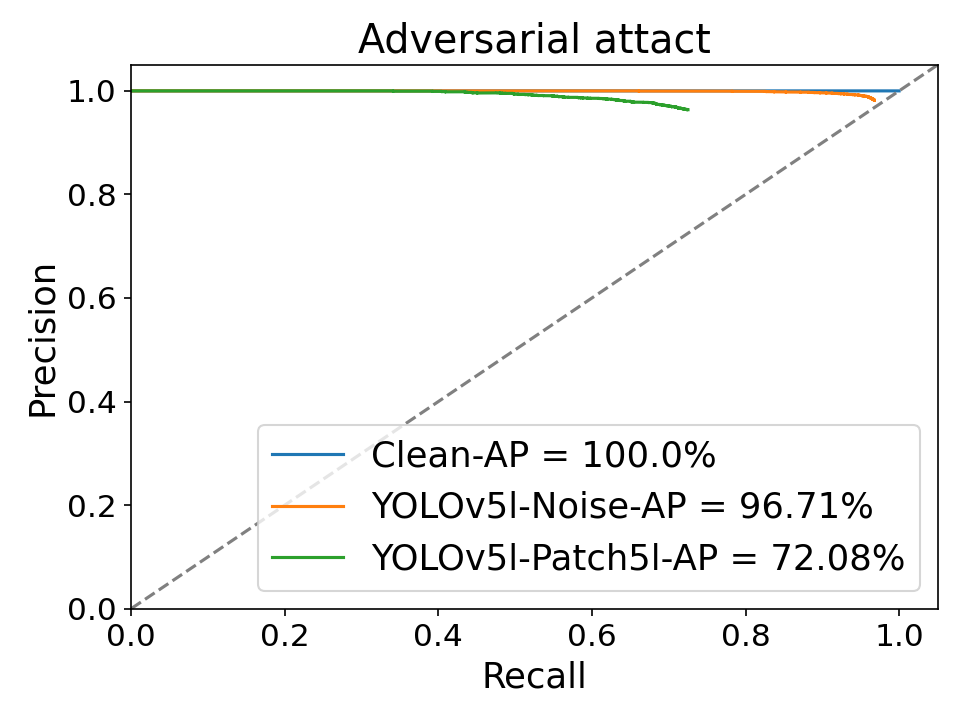}
\includegraphics[width=3.5cm]{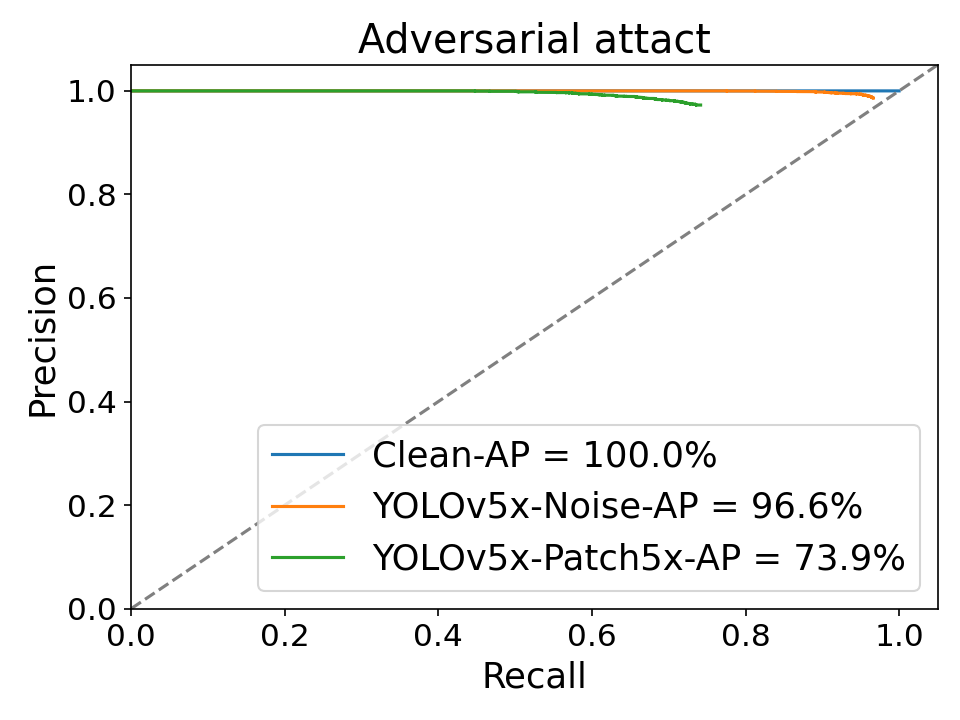}
\includegraphics[width=3.5cm]{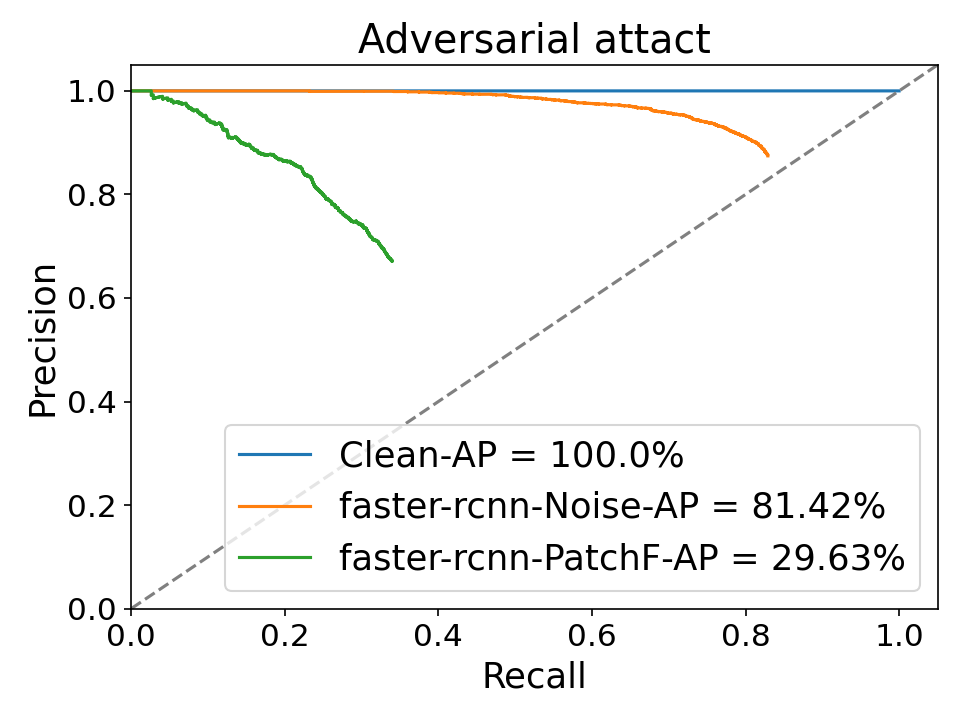}
\includegraphics[width=3.5cm]{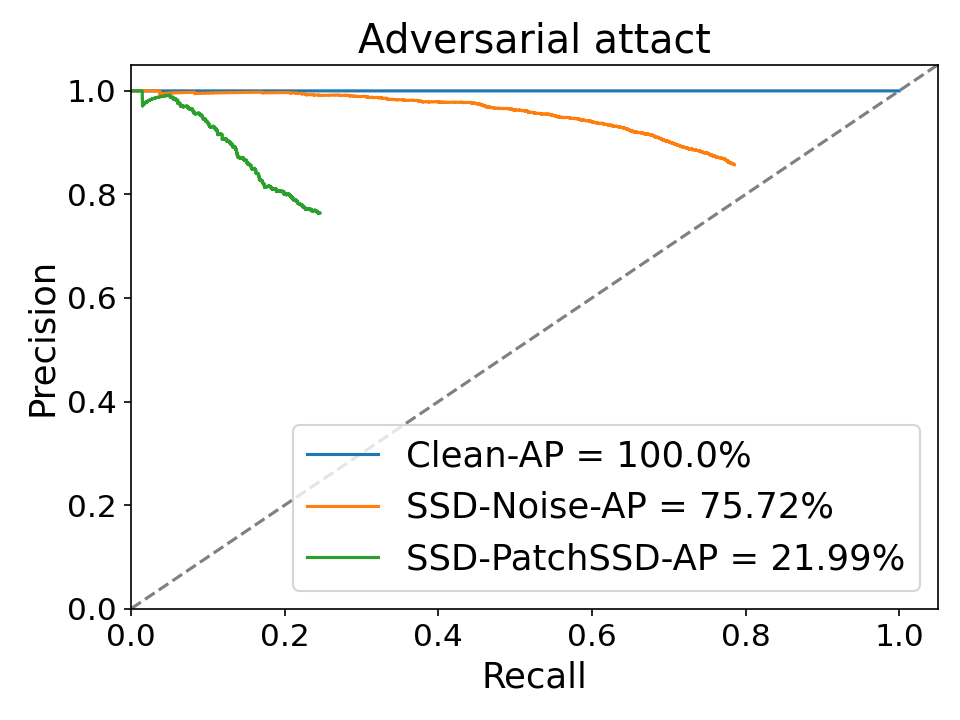}
\includegraphics[width=3.5cm]{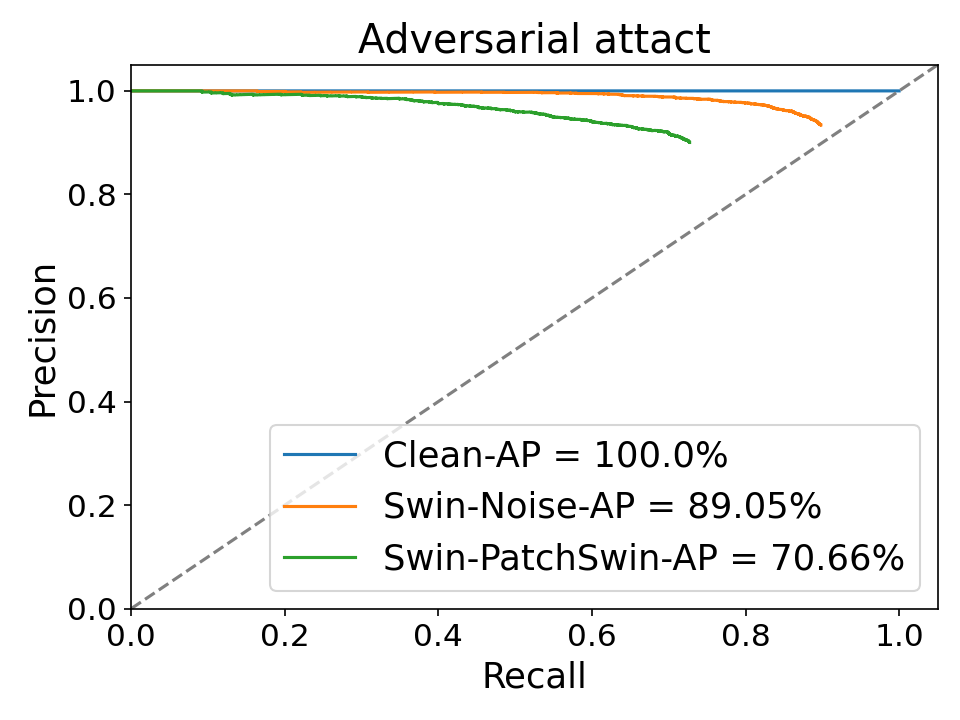}
\end{center}
\caption{P-R curves of adversarial attack against different aerial detectors  with patches on targets.}
\label{fig_pr_curve_center}
\end{figure*}

\begin{figure*}[t!]
\begin{center}
\includegraphics[width=3.5cm]{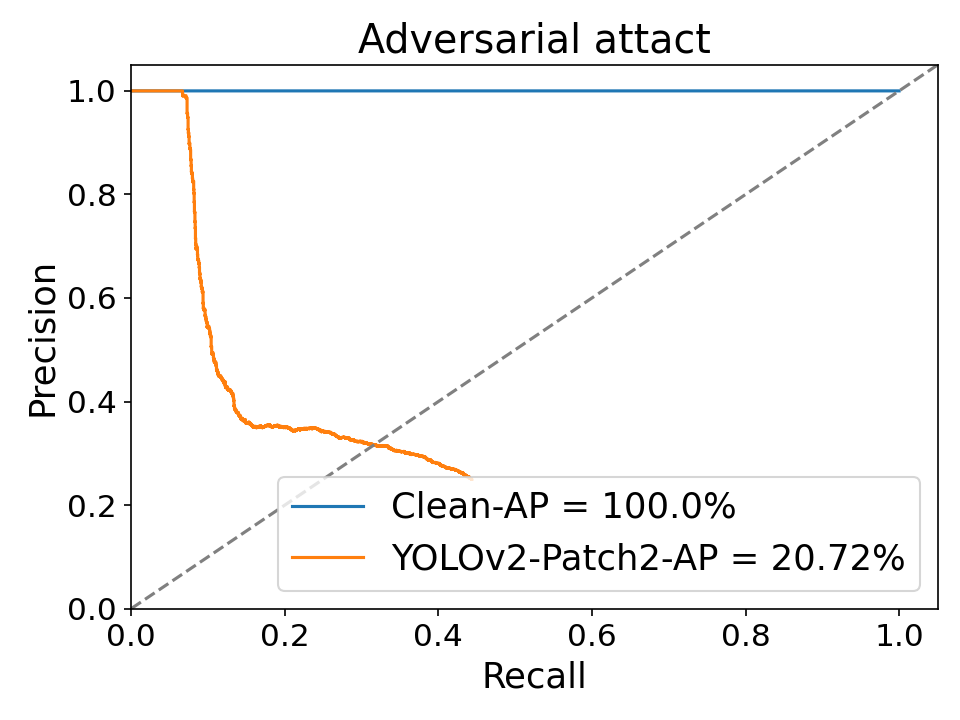}
\includegraphics[width=3.5cm]{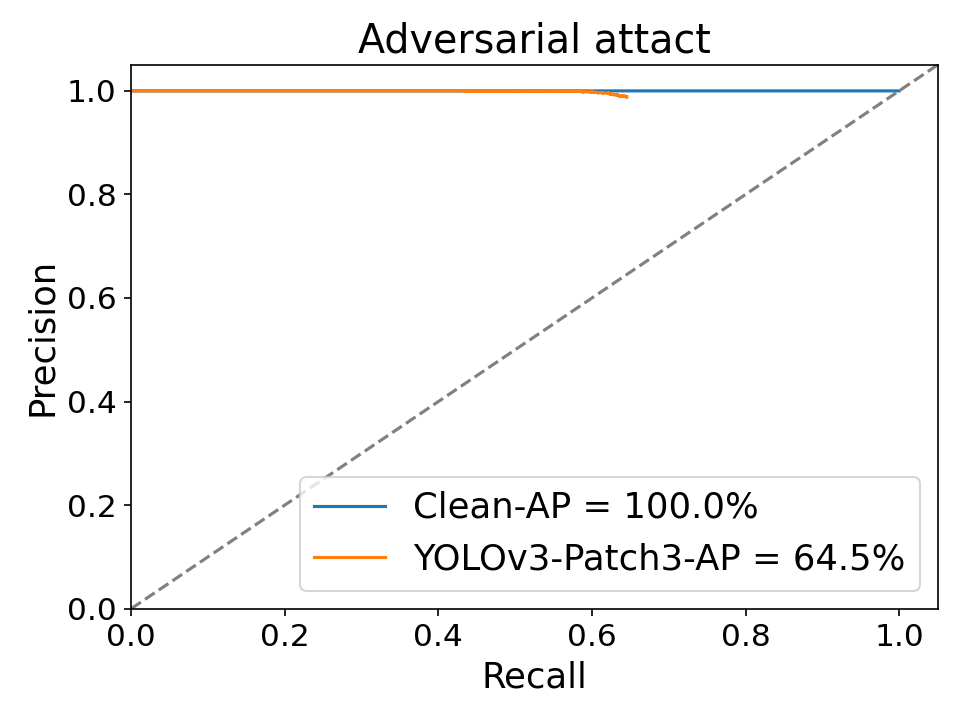}
\includegraphics[width=3.5cm]{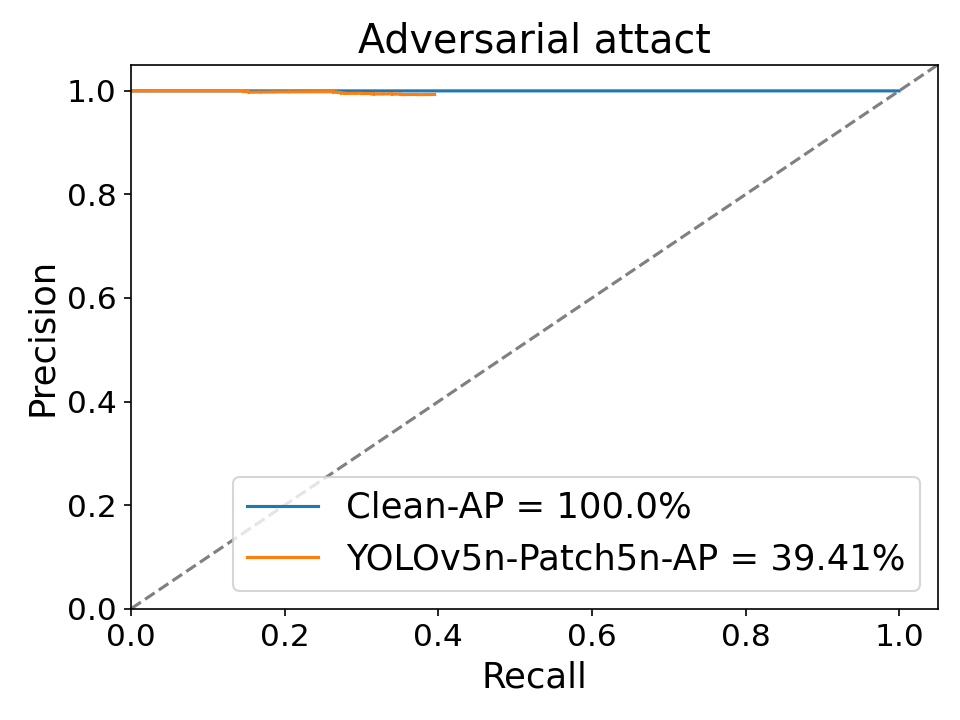}
\includegraphics[width=3.5cm]{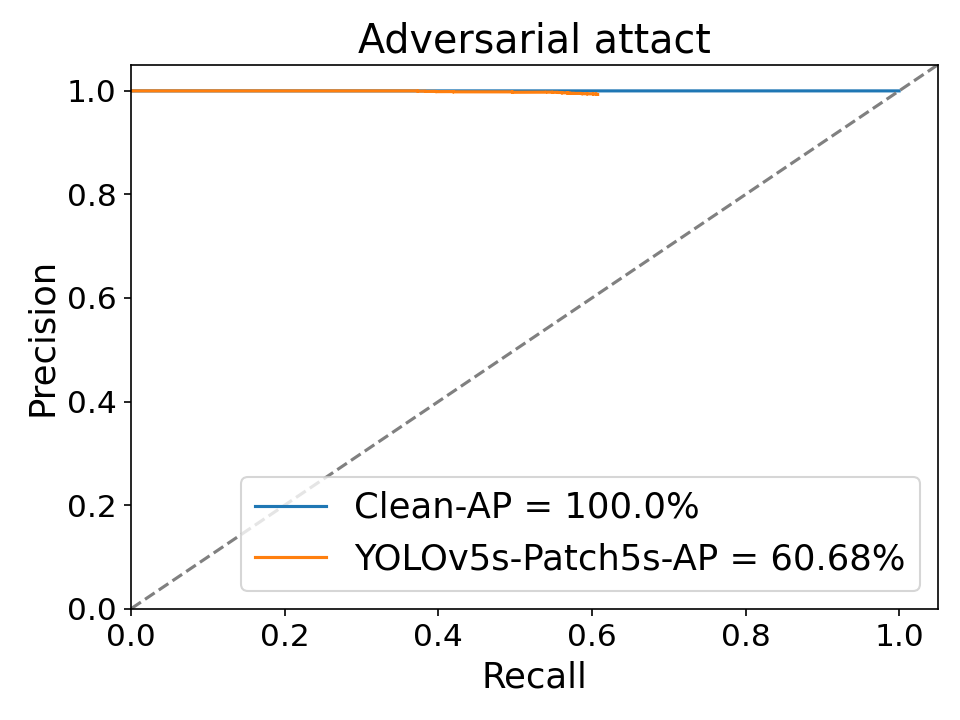}
\includegraphics[width=3.5cm]{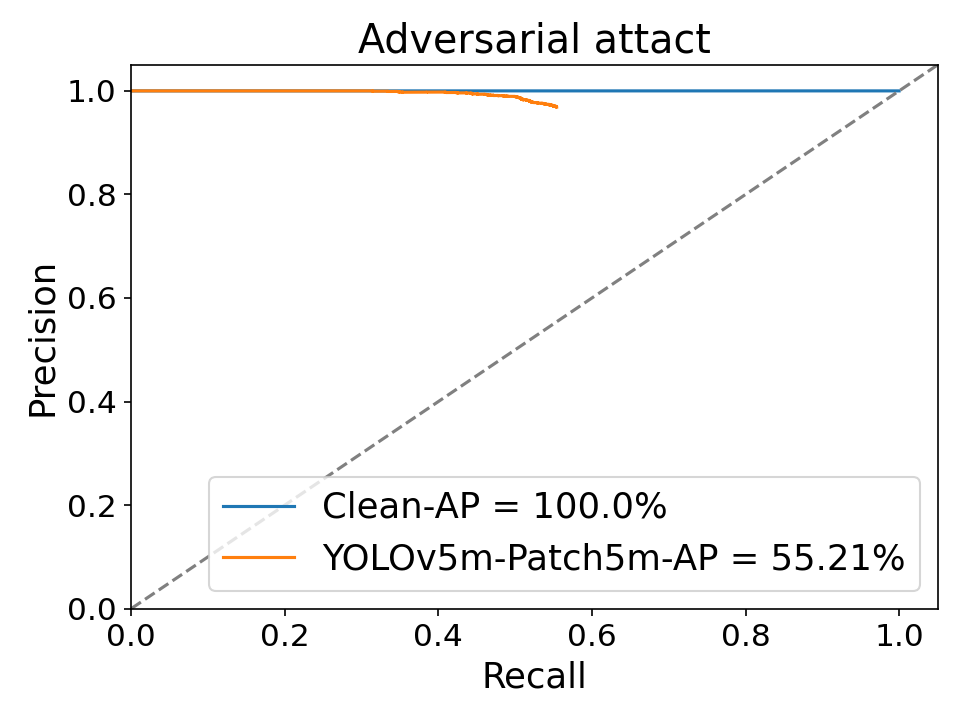}
\includegraphics[width=3.5cm]{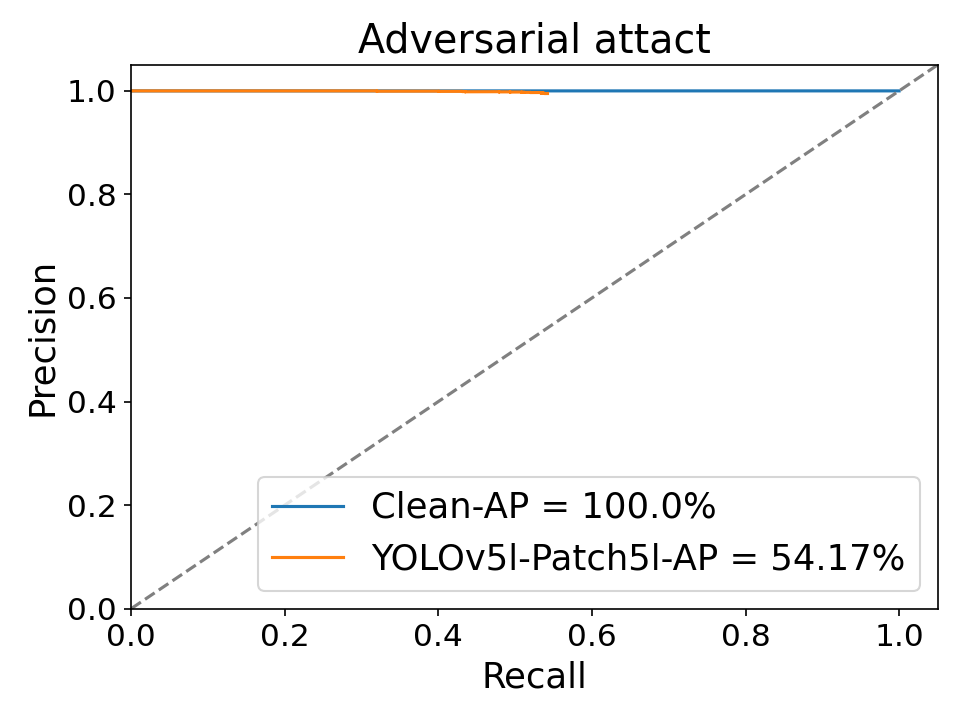}
\includegraphics[width=3.5cm]{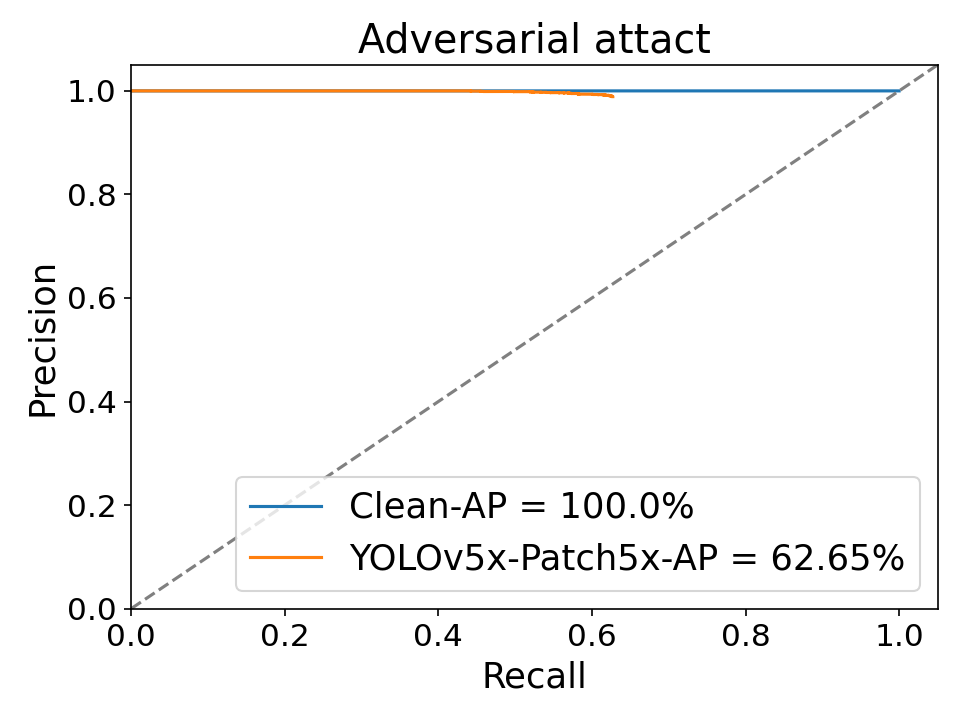}
\includegraphics[width=3.5cm]{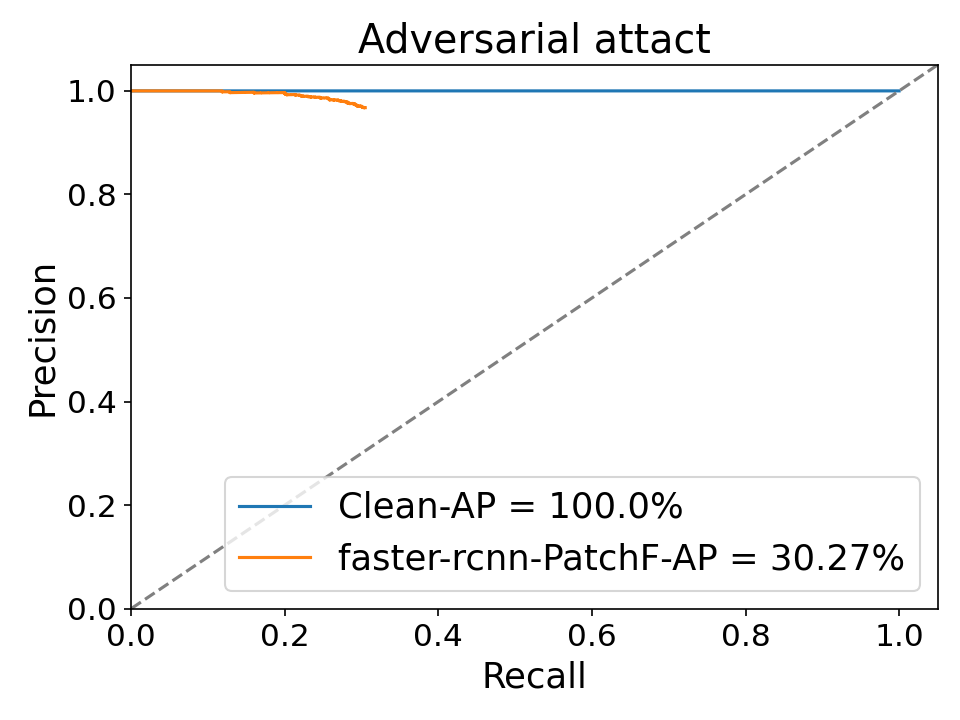}
\includegraphics[width=3.5cm]{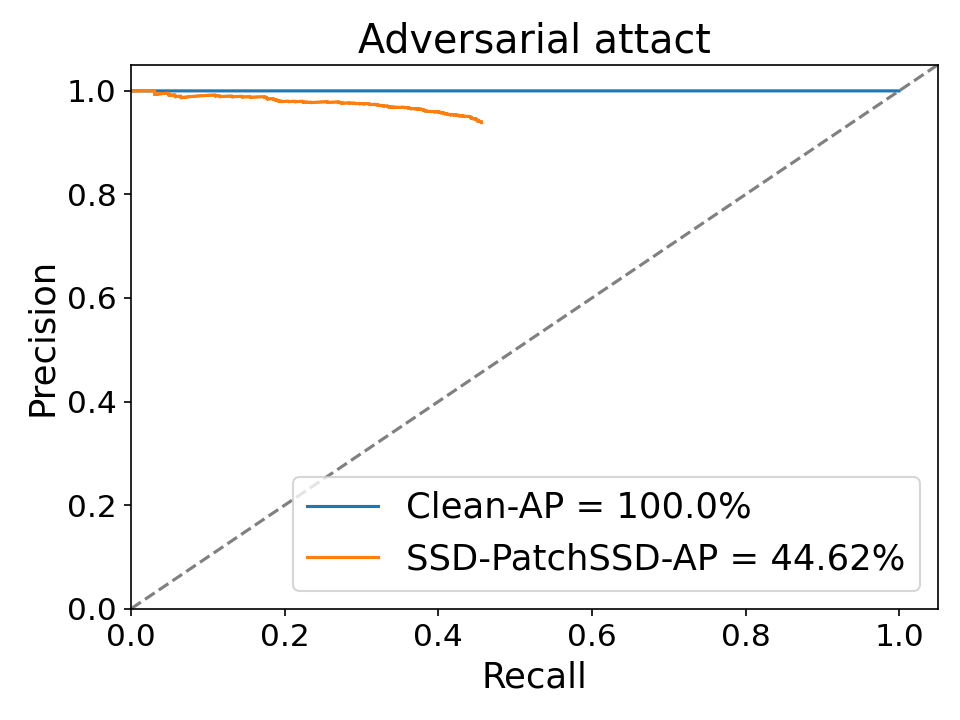}
\includegraphics[width=3.5cm]{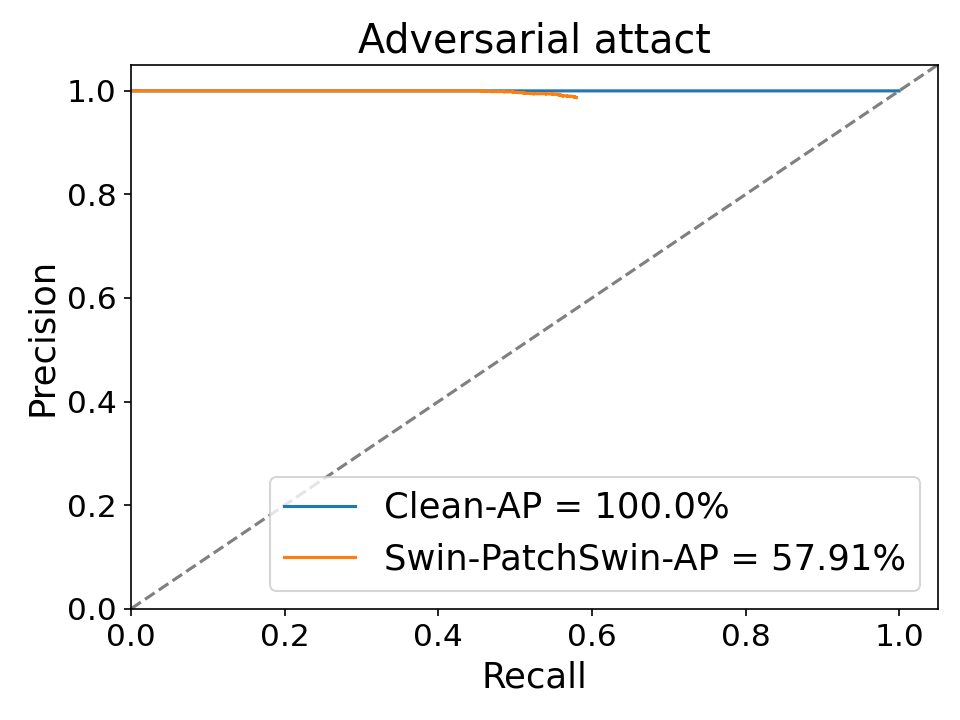}
\end{center}
\caption{P-R curves of adversarial attack against different aerial detectors  with patches outside targets.}
\label{fig_pr_curve_outside}
\end{figure*}

\begin{figure*}[t!]
\begin{center}
\includegraphics[width=3.5cm]{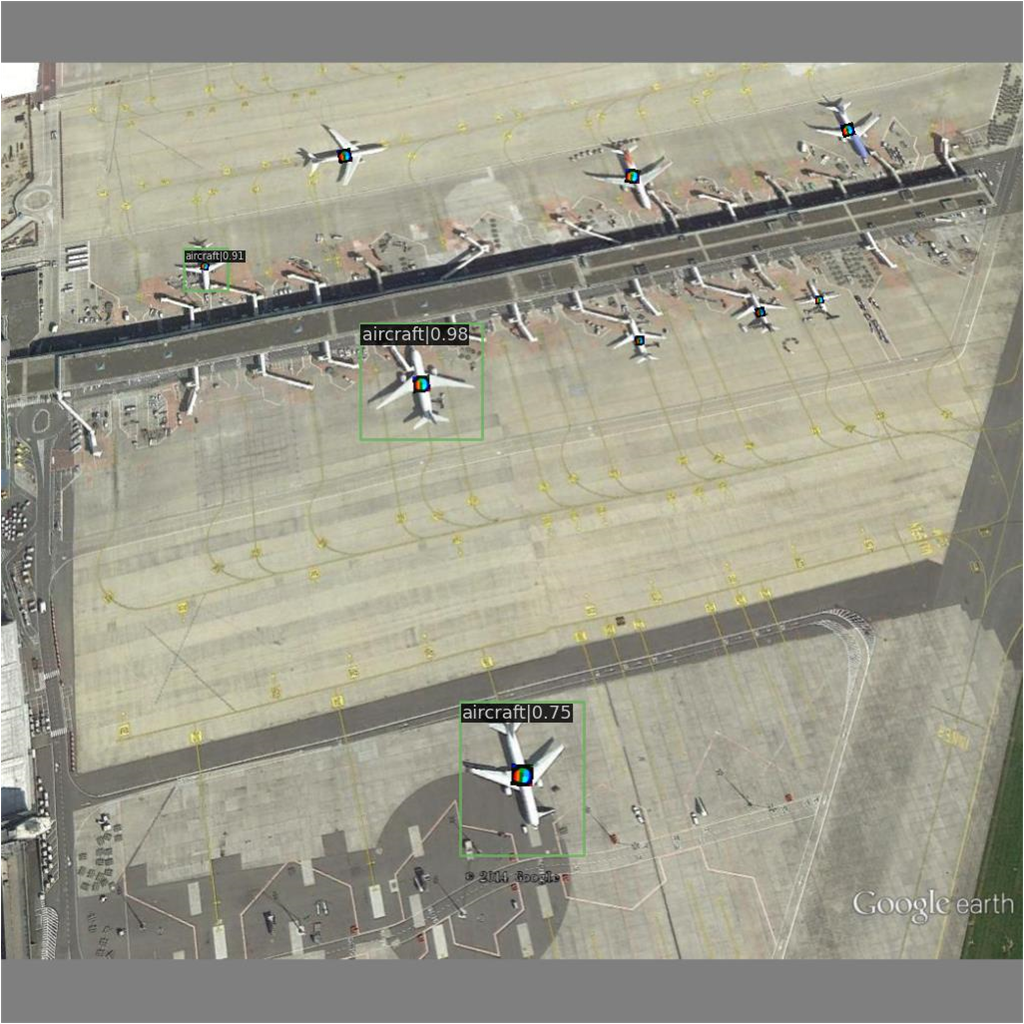}
\includegraphics[width=3.5cm]{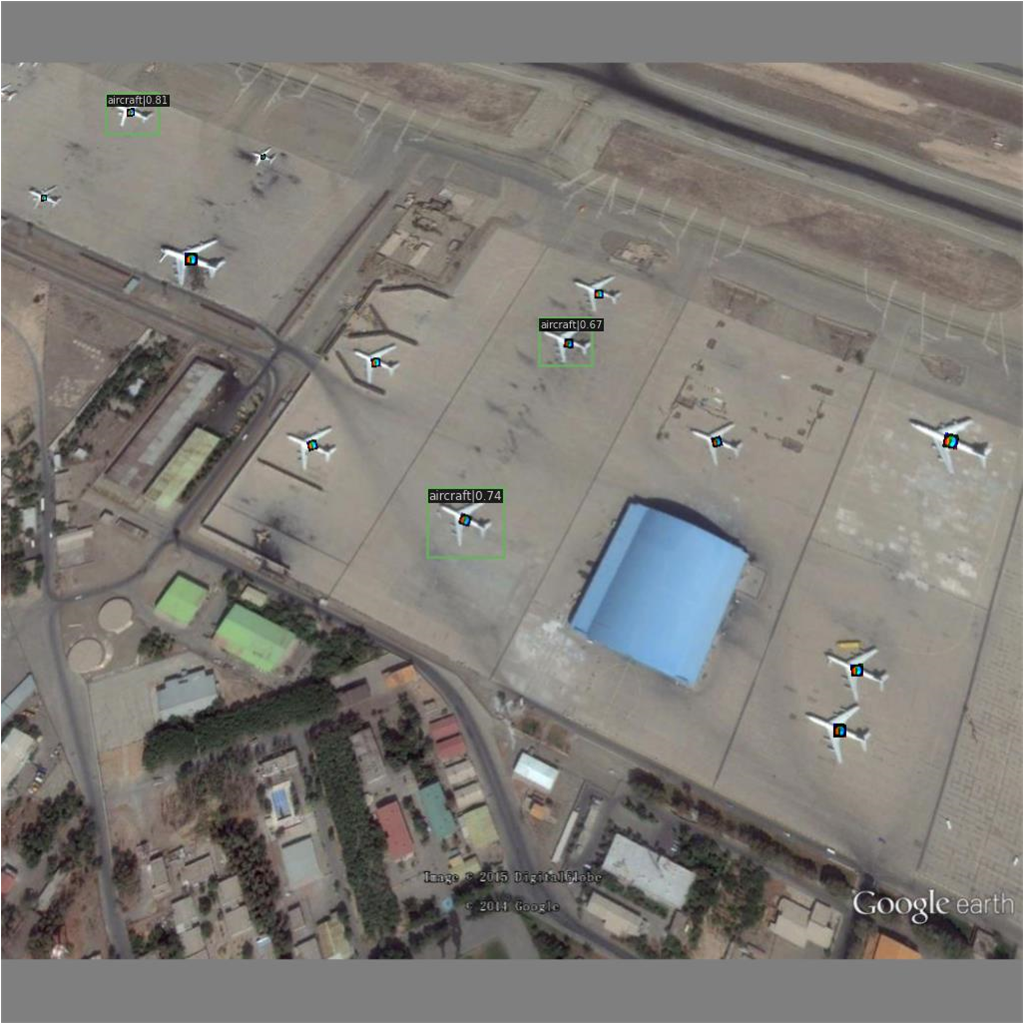}
\includegraphics[width=3.5cm]{aircraft_19_p_center.png}
\includegraphics[width=3.5cm]{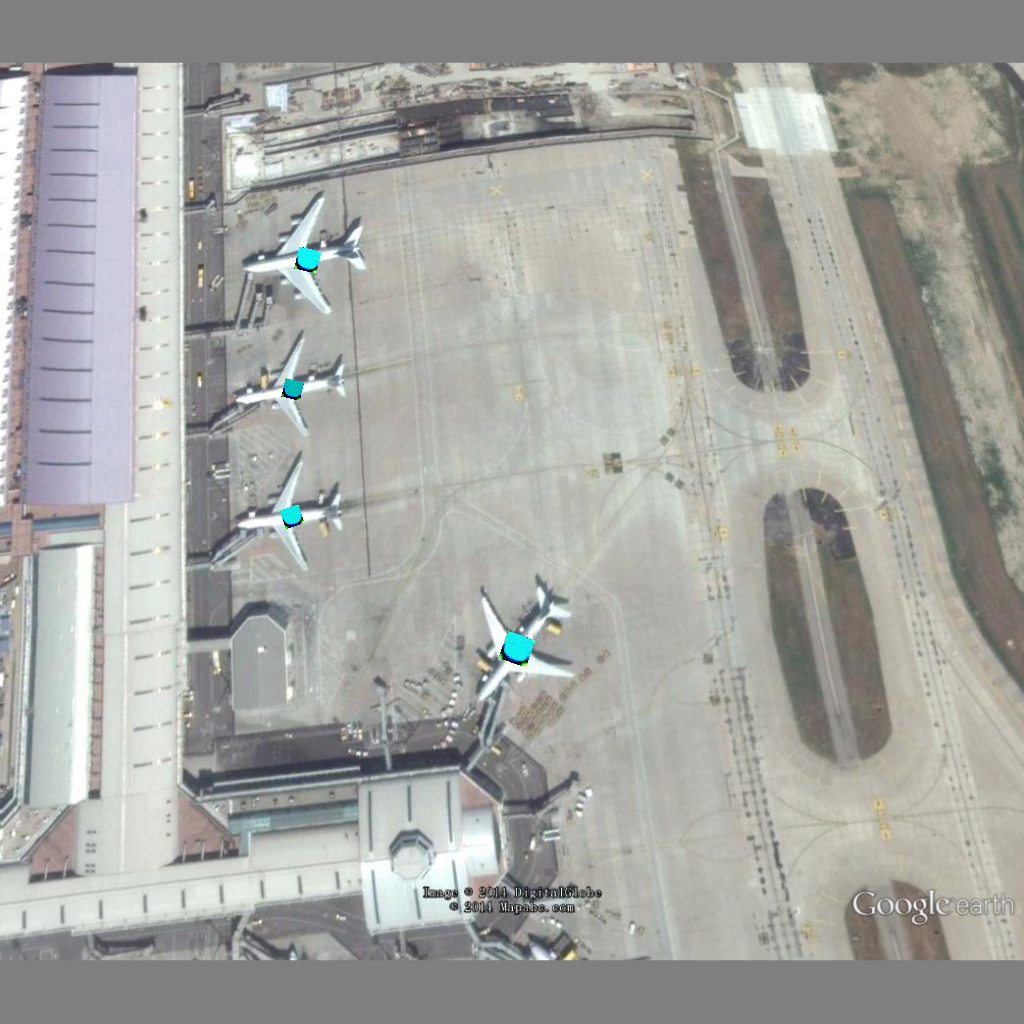}
\includegraphics[width=3.5cm]{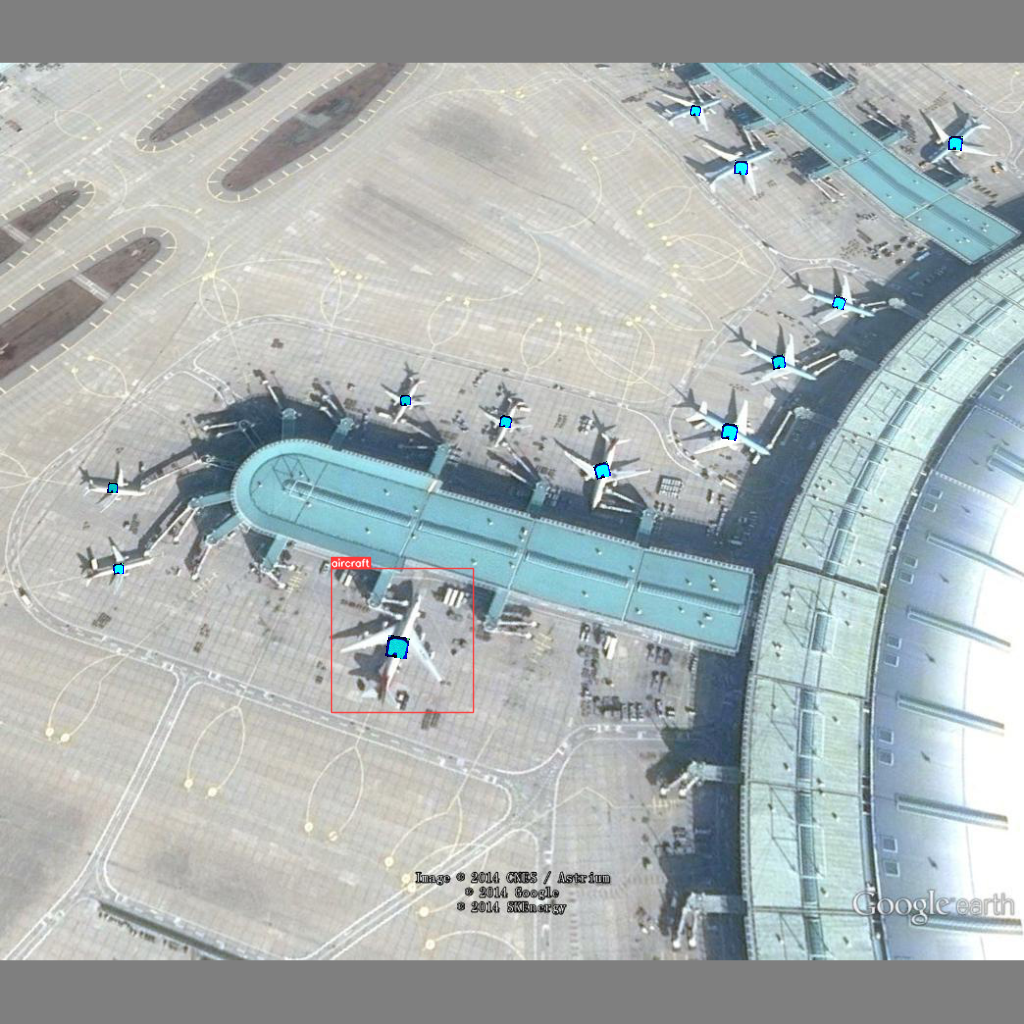}
\end{center}
\begin{center}
\includegraphics[width=3.5cm]{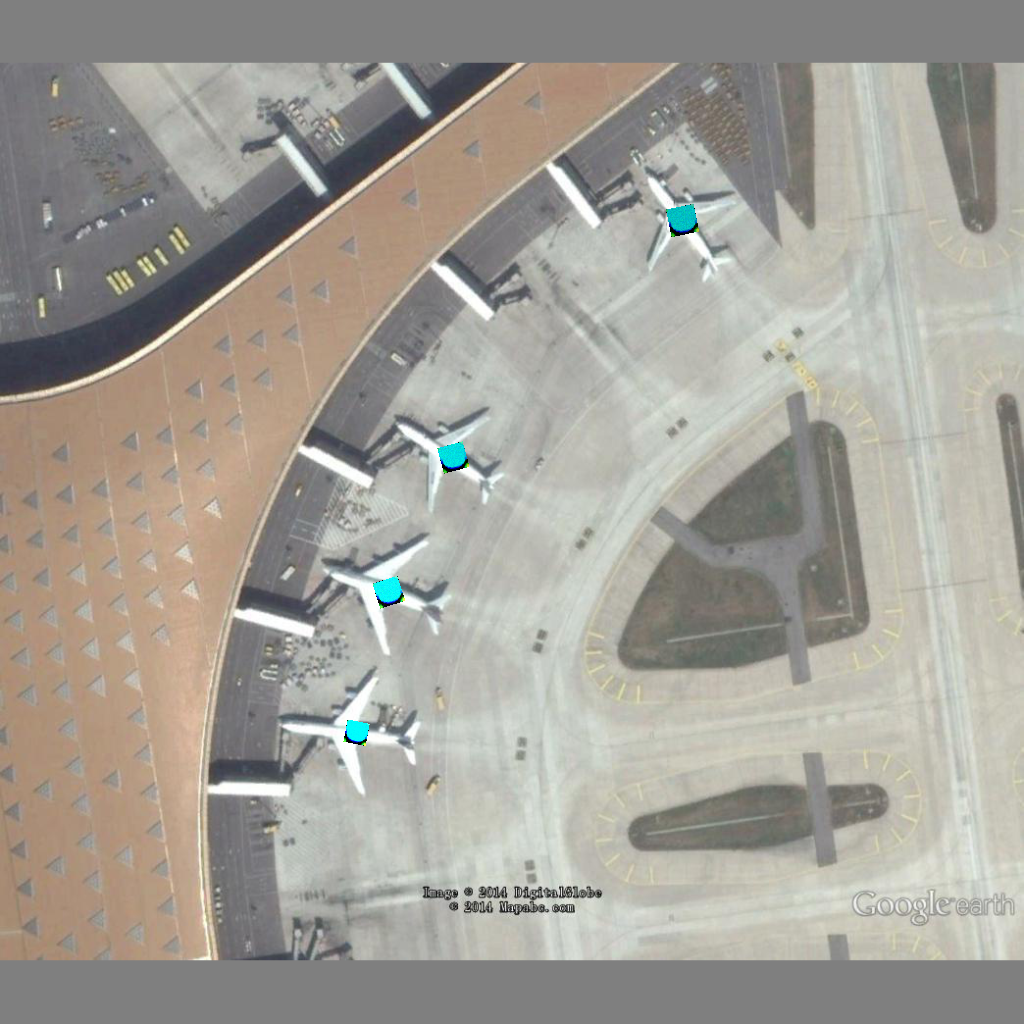}
\includegraphics[width=3.5cm]{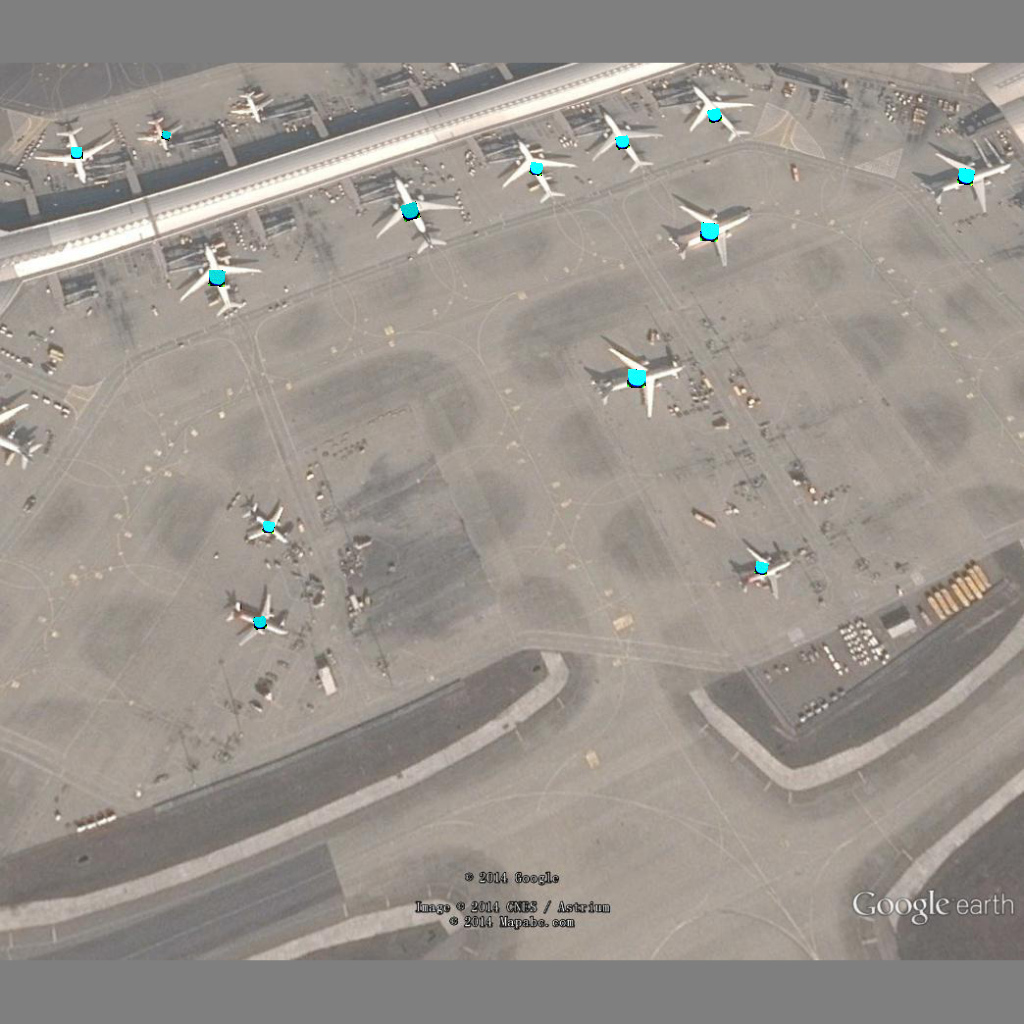}
\includegraphics[width=3.5cm]{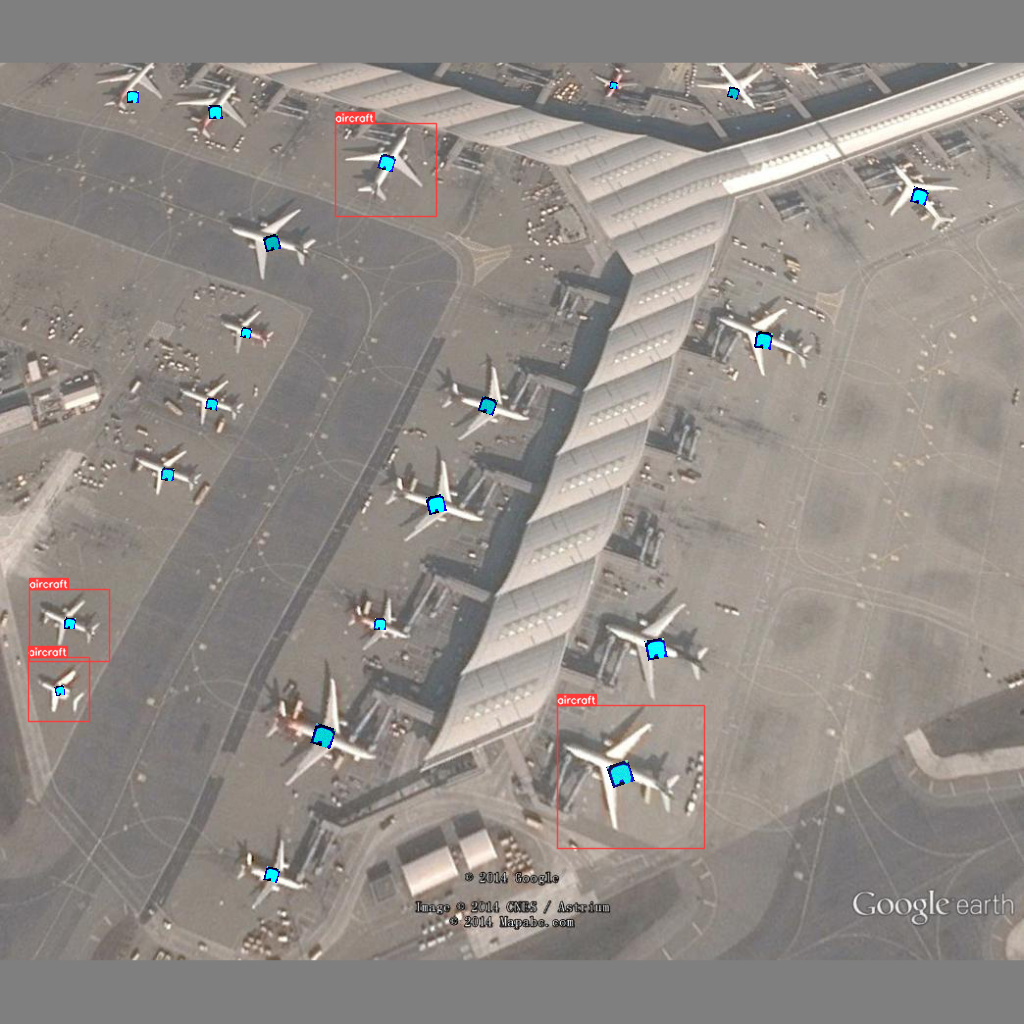}
\includegraphics[width=3.5cm]{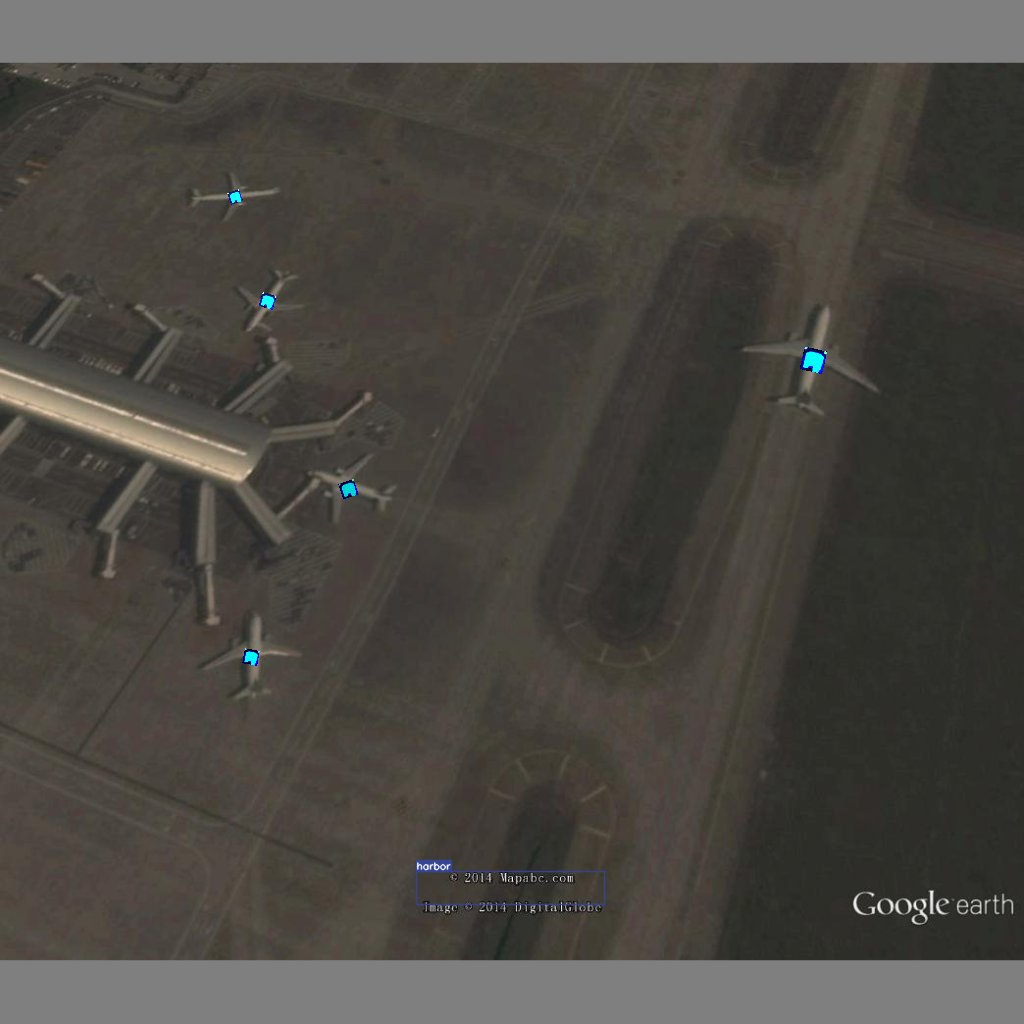}
\includegraphics[width=3.5cm]{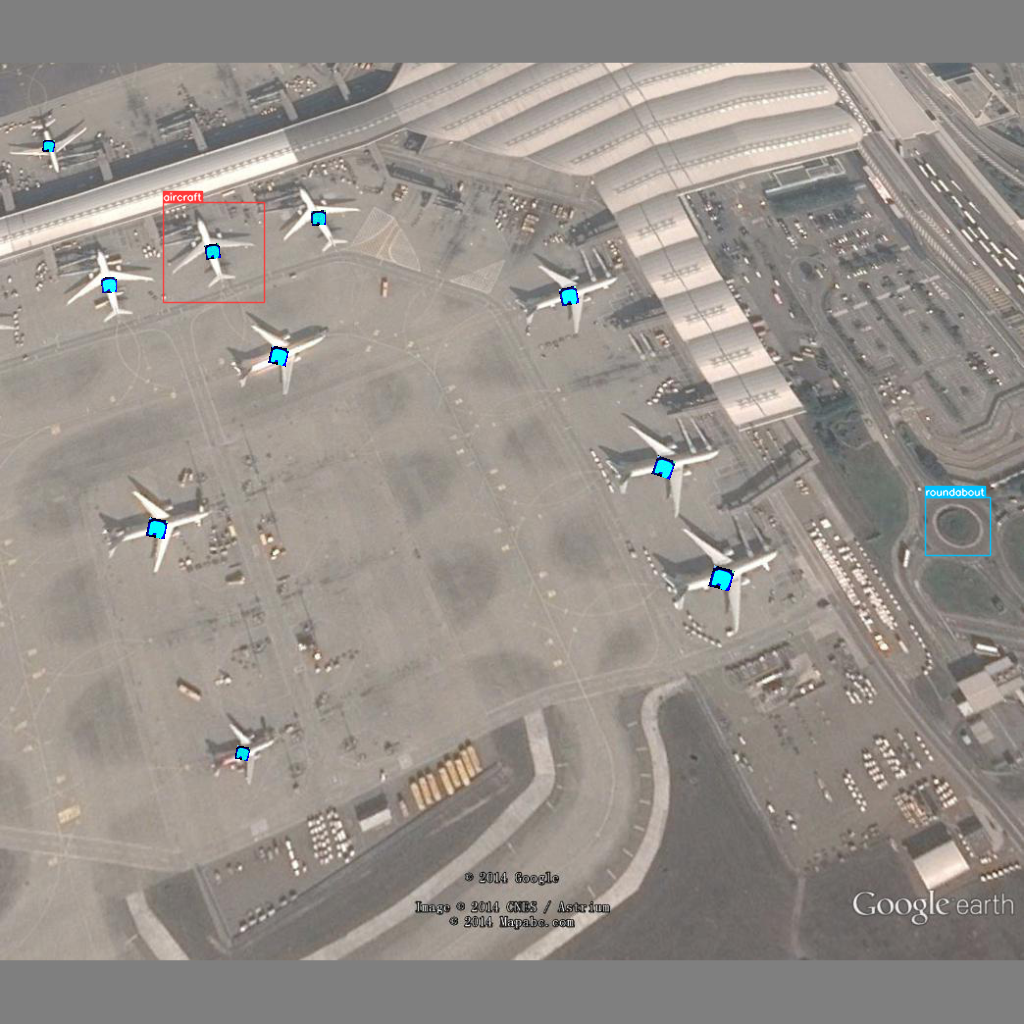}
\end{center}
\caption{Visual examples of attack with adversarial patches on targets.}
\label{fig_visual_examples_on}
\end{figure*}

\begin{figure*}[t!]
\begin{center}
\includegraphics[width=3.5cm]{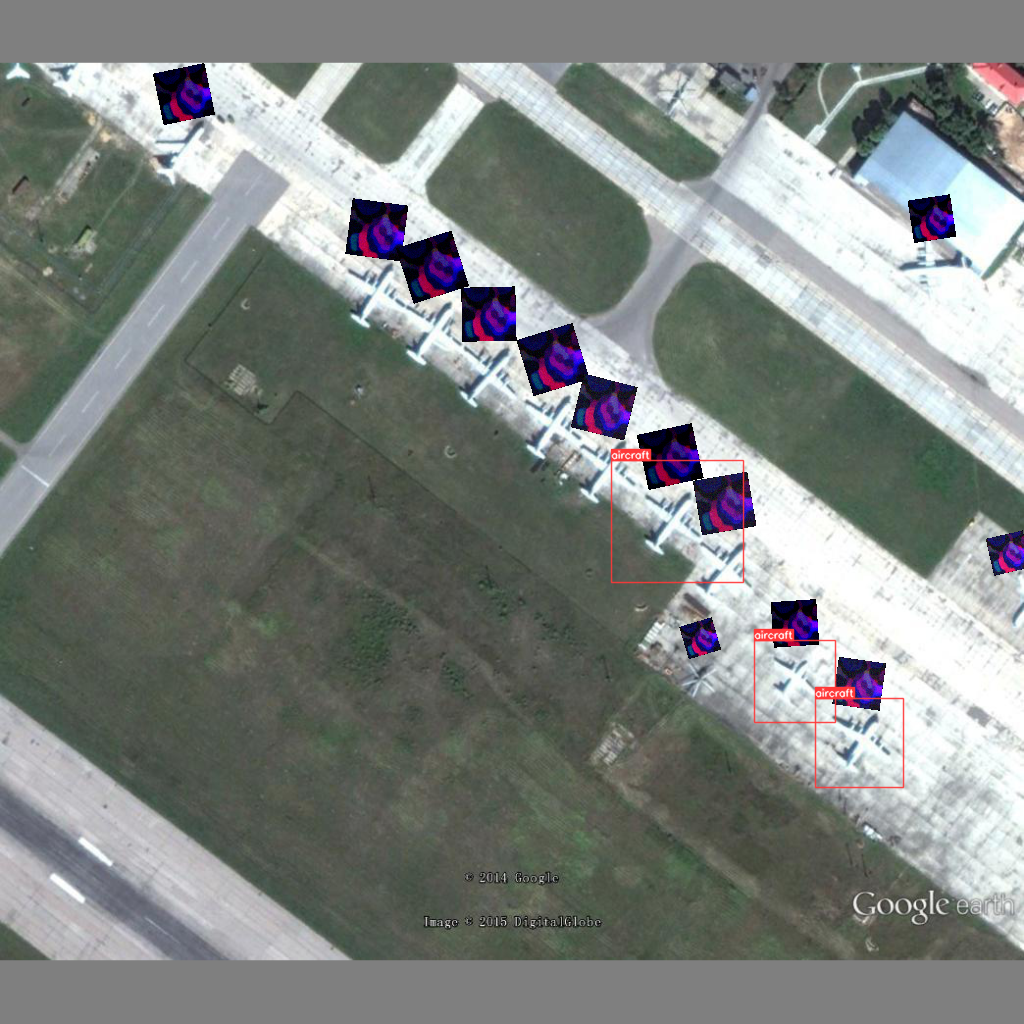}
\includegraphics[width=3.5cm]{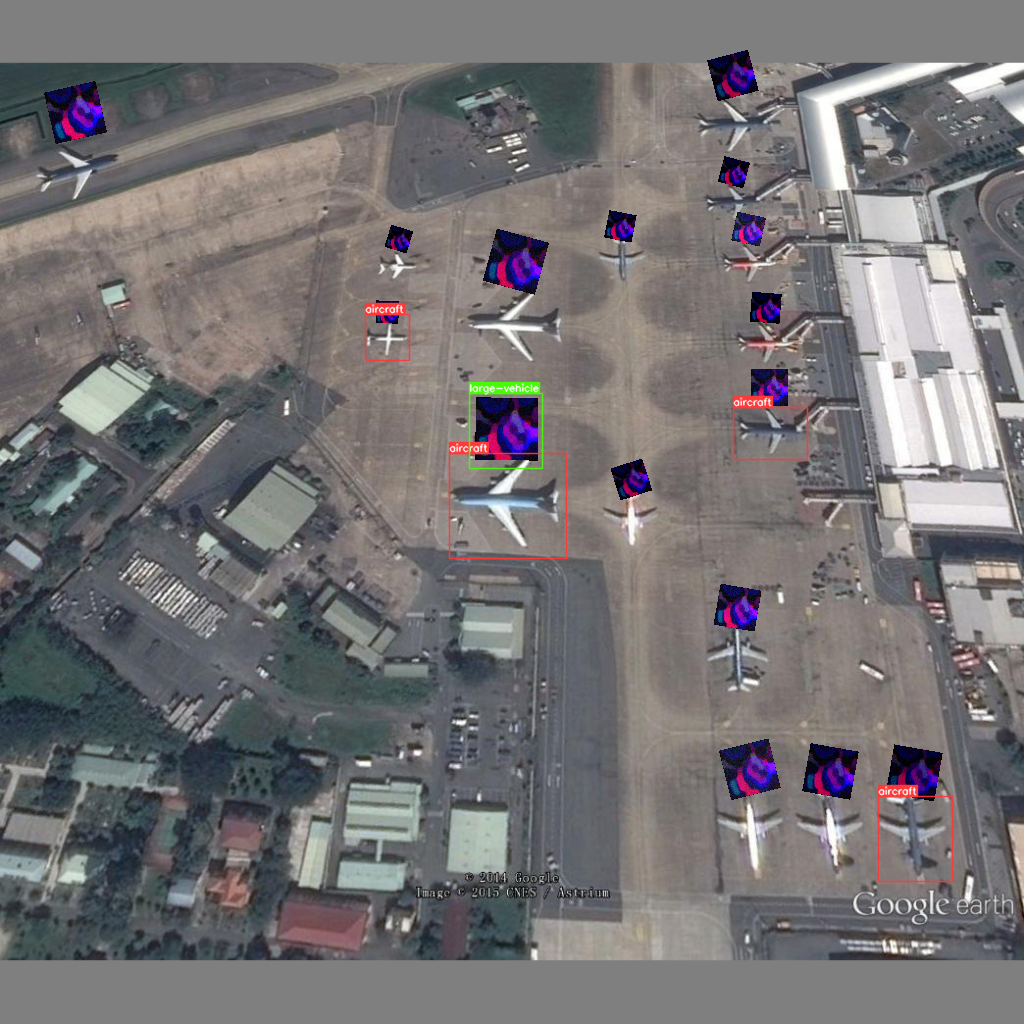}
\includegraphics[width=3.5cm]{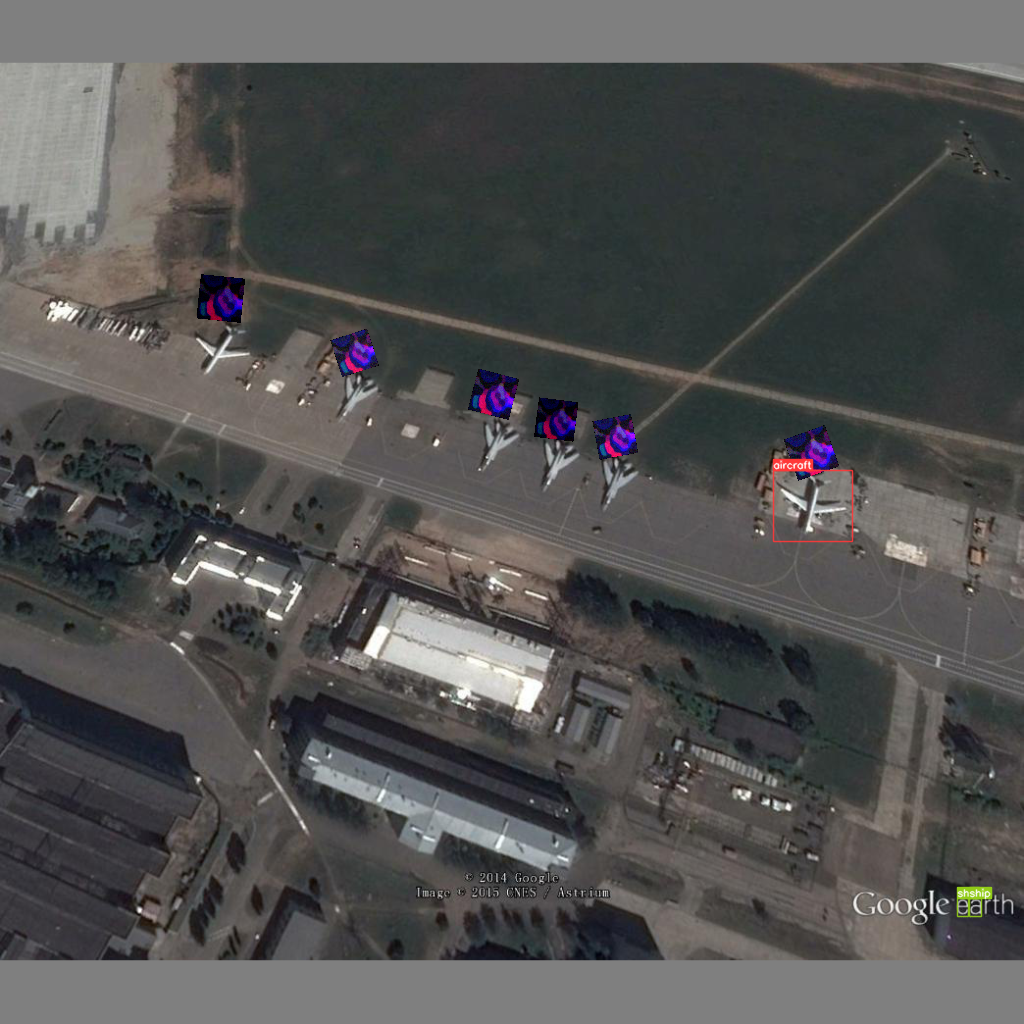}
\includegraphics[width=3.5cm]{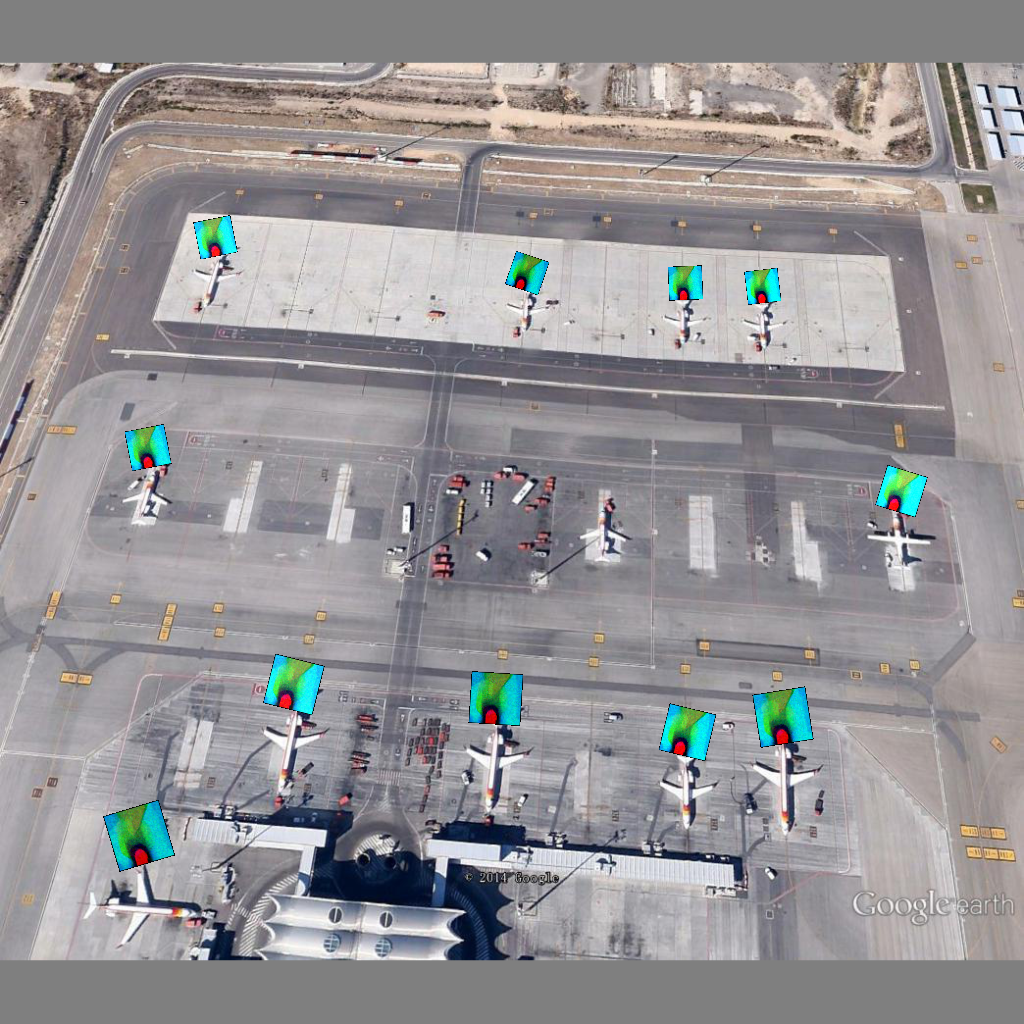}
\includegraphics[width=3.5cm]{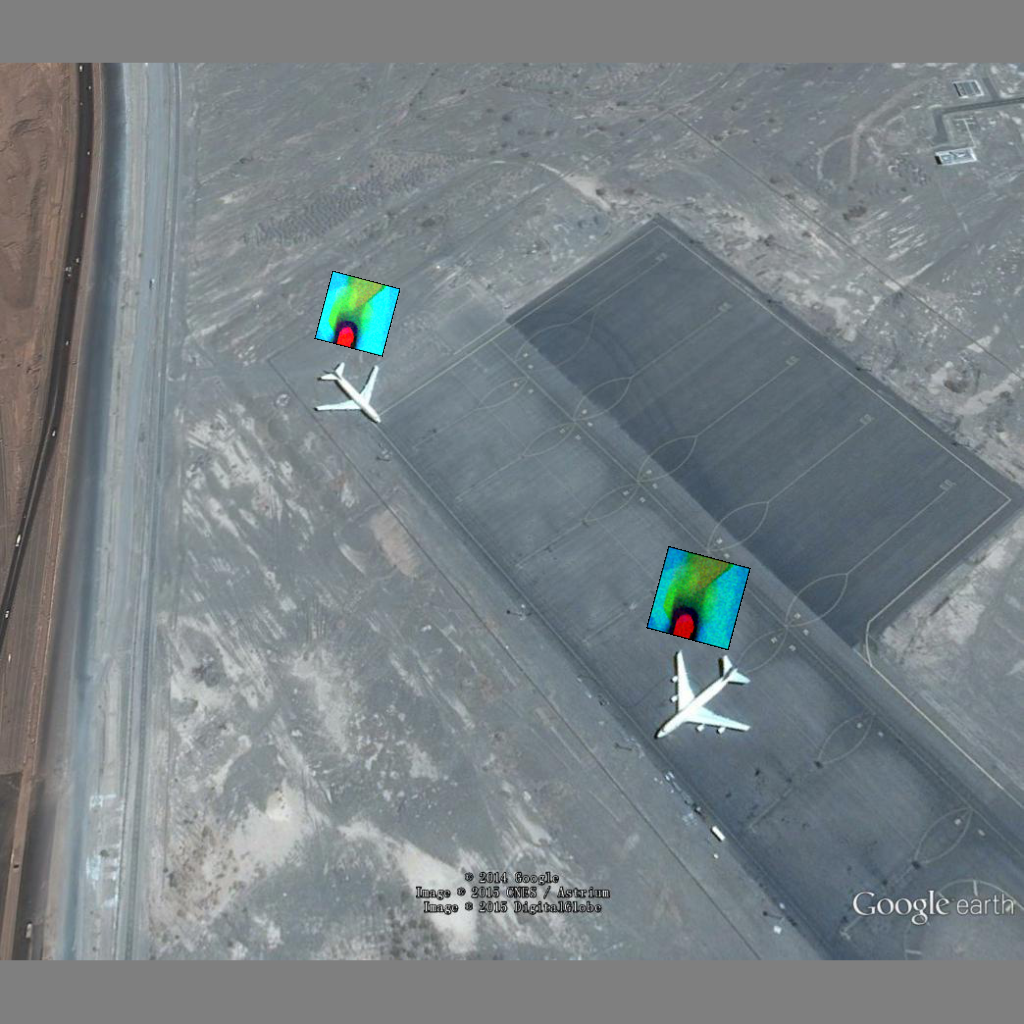}
\end{center}
\begin{center}
\includegraphics[width=3.5cm]{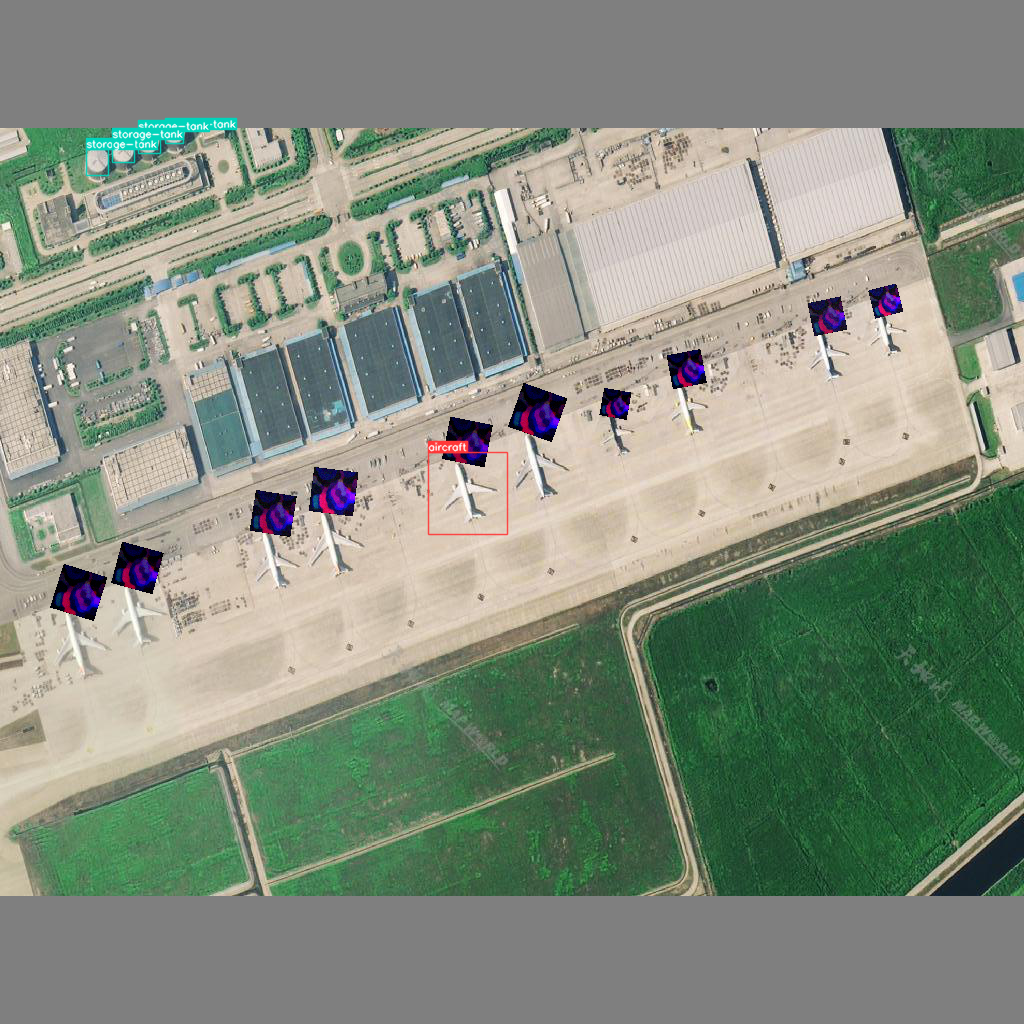}
\includegraphics[width=3.5cm]{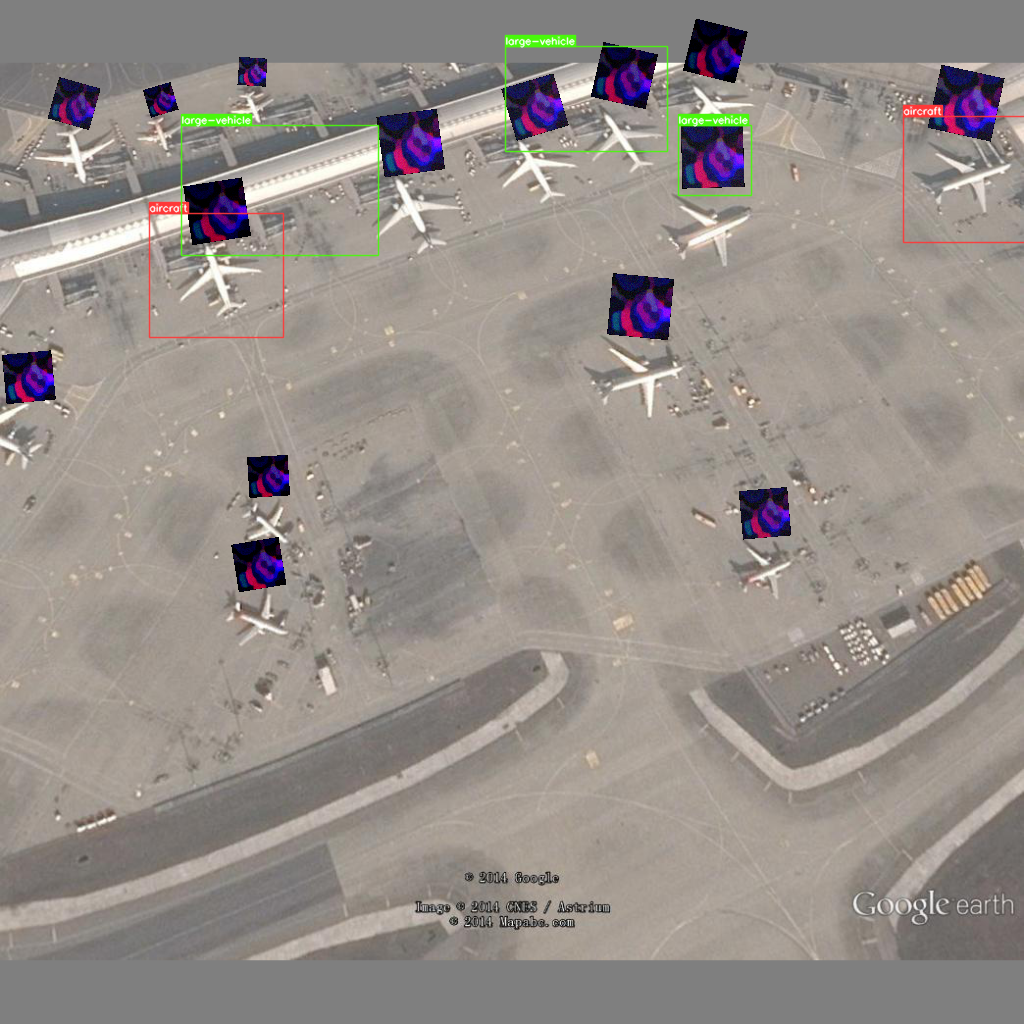}
\includegraphics[width=3.5cm]{aircraft_407_p_upper.png}
\includegraphics[width=3.5cm]{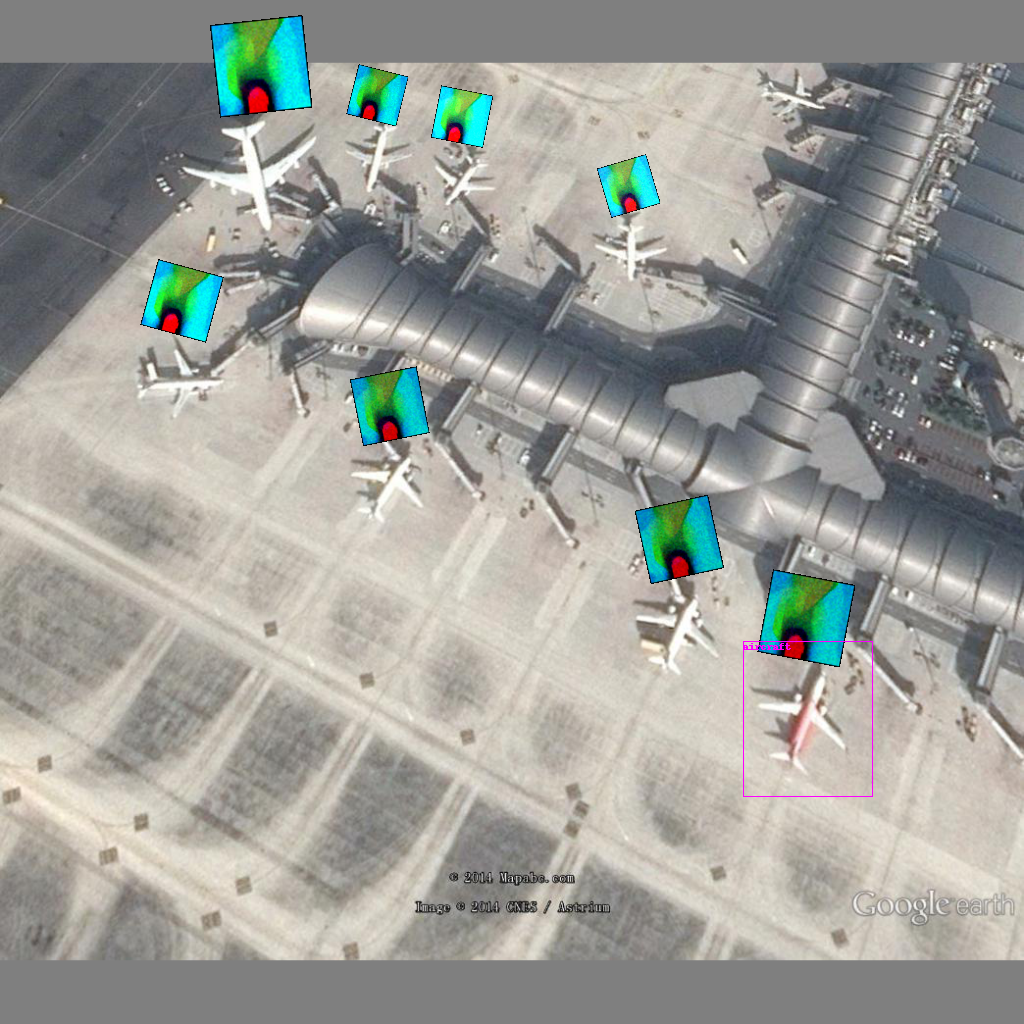}
\includegraphics[width=3.5cm]{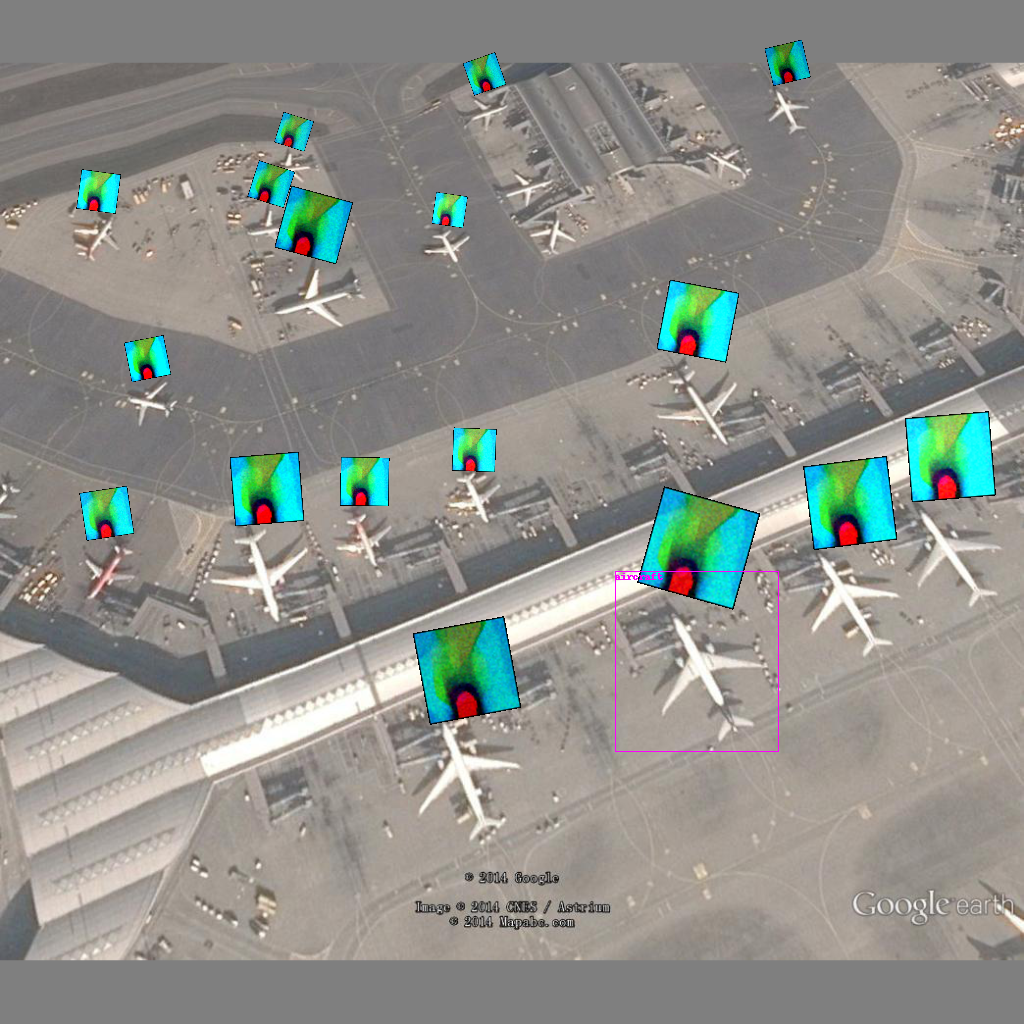}
\end{center}
\caption{Visual examples of attack with adversarial patches outside targets.}
\label{fig_visual_examples_outside}
\end{figure*}

\begin{table*}[t!]
\caption{Comparison of experimental results in the digital domain.}
\label{table_results_comparison_digital}
\centering
\setlength{\tabcolsep}{0.7mm}
\begin{threeparttable}
\begin{tabular*}{\hsize}{cccccccccccc}
\hline\hline
\diaghead{\theadfont Diag ColumnmnHead}{Patches}{Detectors} &--- &YOLOv2 &YOLOv3 &YOLOv5n    & YOLOv5s   &YOLOv5m    &YOLOv5l    &YOLOv5x    &Faster R-CNN   &SSD    &Swin Transformer
\\ \hline\hline
                 &Thys \etal  &9.64\%     &69.35\%    &83.11\%    &72.54\%    &82.09\%    &80.59\%    &80.08\%    &71.74\%    &51.47\%    &86.72\%    \\
YOLOv2\cite{redmon2017yolo9000}           &Ours(On)  &\textbf{6.33}\%     &\textbf{65.80}\%    &80.18\%	&73.05\%	&80.77\%	&78.83\%	&78.44\%	&68.88\%	&\textbf{47.80}\%	&85.98\%    \\
                 &Ours(Out) &20.72\%	&72.18\%    &5\textbf{0.86}\%	&\textbf{64.55}\%	&\textbf{64.87}\%	&\textbf{65.96}\%	&\textbf{67.08}\%	&\textbf{42.23}\%	&48.27\%	&\textbf{68.88}\%    \\
\hline
                 &Thys \etal  &\textbf{6.87}\%     &63.06\%    &80.29\%    &71.53\%    &79.60\%    &76.82\%    &78.41\%    &64.41\%    &46.10\%    &86.25\%    \\
YOLOv\cite{redmon2018yolov3}           &Ours(On)  &19.38\%	&\textbf{59.24}\%    &75.36\%	&\textbf{66.43}\%	&74.20\%	&\textbf{72.54}\%	&75.40\%	&\textbf{35.35}\%	&\textbf{28.88}\%	&82.78\%    \\
                 &Ours(Out) &54.83\%	&64.50\%	&\textbf{57.03}\%	&66.91\%	&\textbf{67.38}\%	&72.76\%	&\textbf{68.05}\%	&46.40\%	&49.35\%	&\textbf{57.87}\%    \\
\hline
                 &Thys \etal  &58.17\%    &70.39\%    &77.15\%    &66.75\%    &80.96\%    &76.81\%    &78.96\%    &\textbf{26.64}\%    &\textbf{33.83}\%    &85.46\%    \\
YOLOv5n\cite{jocher2020yolov5}          &Ours(On)  &63.90\%	&88.57\%	&83.94\%	&85.28\%	&91.60\%	&89.49\%	&92.36\%	&37.16\%	&39.97\%	&81.17\%    \\
                 &Ours(Out) &\textbf{58.02}\%	&\textbf{64.36}\%	&\textbf{39.41}\%	&\textbf{66.07}\%	&\textbf{66.60}\%	&\textbf{69.62}\%	&\textbf{67.51}\%	&40.87\%	&46.64\%	&\textbf{66.9}\%     \\
\hline
                 &Thys \etal  &9.98\%     &66.73\%    &79.82\%    &67.12\%    &76.01\%    &76.11\%    &77.54\%    &60.81\%    &45.48\%    &84.29\%    \\
YOLOv5s\cite{jocher2020yolov5}          &Ours(On)  &\textbf{9.25}\%	    &\textbf{65.17}\%	&78.28\%	&63.60\%	&75.13\%	&74.07\%	&76.48\%	&53.72\%	&\textbf{38.13}\%	&83.81\%    \\
                 &Ours(Out) &55.53\%	&68.68\%	&\textbf{58.07}\%	&\textbf{60.68}\%	&\textbf{70.35}\%	&\textbf{73.60}\%	&\textbf{71.47}\%	&\textbf{34.58}\%	&47.98\%	&\textbf{59.42}\%    \\
\hline
                 &Thys \etal  &15.34\%    &69.62\%    &80.61\%    &70.82\%    &75.55\%    &77.28\%    &80.20\%    &61.95\%    &47.18\%    &84.81\%    \\
YOLOv5m\cite{jocher2020yolov5}          &Ours(On)  &\textbf{12.79}\%	&66.94\%	&78.47\%	&67.49\%	&73.53\%	&75.23\%	&78.31\%	&54.49\%	&42.00\%	&83.75\%    \\
                 &Ours(Out) &55.26\%	&\textbf{63.88}\%	&\textbf{51.83}\%	&\textbf{53.57}\%	&\textbf{55.21}\%	&\textbf{62.67}\%	&\textbf{61.53}\%	&\textbf{41.23}\%	&\textbf{41.06}\%	&\textbf{63.61}\%    \\
\hline
                 &Thys \etal  &13.03\%    &68.15\%    &80.96\%    &70.14\%    &76.56\%    &75.33\%    &78.69\%    &64.45\%    &49.17\%    &86.22\%    \\
YOLOv5l\cite{jocher2020yolov5}          &Ours(On)  &\textbf{11.69}\%	&\textbf{65.50}\%	&78.31\%	&67.17\%	&74.20\%	&72.08\%	&75.30\%	&56.37\%	&\textbf{41.16}\%	&84.34\%    \\
                 &Ours(Out) &56.50\%	&67.09\%	&\textbf{57.96}\%	&\textbf{57.39}\%	&\textbf{63.93}\%	&\textbf{54.17}\%	&\textbf{63.86}\%	&\textbf{33.96}\%	&42.60\%	&\textbf{63.08}\%    \\
\hline
                 &Thys \etal  &15.24\%    &69.47\%    &81.92\%    &69.74\%    &77.52\%    &76.77\%    &76.88\%    &64.93\%    &44.51\%    &83.84\%    \\
YOLOv5x\cite{jocher2020yolov5}          &Ours(On)  &\textbf{8.71}\%	    &65.89\%	&77.64\%	&65.67\%	&75.08\%	&74.53\%	&73.90\%	&55.30\%	&\textbf{40.28}\%	&83.62\%    \\
                 &Ours(Out) &52.90\%	&\textbf{65.70}\%	&\textbf{57.87}\%	&\textbf{61.31}\%	&\textbf{64.42}\%	&\textbf{69.03}\%	&\textbf{62.65}\%	&\textbf{40.45}\%	&48.60\%	&\textbf{65.09}\%    \\
\hline
                 &Thys \etal  &\textbf{12.99}\%    &\textbf{68.13}\%    &77.56\%    &73.91\%    &77.66\%    &80.20\%    &81.86\%    &46.93\%    &38.73\%    &84.78\%    \\
Faster R-CNN\cite{ren2015faster}     &Ours(On)  &14.27\%	&76.86\%	&84.57\%	&80.58\%	&85.02\%	&84.35\%	&84.84\%	&32.90\%	&\textbf{34.29}\%	&80.05\%    \\
                 &Ours(Out) &53.15\%	&72.19\%	&\textbf{56.72}\%	&\textbf{70.27}\%	&\textbf{72.01}\%	&\textbf{75.19}\%	&\textbf{74.19}\%	&\textbf{30.27}\%	&49.42\%	&\textbf{57.72}\%    \\
\hline
                 &Thys \etal  &\textbf{14.87}\%    &\textbf{66.80}\%    &77.39\%    &66.96\%    &75.32\%    &74.26\%    &76.43\%    &34.59\%    &\textbf{24.81}\%    &79.32\%    \\
SSD\cite{liu2016ssd}              &Ours(On)  &27.54\%	&72.54\%	&81.50\%	&77.81\%	&80.22\%	&80.60\%	&81.73\%	&\textbf{30.98}\%	&25.14\%	&78.99\%    \\
                 &Ours(Out) &44.84\%	&68.52\%	&\textbf{54.12}\%	&\textbf{66.02}\%	&\textbf{62.93}\%	&\textbf{71.29}\%	&\textbf{67.11}\%	&42.62\%	&44.62\%	&\textbf{62.94}\%    \\
\hline
                 &Thys \etal  &88.23\%    &92.57\%    &89.88\%    &90.55\%    &93.54\%    &93.68\%    &95.22\%    &56.75\%    &60.81\%    &81.97\%    \\
Swin Transformer\cite{liu2021swin} &Ours(On)  &66.06\%	&81.43\%	&84.70\%	&83.82\%	&85.87\%	&86.31\%	&85.75\%	&\textbf{29.63}\%	&\textbf{33.90}\%	&73.61\%    \\
                 &Ours(Out) &\textbf{47.02}\%	&\textbf{72.05}\%	&\textbf{61.75}\%	&\textbf{71.20}\%	&\textbf{69.90}\%	&\textbf{72.57}\%	&\textbf{70.05}\%	&47.02\%	&51.84\%	&\textbf{57.91}\%    \\
\hline\hline
\end{tabular*}
    \begin{tablenotes}
        \footnotesize
        \item Strongest attack results are highlighted in \textbf{bold}.
        \item On and Out mean patches on and outside targets, respectively.
    \end{tablenotes}
\end{threeparttable}
\end{table*}

\begin{figure*}[t!]
\begin{center}
\includegraphics[width=1.7cm]{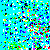}
\includegraphics[width=1.7cm]{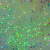}
\includegraphics[width=1.7cm]{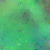}
\includegraphics[width=1.7cm]{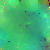}
\includegraphics[width=1.7cm]{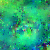}
\includegraphics[width=1.7cm]{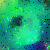}
\includegraphics[width=1.7cm]{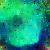}
\includegraphics[width=1.7cm]{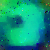}
\includegraphics[width=1.7cm]{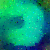}
\includegraphics[width=1.7cm]{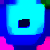}
\end{center}
\begin{center}
\includegraphics[width=1.7cm]{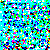}
\includegraphics[width=1.7cm]{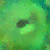}
\includegraphics[width=1.7cm]{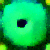}
\includegraphics[width=1.7cm]{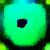}
\includegraphics[width=1.7cm]{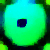}
\includegraphics[width=1.7cm]{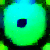}
\includegraphics[width=1.7cm]{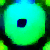}
\includegraphics[width=1.7cm]{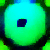}
\includegraphics[width=1.7cm]{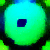}
\includegraphics[width=1.7cm]{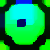}
\end{center}
\begin{center}
\includegraphics[width=1.7cm]{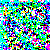}
\includegraphics[width=1.7cm]{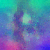}
\includegraphics[width=1.7cm]{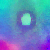}
\includegraphics[width=1.7cm]{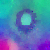}
\includegraphics[width=1.7cm]{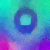}
\includegraphics[width=1.7cm]{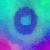}
\includegraphics[width=1.7cm]{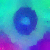}
\includegraphics[width=1.7cm]{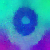}
\includegraphics[width=1.7cm]{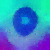}
\includegraphics[width=1.7cm]{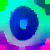}
\end{center}
\caption{Visualization of the adversarial patch (Against Faster R-CNN) optimization process. From top to bottom are the patches crafted by Thys \etal (CVPR) \cite{thys2019fooling}, ours AP-PA (Patch on targets), and AP-PA (Patch outside targets), respectively.}
\label{fig_optimizing_visualization}
\end{figure*}

\begin{table*}[t!]
\caption{Comparison of experimental results of the adversarial patch with different resolutions.}
\label{table_results_comparison_resolution}
\centering
\setlength{\tabcolsep}{1.0mm}
\begin{threeparttable}
\begin{tabular*}{\hsize}{cccccccccccc}
\hline\hline
\diaghead{\theadfont Diag ColumnmnHead}{Patches}{Detectors} &--- &YOLOv2 &YOLOv3 &YOLOv5n    & YOLOv5s   &YOLOv5m    &YOLOv5l    &YOLOv5x    &Faster R-CNN   &SSD    &Swin Transformer
\\ \hline\hline
\multirow{2}{*}{YOLOv2\cite{redmon2017yolo9000}} &150  &18.74\%     &74.95\%    &\textbf{78.03}\%    &76.86\%    &\textbf{78.39}\%    &79.98\%    &81.55\%    &\textbf{14.64}\%    &\textbf{10.77}\%    &\textbf{78.16}\%    \\
           &50  &\textbf{6.33}\%     &\textbf{65.80}\%    &80.18\%	&\textbf{73.05}\%	&80.77\%	&\textbf{78.83}\%	&\textbf{78.44}\%	&68.88\%	&\textbf{47.80}\%	&85.98\%    \\
\hline
\multirow{2}{*}{YOLOv3\cite{redmon2018yolov3}} &150  &\textbf{18.22}\%     &\textbf{59.18}\%    &77.88\%	&73.41\%	&77.11\%	&74.89\%	&78.10\%	&38.95\%	&\textbf{24.62}\%	&83.43\%    \\
           &50  &19.38\%	&59.24\%    &\textbf{75.36}\%	&\textbf{66.43}\%	&\textbf{74.20}\%	&\textbf{72.54}\%	&\textbf{75.40}\%	&\textbf{35.35}\%	&28.88\%	&\textbf{82.78}\%    \\
\hline
\multirow{2}{*}{YOLOv5n\cite{jocher2020yolov5}} &150  &\textbf{20.55}\%    &\textbf{74.20}\%    &\textbf{77.52}\%    &\textbf{73.00}\%    &\textbf{78.97}\%    &\textbf{81.15}\%    &\textbf{82.96}\%    &58.70\%    &\textbf{33.66}\%    &85.09\%    \\
          &50  &63.90\%	&88.57\%	&83.94\%	&85.28\%	&91.60\%	&89.49\%	&92.36\%	&\textbf{37.16}\%	&39.97\%	&\textbf{81.17}\%    \\
\hline
\multirow{2}{*}{YOLOv5s\cite{jocher2020yolov5}} &150  &14.36\%     &68.57\%    &80.73\%    &69.29\%    &75.96\%    &76.86\%    &78.88\%    &59.25\%    &\textbf{34.47}\%    &84.99\%    \\
          &50  &\textbf{9.25}\%	    &\textbf{65.17}\%	&\textbf{78.28}\%	&\textbf{63.60}\%	&\textbf{75.13}\%	&\textbf{74.07}\%	&\textbf{76.48}\%	&\textbf{53.72}\%	&38.13\%	&\textbf{83.81}\%    \\
\hline
\multirow{2}{*}{YOLOv5m\cite{jocher2020yolov5}} &150  &14.02\%    &\textbf{66.10}\%    &78.81\%    &\textbf{67.47}\%    &\textbf{71.41}\%    &\textbf{72.71}\%    &\textbf{77.38}\%    &\textbf{49.37}\%    &\textbf{40.71}\%    &\textbf{83.10}\%    \\
          &50  &\textbf{12.79}\%	&66.94\%	&\textbf{78.47}\%	&67.49\%	&73.53\%	&75.23\%	&78.31\%	&54.49\%	&42.00\%	&83.75\%    \\
\hline
\multirow{2}{*}{YOLOv5l\cite{jocher2020yolov5}} &150  &16.67\%    &66.75\%    &81.09\%    &70.42\%    &74.84\%    &72.09\%    &78.31\%    &59.94\%    &42.09\%    &84.42\%    \\
          &50  &\textbf{11.69}\%	&\textbf{65.50}\%	&\textbf{78.31}\%	&\textbf{67.17}\%	&\textbf{74.20}\%	&\textbf{72.08}\%	&\textbf{75.30}\%	&\textbf{56.37}\%	&\textbf{41.16}\%	&\textbf{84.34}\%    \\
\hline
\multirow{2}{*}{YOLOv5x\cite{jocher2020yolov5}} &150  &18.84\%    &71.36\%    &84.22\%    &72.68\%    &78.54\%    &77.45\%    &77.62\%    &67.36\%    &44.79\%    &84.49\%    \\
          &50  &\textbf{8.71}\%	    &\textbf{65.89}\%	&\textbf{77.64}\%	&\textbf{65.67}\%	&\textbf{75.08}\%	&\textbf{74.53}\%	&\textbf{73.90}\%	&\textbf{55.30}\%	&\textbf{40.28}\%	&\textbf{83.62}\%    \\
\hline
\multirow{2}{*}{Faster R-CNN\cite{ren2015faster}} &150  &\textbf{12.46}\%    &\textbf{73.93}\%    &\textbf{83.44}\%    &\textbf{78.66}\%    &\textbf{83.21}\%    &\textbf{81.66}\%    &\textbf{84.25}\%    &\textbf{26.18}\%    &\textbf{24.18}\%    &80.51\%    \\
     &50  &14.27\%	&76.86\%	&84.57\%	&80.58\%	&85.02\%	&84.35\%	&84.84\%	&32.90\%	&\textbf{34.29}\%	&\textbf{80.05}\%    \\
\hline
\multirow{2}{*}{SSD\cite{liu2016ssd}} &150  &\textbf{9.78}\%    &\textbf{63.78}\%    &\textbf{75.01}\%    &\textbf{63.65}\%    &\textbf{73.21}\%    &\textbf{72.75}\%    &\textbf{76.93}\%    &\textbf{28.43}\%    &\textbf{13.46}\%    &80.02\%    \\
              &50  &27.54\%	&72.54\%	&81.50\%	&77.81\%	&80.22\%	&80.60\%	&81.73\%	&\textbf{30.98}\%	&25.14\%	&\textbf{78.99}\%    \\
\hline
\multirow{2}{*}{Swin Transformer\cite{liu2021swin}} &150  &92.04\%    &94.12\%    &92.25\%    &93.35\%    &94.88\%    &95.65\%    &96.13\%    &75.01\%    &73.44\%    &86.47\%    \\
 &50  &\textbf{66.06}\%	&\textbf{81.43}\%	&\textbf{84.70}\%	&\textbf{83.82}\%	&\textbf{85.87}\%	&\textbf{86.31}\%	&\textbf{85.75}\%	&\textbf{29.63}\%	&\textbf{33.90}\%	&\textbf{73.61}\%    \\
\hline\hline
\end{tabular*}
    \begin{tablenotes}
        \footnotesize
        \item Strongest attack results are highlighted in \textbf{bold}.
    \end{tablenotes}
\end{threeparttable}
\end{table*}

\begin{figure}[t!]
\begin{center}
\includegraphics[width=4.3cm]{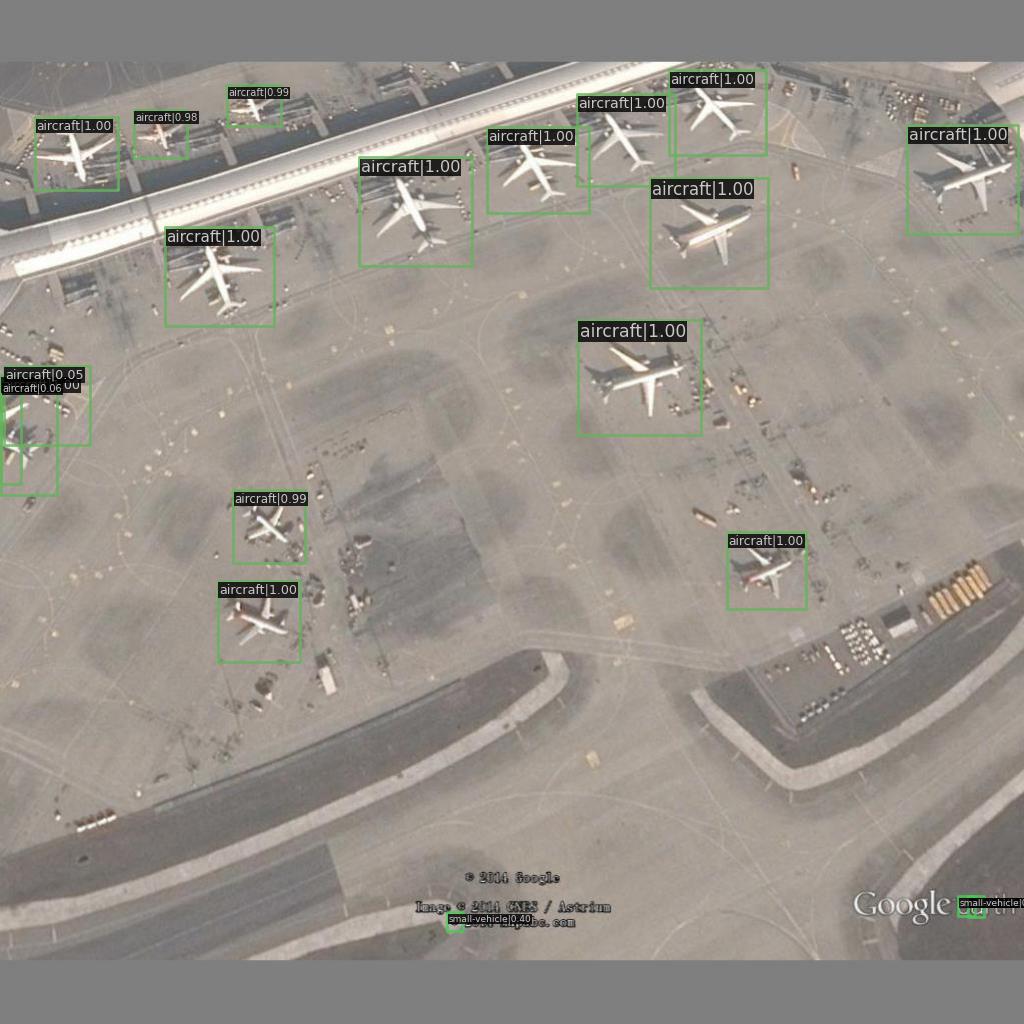}
\includegraphics[width=4.3cm]{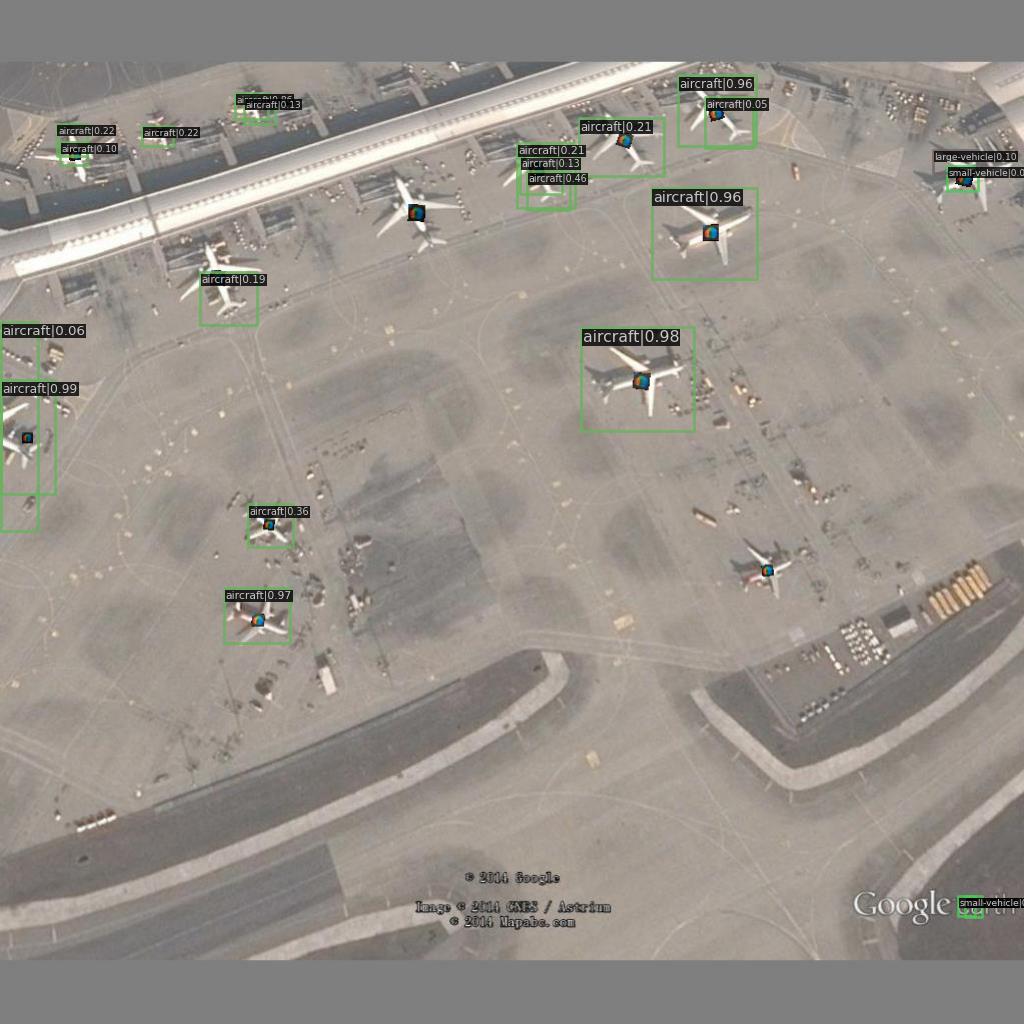}
\end{center}
\caption{Comparison of visual examples before and after adversarial attack with predicted probabilities.}
\label{fig_probability_results_example_physical}
\end{figure}

\begin{figure}[t!]
\begin{center}
\includegraphics[width=8.0cm]{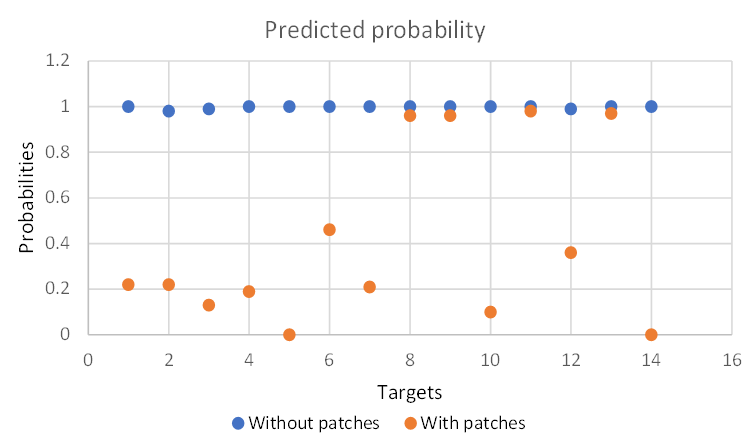}
\end{center}
\caption{The predicted probabilities of targets before and after the adversarial attack in the physical condition.}
\label{fig_probability_results_comparison_physical}
\end{figure}

\begin{figure*}[t!]
\begin{center}
\includegraphics[width=3.5cm]{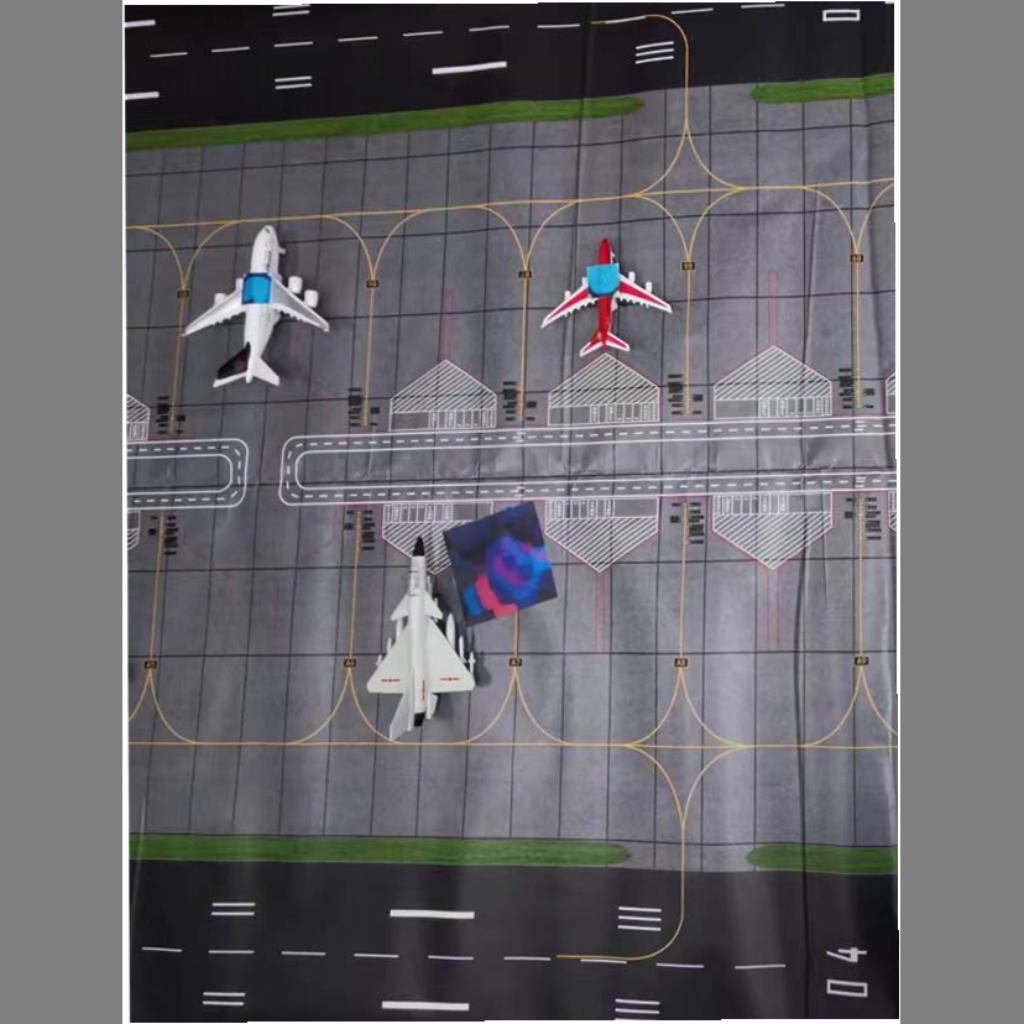}
\includegraphics[width=3.5cm]{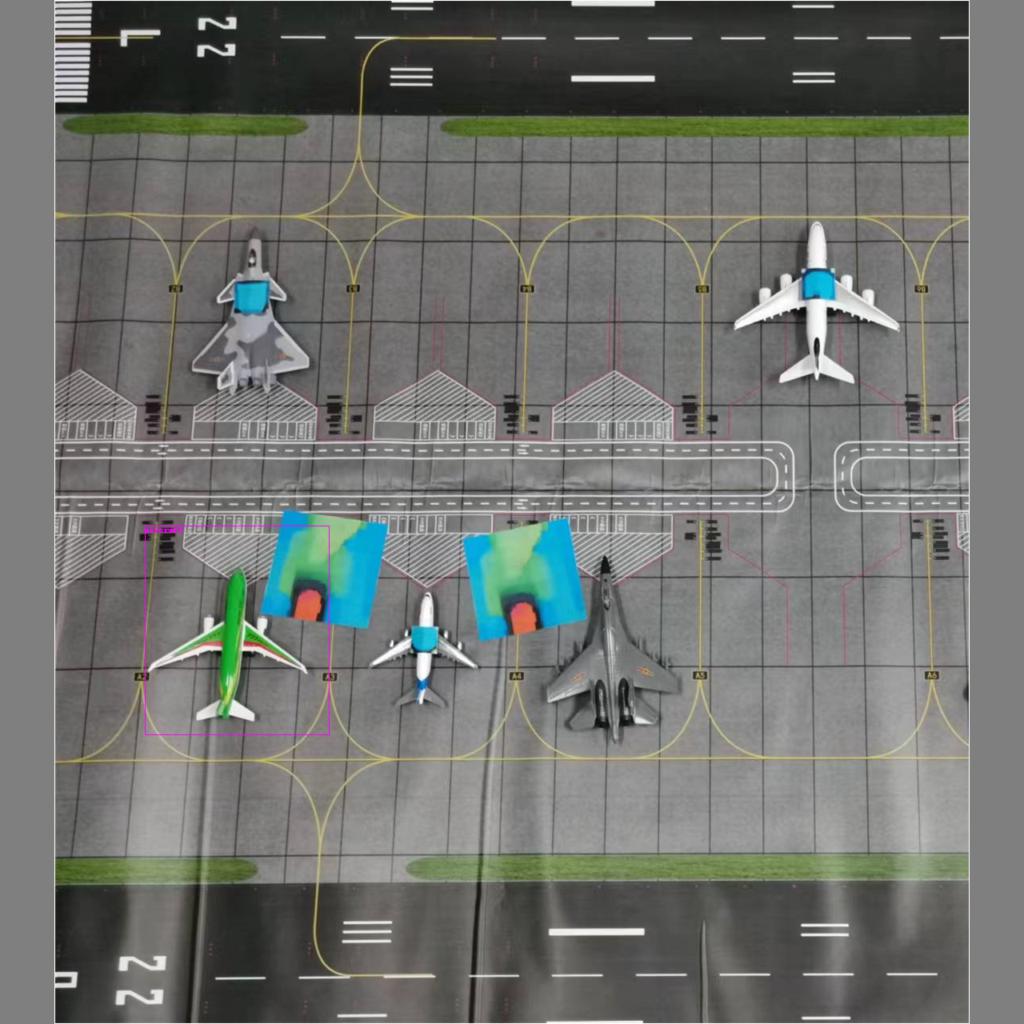}
\includegraphics[width=3.5cm]{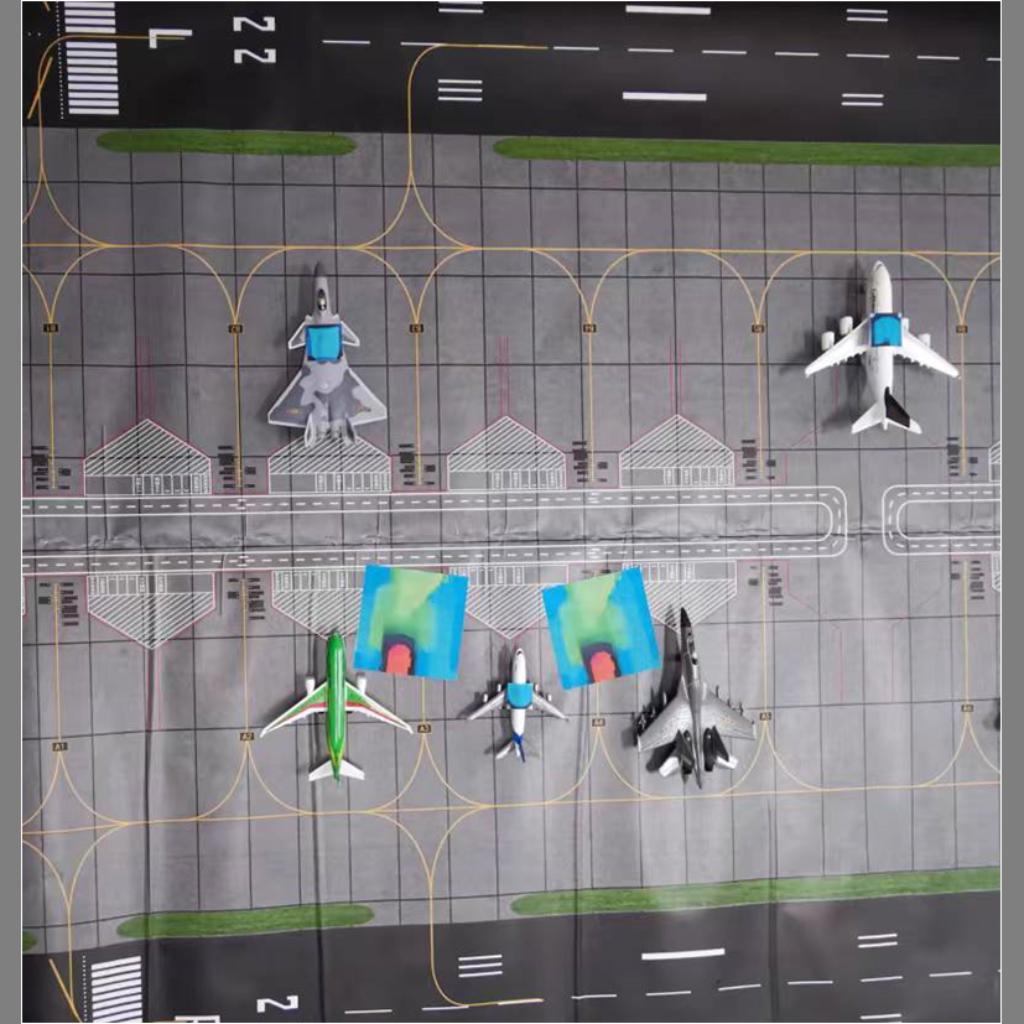}
\includegraphics[width=3.5cm]{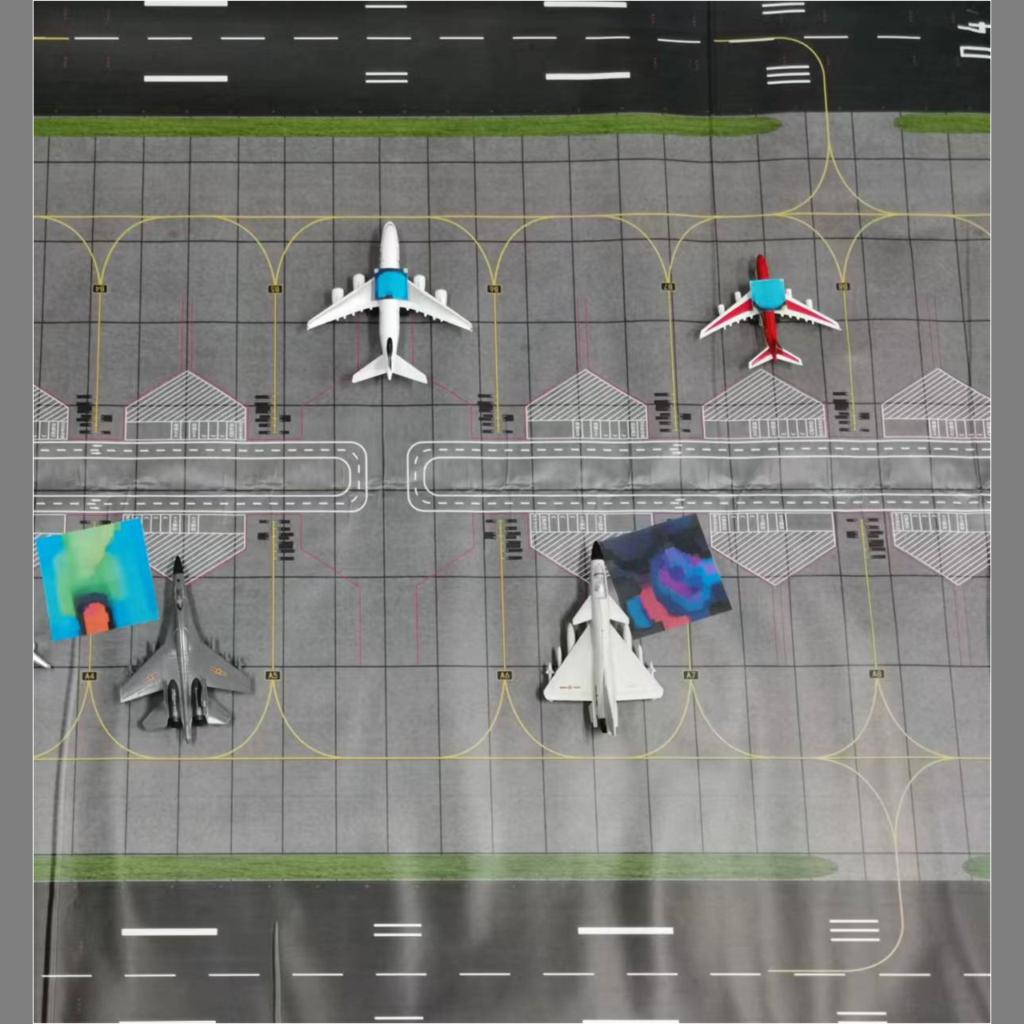}
\includegraphics[width=3.5cm]{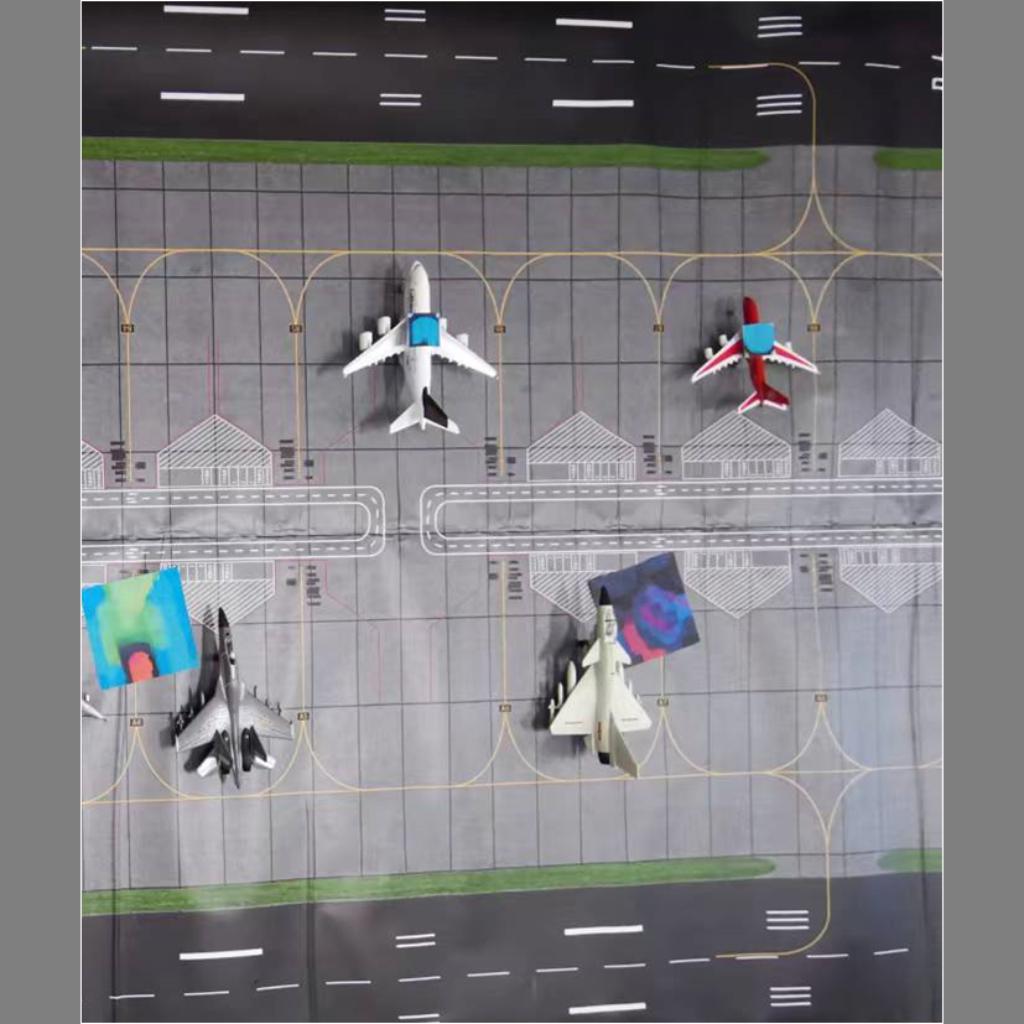}
\end{center}
\begin{center}
\includegraphics[width=3.5cm]{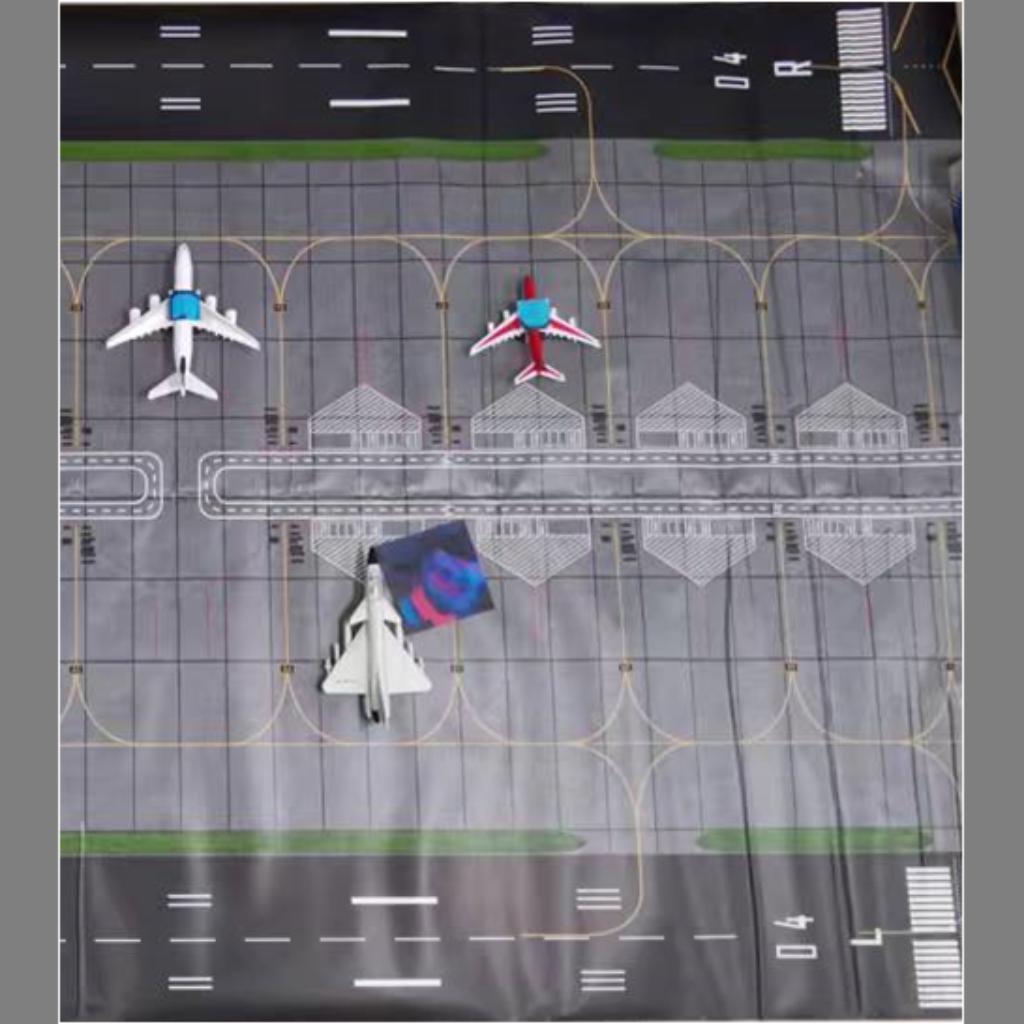}
\includegraphics[width=3.5cm]{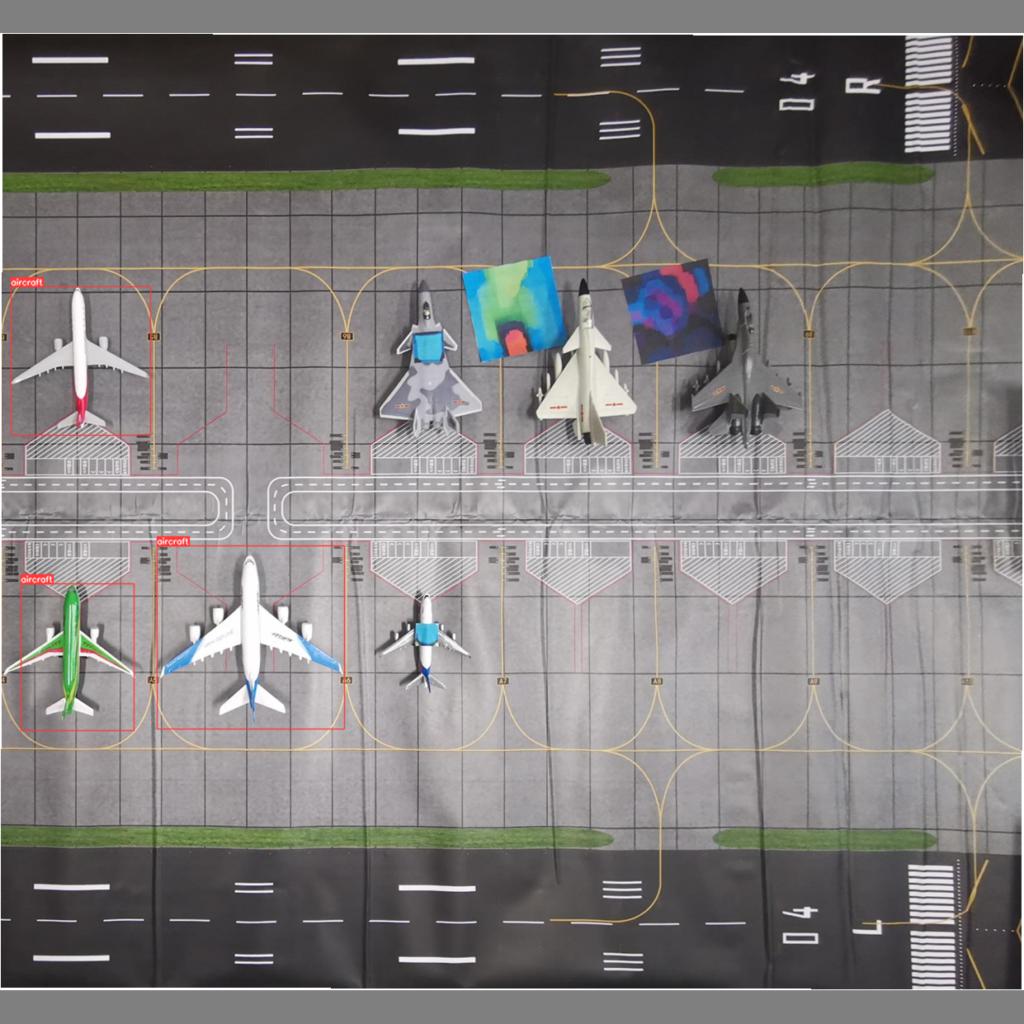}
\includegraphics[width=3.5cm]{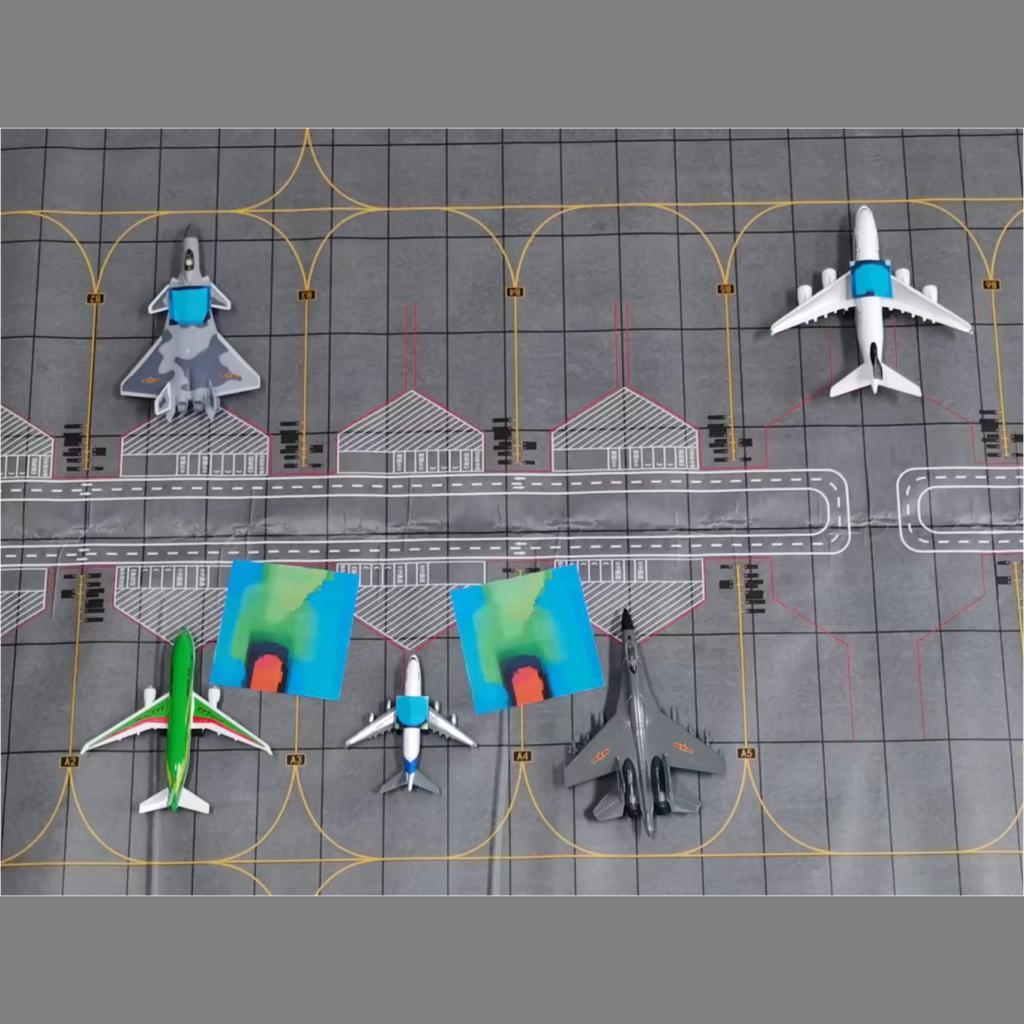}
\includegraphics[width=3.5cm]{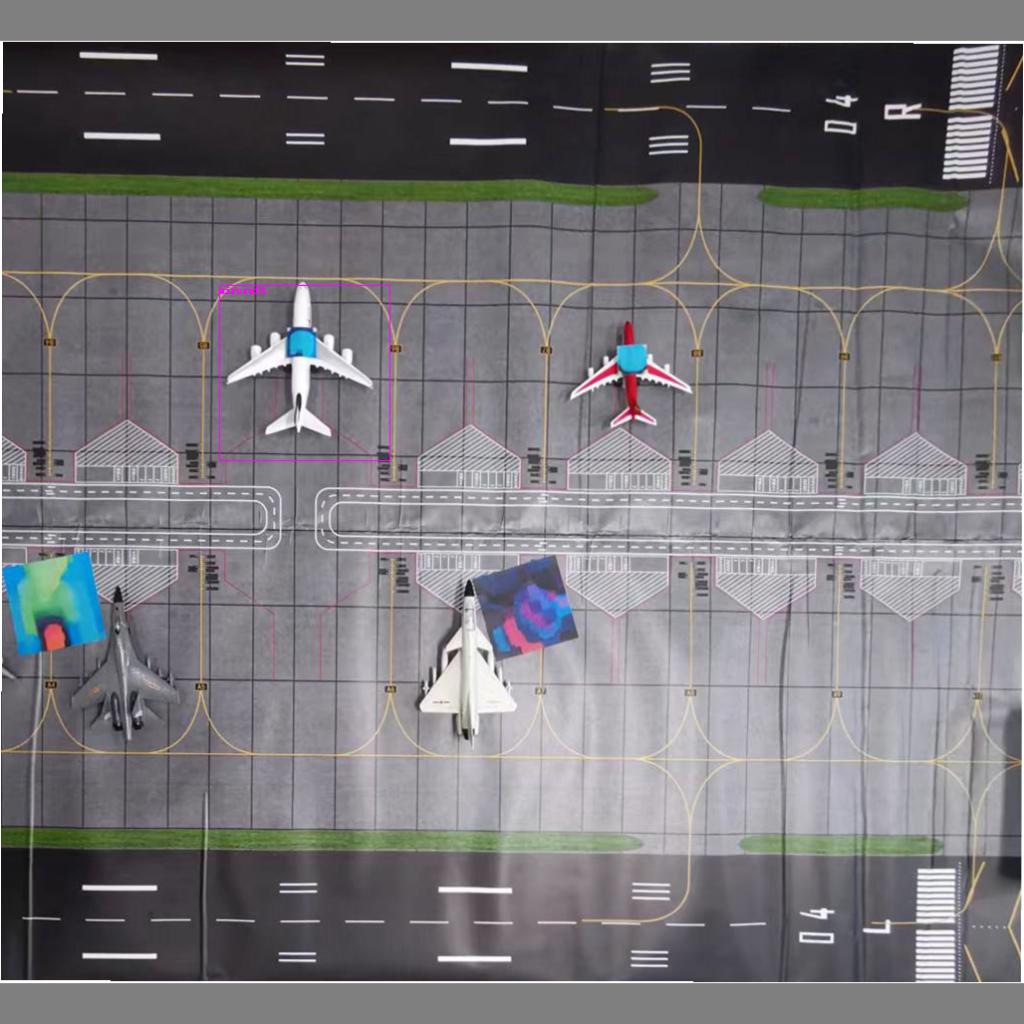}
\includegraphics[width=3.5cm]{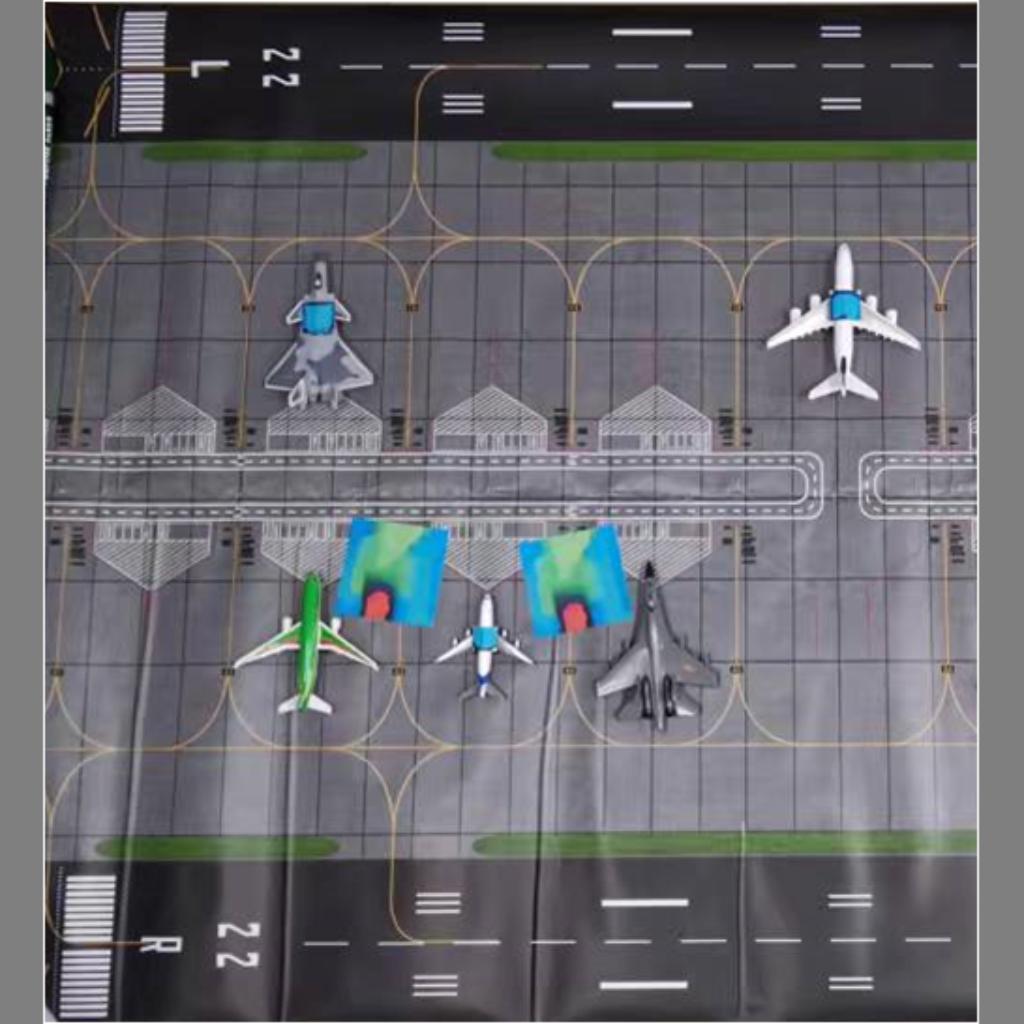}
\end{center}
\caption{Visual examples of attack effectiveness in the proportionally scaled scenarios with different physical conditions including angles and distances, where A and D represent angle and distance, respectively.}
\label{fig_visual_examples_physical}
\end{figure*}

We report the attack effectiveness of our proposed AP-PA method against YOLOv2, YOLOv3, YOLOv5n, YOLOV5s, and YOLOv5m, YOLOv5l, YOLOv5x, Faster R-CNN, SSD, and Swin Transformer, respectively. We place the adversarial patch both on and outside the target to perform a dodging attack under the aerial detection task and evaluate the recall, precision, and AP, respectively. The quantitative descriptions of attack efficacy in detail are displayed in Table \ref{table_on_target} and Table \ref{table_outside_target}. Moreover, visual representations of Table \ref{table_on_target} and Table \ref{table_outside_target} are displayed in Fig.\ref{fig_3d_bar_on} and Fig.\ref{fig_3d_bar_outside} respectively to show the patches' attack transferability between different aerial detectors.

It is clear that in all white-box attack scenarios where the target model is the same as the model for generating adversarial patches (proxy model), our proposed AP-PA method can dramatically drop the detection AP of aerial detectors, especially for some relatively earlier methods such as YOLOv2 ($6.33\%$ and $20.72\%$), Faster R-CNN ($32.90\%$ and $30.27\%$), and SSD ($21.14\%$ and $44.62\%$). For black-box attack scenarios, namely that the target model is different from the proxy model. Similarly, the AP-PA can also achieve strong attack effectiveness and good transferability as shown in Fig.\ref{fig_3d_bar_on} and Fig.\ref{fig_3d_bar_outside}. An interesting discovery is that the effect of a transfer-based attack has a close relationship with the robustness of the target model. In other words, the more robust the target model is the worse performance a transfer-based attack can achieve and vice versa.

The P-R curve of each aerial detector is displayed in Fig.\ref{fig_pr_curve_center} and Fig.\ref{fig_pr_curve_outside}. From these P-R curves, we can observe the influence of elaborated adversarial patches compared to random noise patches (Contrasting the effect of occlusion). A widely adopted way to choose a nice point on the P-R curve for detection is to draw a diagonal line on the P-R curve. In this work, we use 0.4 as a reference and set the detection results of clean images as 100\% AP, the proposed AP-PA can significantly drop the precision and recall of the aerial detectors no matter the patches on or outside targets. Finally, several visual examples of attack are given in Fig.\ref{fig_visual_examples_on} and Fig.\ref{fig_visual_examples_outside}.

\subsubsection{Comparisons}

We adopt a state-of-the-art physical attack approach from the work \cite{thys2019fooling} (CVPR) as the comparison algorithm due to the existing related physical attack methods \cite{den2020adversarial,lu2021scale,du2021adversarial,du2022physical} against aerial detectors all derived from this method. The target detector is still YOLOv2, YOLOv3, YOLOv5n, YOLOV5s, YOLOv5m, YOLOv5l, YOLOv5x, Faster R-CNN, SSD, and Swin Transformer. In addition, we also take patches with different positions into comparison.

In Table \ref{table_results_comparison_digital}, we compare the attack efficacy of the proposed AP-PA with another state-of-the-art method from \cite{thys2019fooling} (CVPR). We can see that the proposed AP-PA approach can generate adversarial patches with stronger fooling effectiveness, which can significantly drop the detection AP of aerial detectors both in white-box and black-box settings than the method in \cite{thys2019fooling} in most circumstances. Moreover, we can also see better attack transferability by using adversarial patches elaborated by our proposed AP-PA approach.

Three adversarial patches from different algorithms against Faster R-CNN are chosen to compare the optimization process. The optimizing process of adversarial patches is visualized and displayed in Fig.\ref{fig_optimizing_visualization}. Several adversarial patches are selected (Iterations: 0, 200, 400, 600, 800, 1000, 1200, 1400, 1600) from the optimizing process to analyze the evolutionary progress of adversarial patches. We can observe that our method can enormously accelerate the optimizing efficiency of adversarial patches because the selected adversarial patches have greater similarity with the final optimized adversarial patches with stable patterns.

The main reason for the better attack performance of our AP-PA is that we consider all detected objects, \ie, using the mean scores of all detected objects to optimize adversarial patches instead of the only one object with the biggest objectiveness score (\cite{thys2019fooling,den2020adversarial,lu2021scale,du2021adversarial,du2022physical}), which can not only greatly drop the number of detected objects but also significantly improve the optimizing efficiency of adversarial patches.

Unquestionably, the bigger the patch size, the stronger the attack efficacy. In this work, we also delve into the influences of another attribution of the adversarial patch, namely the resolution of the adversarial patch. We compare the different resolutions of adversarial patches (50 and 150, respectively) to see the impact of the patch's resolution. The comparison result is shown in Table \ref{table_results_comparison_resolution}, we can observe that the attack effectiveness of adversarial patches is slightly swayed by its resolution for most cases, while a stronger adversarial patch is acquired for attacking Swin Transformer by reducing the patch's resolution. During the rest part of the experiments, we adopt $50\times50$ as patches' resolution, because we believe that the simpler the adversarial patches are, the less loss of attack efficacy during the physical-digital transformation.

\subsection{Proportionally scaled validation experiments in Physical Domain}
\label{section3}

In this part, we report the attack effectiveness of adversarial patches elaborated by our AP-PA method in physical scenarios. Firstly, we place the adversarial patches acquired by a camera on the targets of an aerial image from the public dataset, Figs.\ref{fig_probability_results_example_physical} and \ref{fig_probability_results_comparison_physical} show the predicted probabilities of several targets corresponding to the ground-truth label before and after conducting attacks in real scenarios. The results demonstrate that the predicted probabilities and IOUs of different targets have been significantly dropped, and the maximum reduction is 1.00, which means the targets can not be detected at all. This illustrates that the adversarial patches generated digitally by the proposed AP-PA method can maintain a stable attack efficacy when applied to real physical scenarios.

In addition, we use aircraft models to simulate aerial detection in real scenarios with different angles and distances, and the experimental results are shown in Fig.\ref{fig_visual_examples_physical}. It illustrates that the proposed AP-PA can be robust enough to achieve strong attack effectiveness for most targets with different physical conditions, such as image acquire angles and distances, in real proportional scaled scenarios.

\section{Conclusion and future work}

In this article, we proposed the AP-PA adversarial attack algorithm, a physically practicable attack based on adversarial patches for both white-box and black-box settings in real physical scenarios. We perform attacks based on adversarial patches in both digital and physical domains. However, attacking aerial detectors poses a greater challenge than attacking image classifiers, especially broadening attacks from the digital domain to real physical scenarios, which requires the adversarial patch to be robust enough to survive real-world distortions due to some uncontrollable physical dynamics, such as different viewing distances, object scales, and lighting conditions. To solve the above problems, we devised the AP-PA method, which aims to generate adversarial patches to hide objects from aerial detectors in the physical world. To reach this, aircraft is chosen as the target object to conduct experiments, and the different target objects can be simply derived. During the training process, the weights and bias of the targeted aerial detectors should be fixed, instead, the pixel values of the adversarial patch should be updated iteratively, which means we are ``training'' a patch instead of a model. Furthermore, a new loss is devised to consider more available information of detected objects to optimize the adversarial patches, which can significantly improve the patch’s attack efficacy and optimize efficiency. In addition, most of the existing adversarial attack methods focus on digital attacks and individual object detectors. So we also establish one of the first comprehensive, coherent, and rigorous benchmarks to evaluate the attack robustness of adversarial patches on aerial detection tasks in both digital and physical domains. Extensive experiments on aerial detection in both white-box and black-box settings demonstrated the robust attack efficacy and transferability of our proposed AP-PA.

In contrast with imperceptible perturbations, patch-based attacks are easier to fool aerial detectors in real physical scenarios due to the adversarial perturbations concentrated on a small area, which can be easily captured by imaging devices with less loss and distortion of adversarial attack efficacy. In this paper, we focus on improving the generating efficiency and attack efficacy of adversarial patches and comprehensively evaluating adversarial patches. In future work, we would like to camouflage adversarial patches and further improve the attack effectiveness against robust detectors, such as YOLOv5, Swin Transformer, \etc. Additionally, other directions where more research should be conducted are to search for the patch's optimal position and shape.

\bibliographystyle{IEEEtran}
\bibliography{references}

\end{document}